\documentclass{article} 
\usepackage{iclr2027_conference,times}

\usepackage{amsmath,amsfonts,bm}

\def\eqref#1{equation~\ref{#1}}

\def\1{\bm{1}}

\DeclareMathAlphabet{\mathsfit}{\encodingdefault}{\sfdefault}{m}{sl}
\SetMathAlphabet{\mathsfit}{bold}{\encodingdefault}{\sfdefault}{bx}{n}

\usepackage{wrapfig}
\usepackage{graphicx}
\usepackage{hyperref}
\usepackage{url}
\usepackage{booktabs}
\usepackage{graphicx}
\usepackage{enumitem}
\usepackage{wrapfig}
\usepackage{caption}
\usepackage{pgfplots}
\pgfplotsset{compat=1.17}
\usepackage{amssymb}
\usepackage{placeins}
\usepackage{titletoc}
\usepackage{multicol}
\usepackage{needspace}
\titlecontents{part}[0em]{\addvspace{6pt}\scshape\bfseries}{}{}{}[\addvspace{1pt}]
\titlecontents{section}[1.9em]{\addvspace{2pt}\bfseries}{\contentslabel{1.9em}}{\hspace*{-1.9em}}{\hfill\contentspage}
\titlecontents{subsection}[4.4em]{\addvspace{0.5pt}}{\contentslabel{2.5em}}{\hspace*{-2.5em}}{\titlerule*[0.7pc]{.}\contentspage}
\usepackage{pifont}
\usepackage{multirow}
\usepackage{tabularx}
\usepackage{fvextra}
\usepackage{newunicodechar}
\newunicodechar{→}{\ensuremath{\rightarrow}}
\newunicodechar{←}{\ensuremath{\leftarrow}}
\newunicodechar{—}{\textemdash}
\newunicodechar{–}{\textendash}
\newunicodechar{×}{\ensuremath{\times}}
\newunicodechar{°}{\ensuremath{^\circ}}
\newunicodechar{≥}{\ensuremath{\geq}}
\newunicodechar{≤}{\ensuremath{\leq}}
\newunicodechar{±}{\ensuremath{\pm}}
\newunicodechar{…}{\ldots}
\newunicodechar{·}{\ensuremath{\cdot}}
\newunicodechar{✗}{\ding{55}}
\newunicodechar{✓}{\ding{51}}
\newunicodechar{’}{'}
\newunicodechar{‘}{`}
\newunicodechar{“}{``}
\newunicodechar{”}{''}
\newunicodechar{Δ}{\ensuremath{\Delta}}
\newcommand{\appendixpart}[1]{\par\needspace{14\baselineskip}\addvspace{10pt}\phantomsection\addcontentsline{toc}{part}{#1}%
  {\noindent\large\scshape\bfseries #1\par}\nobreak\vspace{2pt}\hrule\nobreak\vspace{6pt}\nobreak}
\newcommand{\listingfile}[1]{\VerbatimInput[fontsize=\scriptsize,breaklines,frame=lines,framesep=1.5mm]{#1}}

\title{A Controlled Study of Models, Skills, \\ Feedback, and Curriculum \\ in Agentic Robotics}

\author{Kaiyuan Chen$^{3*}$, Xinyue Zhu$^{3*}$, Shuangyu Xie$^{3}$, Uksang Yoo$^{3}$, Thi Thu Phuong Luong$^{1}$, \\
\bf Marcus Gualtieri$^{1}$, Christian Juette$^{1}$, Liu Ren$^{1}$, Spencer Huang$^{2}$, Ken Goldberg$^{3}$ \\
\normalfont $^{1}$Bosch \quad $^{2}$NVIDIA \quad $^{3}$UC Berkeley \\
\normalfont $^{*}$Equal contribution}

\iclrfinalcopy 
\begin{document}

\maketitle
\lhead{Preprint}

\begin{abstract}

Agentic robotics asks coding agents to implement and iteratively improve robot policies from simulated and real-world physical interactions. We study the self-learning loop of agentic robotics through a controlled simulation study spanning nine manipulation tasks, 631 different experiment configurations, and 28K coding agent rollouts, systematically varying model capability, skill priors, feedback structure, and curriculum. The experiment data suggests that the bottleneck is often not the access to sufficient information, but interpreting it into the right policy update. 
Across these factors, a successful self-learning loop depends on how effectively agents attribute failures, reason task-relevant spatial and geometric structure, and selectively restructure code to improve policies.

\end{abstract}

\begin{figure}[!ht]
    \centering
    \includegraphics[width=0.8\linewidth]{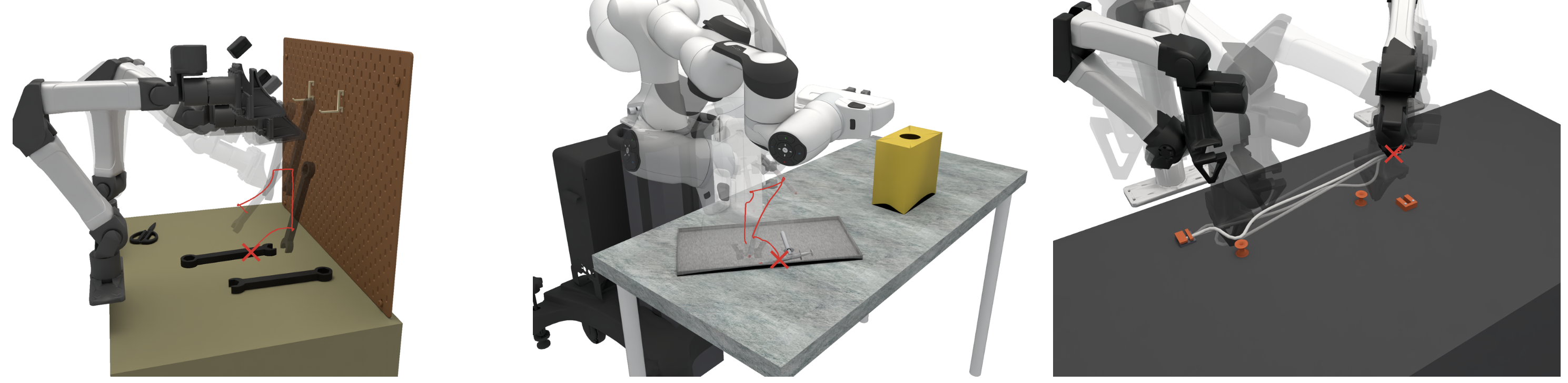}
    \caption{\textbf{Representative failure modes in Agentic Robotics.}
    Left: \emph{Manipulation planning} fails to select a grasp and approach strategy compatible with the downstream placement.
    Center: \emph{Motion planning} fails to jointly reason about held-object geometry, obstacles, and clearance to produce a collision-free trajectory.
    Right: \emph{Deformable manipulation} requires reasoning over object configuration and contact that is difficult to capture through local motion decisions.}
    \label{fig:teaser}
\end{figure}

\section{Introduction}

Recent coding agents, such as GPT Astra ~\citep{openai2026astra} and Claude Fable ~\citep{anthropic2026fable51}, have become increasingly capable of generating and iteratively repairing executable programs, motivating a growing class of Agentic Robotics (AR) systems in which LLM- and VLM-based agents write code that composes perception, planning, and control skills into robot policies~\citep{liang2023code,singh2022progprompt,fu2026capx,chen2026gap}.
However, self-improvement in robotics differs fundamentally from software debugging. In software, failures, such as compiler errors, are often exposed through symbolic feedback that is already closely aligned with the program; but in robotics, the evidence to improve a policy is instead embedded in physical interactions: object geometry, spatial relationships, contact, motion, and changes in the scene over time. As illustrated in Figure~\ref{fig:teaser}, a failed placement can be attributed to an earlier incompatible grasp and a collision requires reasoning about the geometry of an object after it becomes part of the moving robot. 
In this paper, we ask: {what determines whether an agent can interpret physical interaction and convert it into effective robot-policy improvement?}

\begin{figure}[t]
  \centering
  \includegraphics[width=0.8\textwidth]{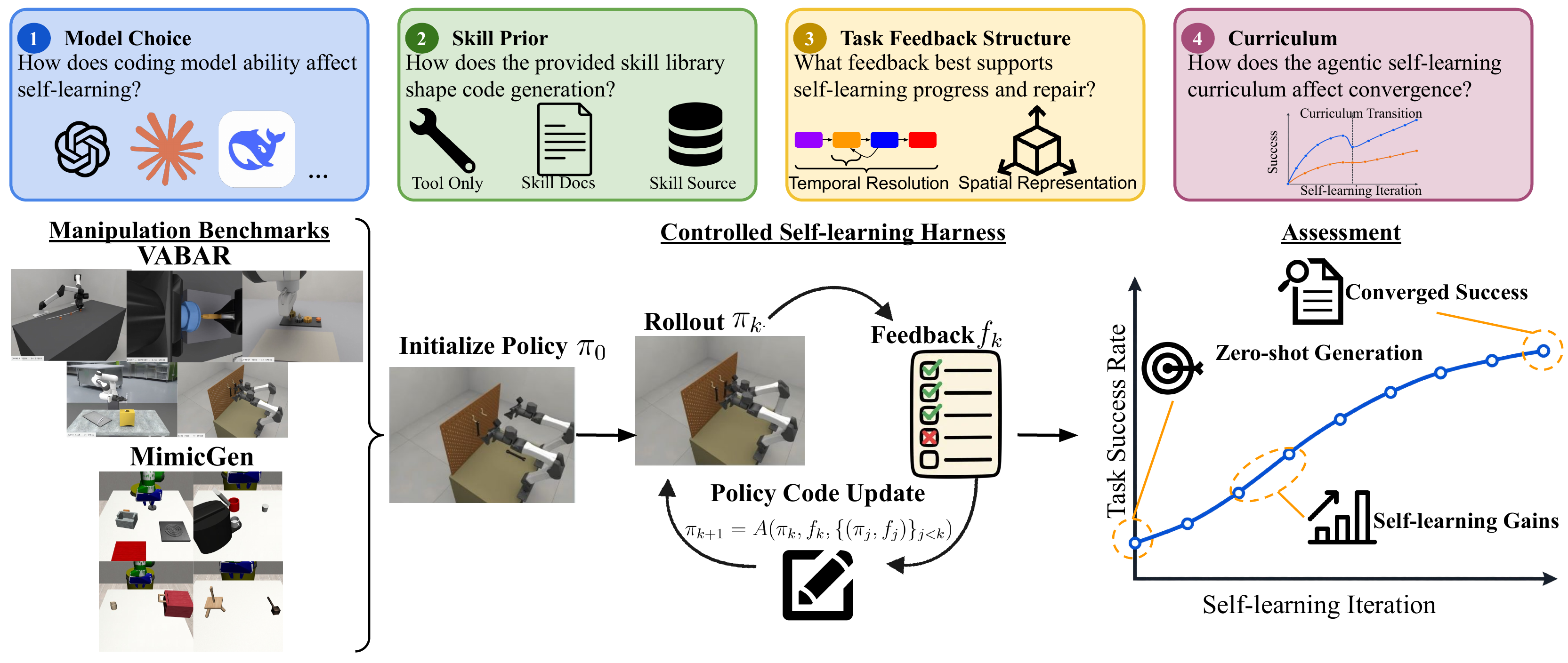}
  \caption{
    Overview of our controlled study of agentic robot self-learning of {model}, {skill}, {feedback}, and {curriculum design}. A coding agent initializes a policy $\pi_0$, executes the current policy $\pi_k$ in simulation, receives feedback $f_k$, and iteratively edits its policy as $\pi_{k+1}=A\!\left(\pi_k,f_k,\{(\pi_j,f_j)\}_{j<k}\right)$. We evaluate zero-shot and self-learning performance, and final task success across VABAR and MimicGen manipulation tasks.
  }
  \vspace{-10pt}
  \label{fig:overview}
\end{figure}

We design a controlled simulation study of agentic robot self-learning across five contact-rich VABAR manipulation tasks~\cite{vabar} and four longer-horizon MimicGen tasks~\cite{mandlekar2023mimicgen}. Holding the overall self-learning harness fixed, we systematically vary four components of the learning loop: {model capability}, {skill prior}, {feedback structure}, and {curriculum}, as illustrated in Figure~\ref{fig:overview}.
Across these studies, we find that self-learning is limited less by the availability of useful information than by the agent's ability to turn the information into effective policy updates. More execution feedback is not necessarily more useful: feedback is most effective when aligned with semantically meaningful policy decisions rather than reported at the finest temporal resolution. Similarly, world-frame states contain all the information needed for manipulation, yet agents often fail to recover task-relevant object-relative and functional relationships unless these are made explicit. Even when agents identify local geometric errors, they often struggle to translate them into coherent collision-aware motion plans. We find that this limitation can also affect curriculum design:  agents tend to preserve and patch policy structure inherited from earlier stages rather than fundamentally reorganize it, so privileged learning can either provide a useful initialization or constrain later learning when the inherited policy depends on privileged precision or brittle manipulation strategies.
Our results further suggest that progress in agentic robot self-learning comes from  the better structures of skills, feedback, and state representations and the learned physical reasoning ability from the underlying models.

This paper makes three contributions:
(1) {A controlled study design for Agentic Robotics} that isolates four important components of simulation-guided self-learning, (2) Experiment data from 631 implemented experiments with 28 thousand simulation rollouts on Agentic robotics policies, (3) Six findings on model capability, skill priors, feedback structure, and curriculum.

\section{Related Work}


\textbf{Agentic Robotics.} Foundation models have been used to compose robot skills and generate executable policies through language-model planning and code generation~\citep{codeaspolicies,ahn2022can,singh2022progprompt}. More recent work integrates foundation models with explicit spatial representations, geometric constraints, and task-and-motion planning to support open-vocabulary and multi-step manipulation~\citep{huang2023voxposer,huang2024rekep,shen2026tiptop,murray2025showtell}. Other approaches close the loop by using simulation or real-world execution feedback to revise policies, acquire skills, and improve task performance over repeated interactions~\citep{fu2026capx,chen2026gap,enpire,aspire}. Rather than introducing a new agent architecture, we study how individual components of this self-learning loop affect performance.

\textbf{Controlled Evaluation of Agentic Robotics.} Robot-learning benchmarks increasingly use standardized task suites to evaluate generalization, long-horizon reasoning, and manipulation under controlled variation~\citep{gu2023maniskill2,behavior1k,calvin,liu2023libero,zheng2022vlmbench}. More recent benchmarks extend controlled evaluation to foundation-model agents by measuring interactive decision making, embodied reasoning, perception, and action across diverse environments~\citep{liu2024agentbench,li2024eai,yang2025embodiedbench}. Robot-specific evaluations further study robot-centric perception and decision making, coding-agent behavior under different abstraction and interaction settings, and repeated manipulation under controlled task variation~\citep{xie2026robovista,fu2026capx,vabar}. Unlike prior agentic robotics benchmarks that primarily vary tasks, environments, or complete agent systems, we hold the self-learning harness fixed and independently vary model capability, skill priors, feedback structure, and curriculum to isolate how each factor affects initial policy generation and subsequent self-improvement.

\textbf{Skills, Feedback, and Curriculum for Self-Learning.} Prior work has shown that reusable skills and execution feedback can improve embodied agent learning~\citep{ahn2022can,wang2023voyager,huang2022inner,fu2026capx}. Skill libraries provide reusable action abstractions that constrain generation and support compositional behavior, while systems such as Voyager jointly combine an executable skill library, environment feedback, and an automatic curriculum for continual improvement~\citep{wang2023voyager}. More broadly, language agents can use environment observations, linguistic critiques, or tool-derived feedback to revise plans and outputs across interaction or repeated attempts~\citep{yao2023react,shinn2023reflexion,madaan2023selfrefine,gou2024critic}. However, prior work also shows that self-correction can be unreliable without sufficiently informative external feedback~\citep{huang2024large}. Curriculum learning instead structures the progression of experience~\citep{bengio2009curriculum}, while privileged-information methods use state available during training to simplify learning before transferring to image-based or otherwise non-privileged observations~\citep{pinto2017asymmetric,chen2020learning}. These factors are often introduced jointly within a single system. We instead study them separately by varying the skill prior, temporal and spatial feedback representation, and privileged-to-vision curriculum.
\section{Problem Formulation}
\label{sec:protocol}

\paragraph{Agentic robotics for manipulation.}
A robot manipulation task class can be defined as a tuple
$\mathcal{T} = \langle \mathcal{L}, \mathcal{E}, \mathcal{R}, \mathcal{O}, \mathcal{X}, \mathcal{B}, \mathcal{J} \rangle$: a natural-language instruction $\mathcal{L}$, a fixed workcell $\mathcal{E}$ with a known robot and sensor configuration $\mathcal{R}$, a bounded set of objects $\mathcal{O}$ with known models, a state space $\mathcal{X}$ of robot configurations and object poses, a belief space $\mathcal{B}$ giving the distribution over initial object poses, and a reward $\mathcal{J}$, here the per-object success indicator $\mathbb{I}(\text{success})$. A task instance $\tau_i = \langle o_i, x_i \rangle$ is drawn from $\mathcal{O}$ and $\mathcal{B}$.
The agent produces a policy $\pi$ represented as executable source code composed from reusable robot skills. Following the graph-based policy representation of GaP~\citep{chen2026gap}, we view $\pi$ as a computation graph whose nodes invoke skills and whose edges define their execution and control flow. The policy is constructed over a skill library
$\mathcal{S} = \mathcal{S}_0 \cup \mathcal{S}_g \cup \mathcal{S}_{\tau} \cup \mathcal{S}_A$,
where $\mathcal{S}_0$ contains basic robot and sensor primitives;
$\mathcal{S}_g$ contains reusable general-purpose skills;
$\mathcal{S}_{\tau}$ contains task-specific skills provided by a human expert; and
$\mathcal{S}_A$ contains task-specific skills synthesized by the agent.
Given $\mathcal{S}$, an LLM agent $A$ constructs $\pi$ by selecting, parameterizing, and composing available skills, while implementing new skills when the existing library is insufficient.

On a task instance $\tau_i$, the resulting policy
$\pi(a \mid \mathcal{I})$ maps observations $\mathcal{I}$ to robot actions $a$ in closed loop, and the resulting episode is evaluated by $\mathcal{J}$.
Given a task class $\mathcal{T}$ and skill library $\mathcal{S}$, the agent seeks to construct a policy that succeeds across the belief space:
$
\pi^* = \arg\max_{\pi}\;
\mathbb{E}_{x_i \sim \mathcal{B}}
\left[\mathcal{J}\big(\pi(a \mid \mathcal{I})\big)\right],
$
using only $\mathcal{L}$, $\mathcal{S}$, and the outcomes of its own policies.
Policy improvement may therefore modify both the composition of the policy graph and the implementation of its constituent skills.

\paragraph{Self-learning loop.}
Unlike reinforcement learning, feedback in the self-learning loop is not restricted to a scalar reward. At iteration $k$, the current policy $\pi_k$ runs in simulation on the training instances $\{\tau_i\}$. A feedback function $F$ turns these runs into a textual (and optionally visual) report $f_k = F(\pi_k, \{\tau_i\})$. The agent then edits its own policy, conditioned on the report and on its edit history $\pi_{k+1} = A\big(\pi_k,\, f_k,\, \{(\pi_j, f_j)\}_{j<k}\big)$.
The loop runs for $K$ iterations. 
Success is reported on a held-out set of instances for the initial policy $\pi_0$ (zero-shot, $k{=}0$) and for the iterate with the best validation score, $\pi_{k^*}$ with $k^* = \arg\max_k \mathrm{val}(\pi_k)$. Held-out success at every iteration gives a learning curve for each setting and task. Each study below varies one component of this loop: the agent $A$, the skills $\mathcal{S}$, the feedback $F$, or the curriculum over training instances.


\paragraph{Assumptions and Scope.}

We assume access to a reliable physics simulator. Training and evaluation remain within the same manipulation task class, with a fixed robot and sensor configuration and bounded scene variation. We evaluate generalization to held-out instances of these task classes and do not study sim-to-real transfer or generalization to new embodiments.

\section{Study Design}

\begin{figure}[t]
  \centering
  \newlength{\taskpw}\setlength{\taskpw}{\dimexpr(\textwidth-29pt)/7\relax}
  \setlength{\lineskip}{0pt}%
  \newcommand{\taskpanel}[3]{
    \begin{tikzpicture}[inner sep=0pt, baseline=(p.south)]
      \node[anchor=south west] (p) at (0,0) {\includegraphics[width=\taskpw]{#1}};
      \if\relax\detokenize{#2}\relax\else
        \node[overlay, anchor=north west, fill=white, fill opacity=0.85, text opacity=1, rounded corners=1pt, inner sep=1.1pt,
              font=\fontsize{5.5}{6}\selectfont] at ([shift={(1.2pt,-1.2pt)}]p.north west) {#2};
      \fi
      \if\relax\detokenize{#3}\relax\else
        \node[overlay, anchor=south west, fill=white, fill opacity=0.85, text opacity=1, rounded corners=1pt, inner sep=1.1pt,
              text={rgb,255:red,214;green,96;blue,0}, font=\fontsize{5.5}{6}\selectfont\bfseries] at ([shift={(1.2pt,1.2pt)}]p.south west) {#3};
      \fi
    \end{tikzpicture}}%
  \newlength{\taskph}\setlength{\taskph}{0.75\taskpw}
  \newcommand{\tasksep}[1]{\hspace{1.8pt}\smash{\textcolor{black!25}{\rule[-#1]{0.4pt}{\dimexpr\taskph+#1\relax}}}\hspace{1.8pt}}%
  \newcommand{\benchsep}[1]{\hspace{3.6pt}\smash{\textcolor{black!75}{\rule[-#1]{0.8pt}{\dimexpr\taskph+#1\relax}}}\hspace{3.6pt}}%
  \makebox[\dimexpr5\taskpw+16pt]{\scriptsize VABAR}\hspace{8pt}\makebox[\dimexpr2\taskpw+4pt]{\scriptsize MimicGen}\\[1pt]
  \taskpanel{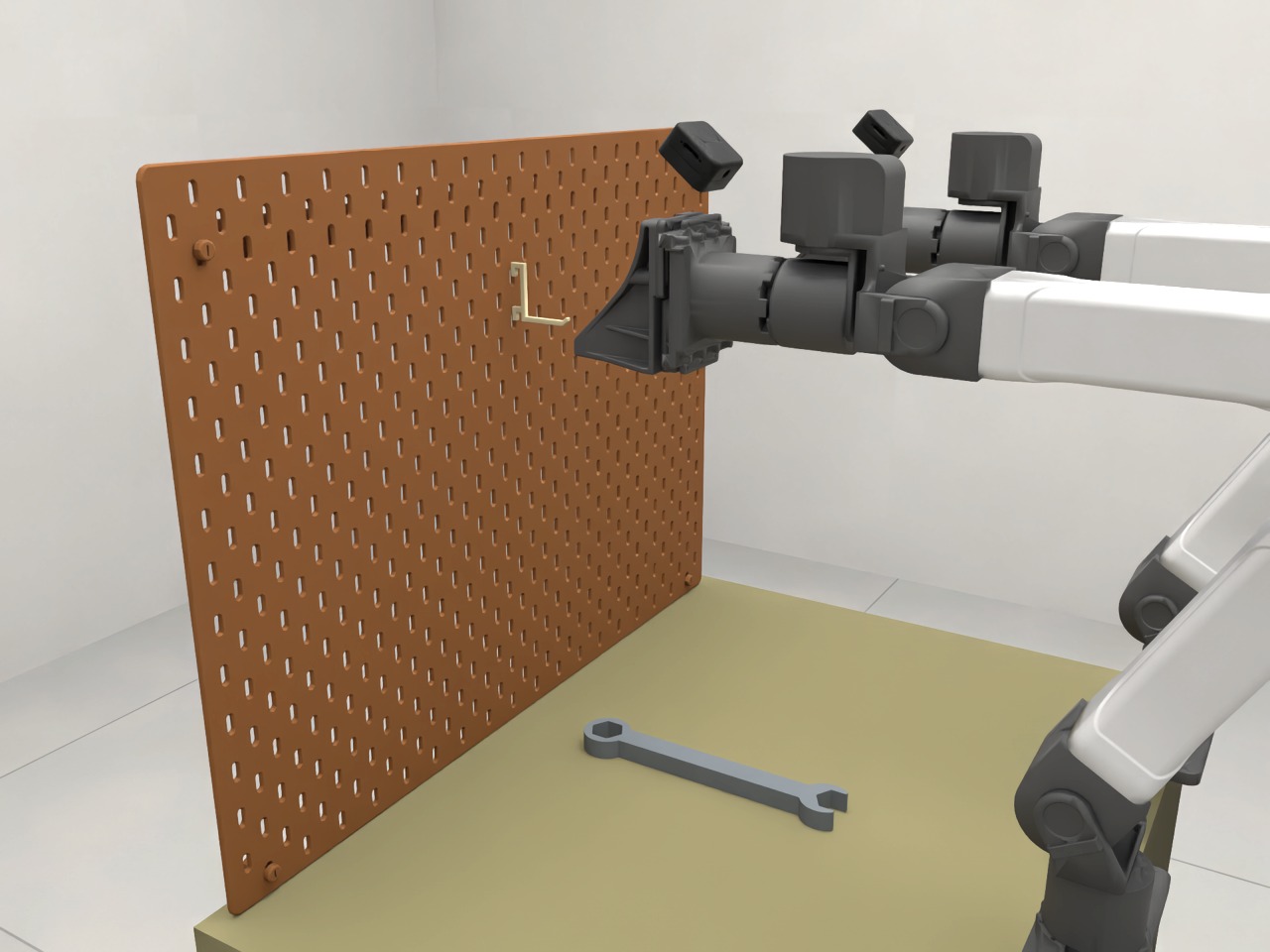}{Tool Hanging}{Before}\tasksep{0.5pt}%
  \taskpanel{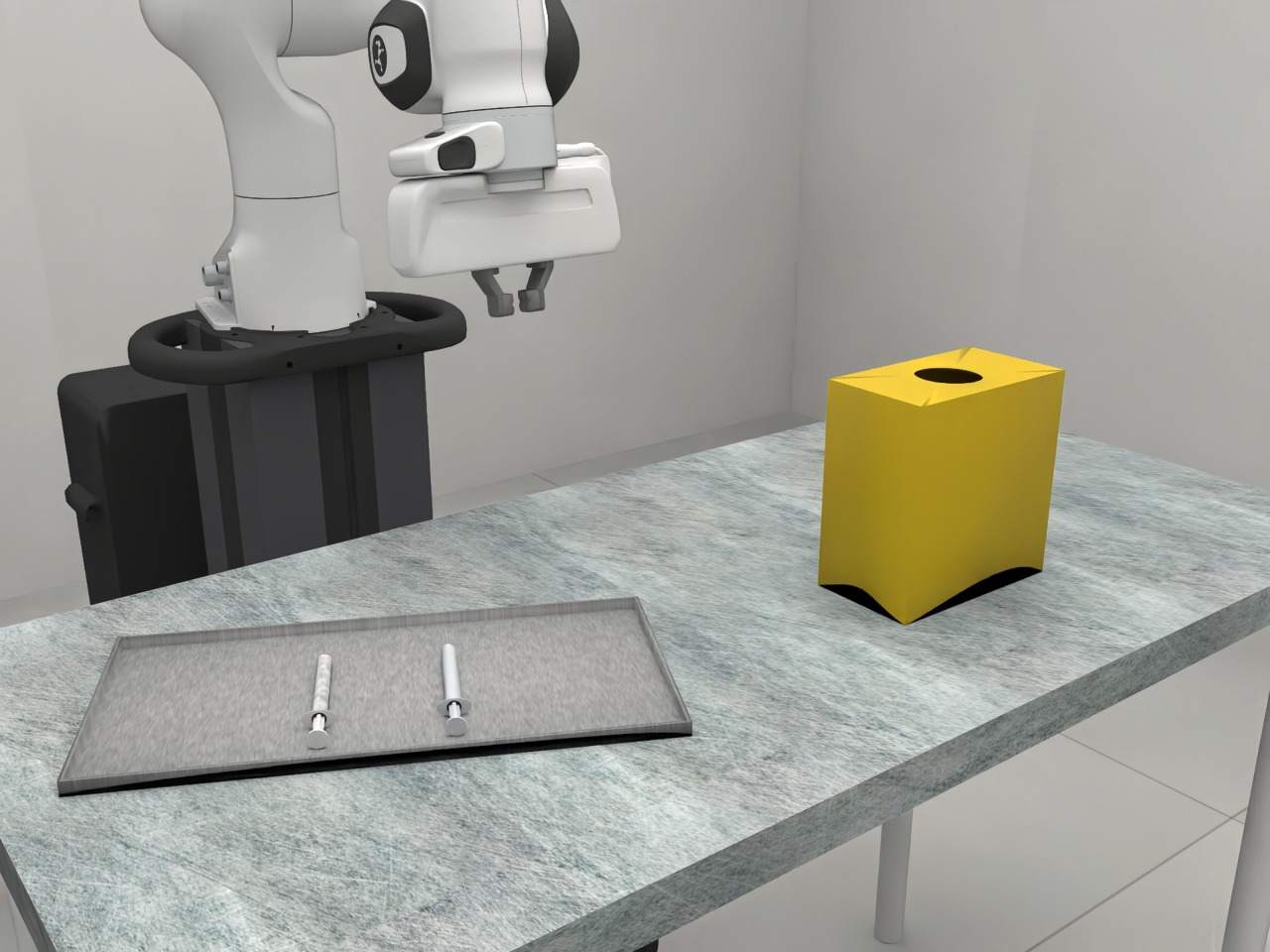}{Syringe Packing}{Before}\tasksep{0.5pt}%
  \taskpanel{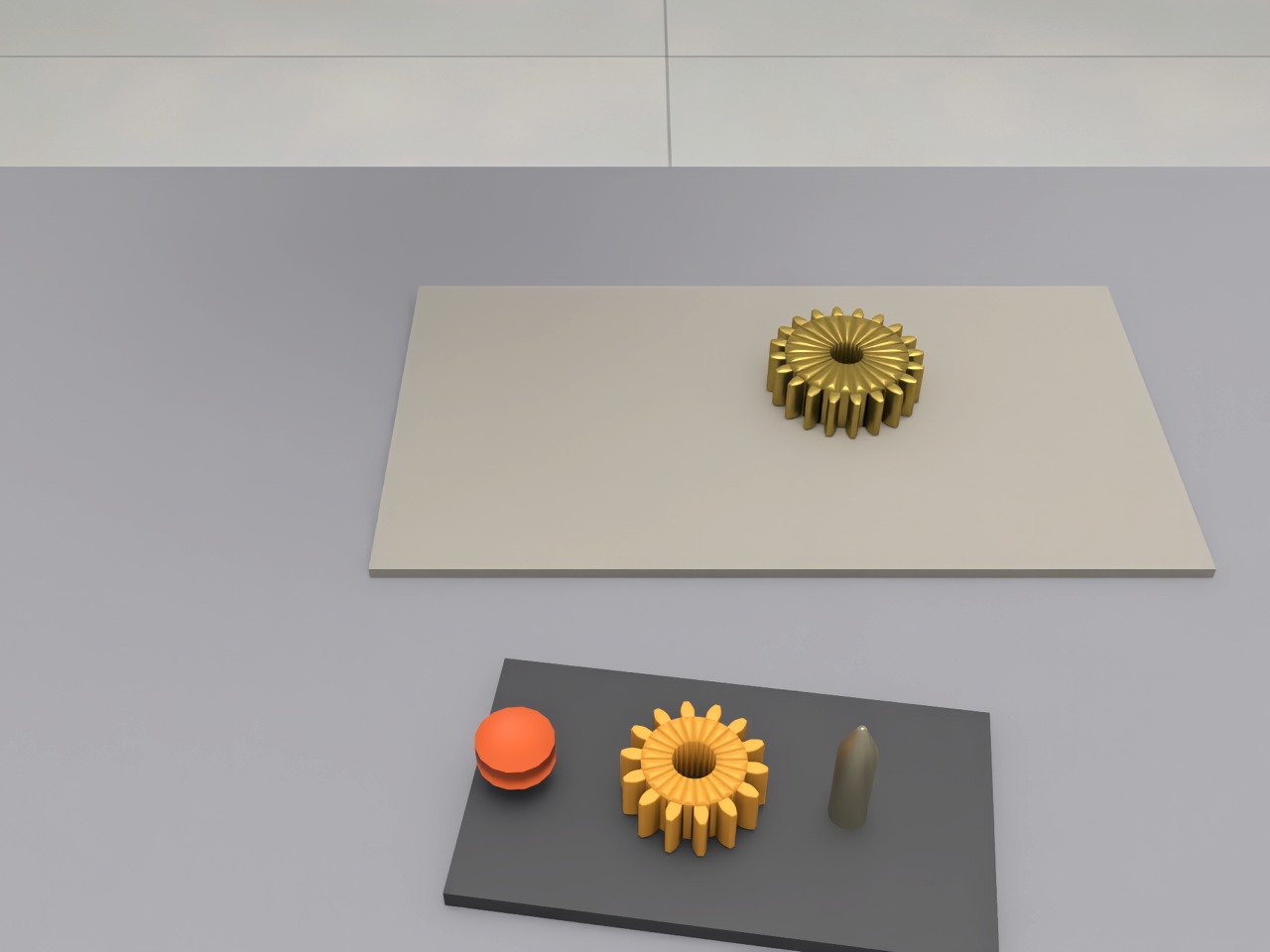}{Gear Meshing}{Before}\tasksep{0.5pt}%
  \taskpanel{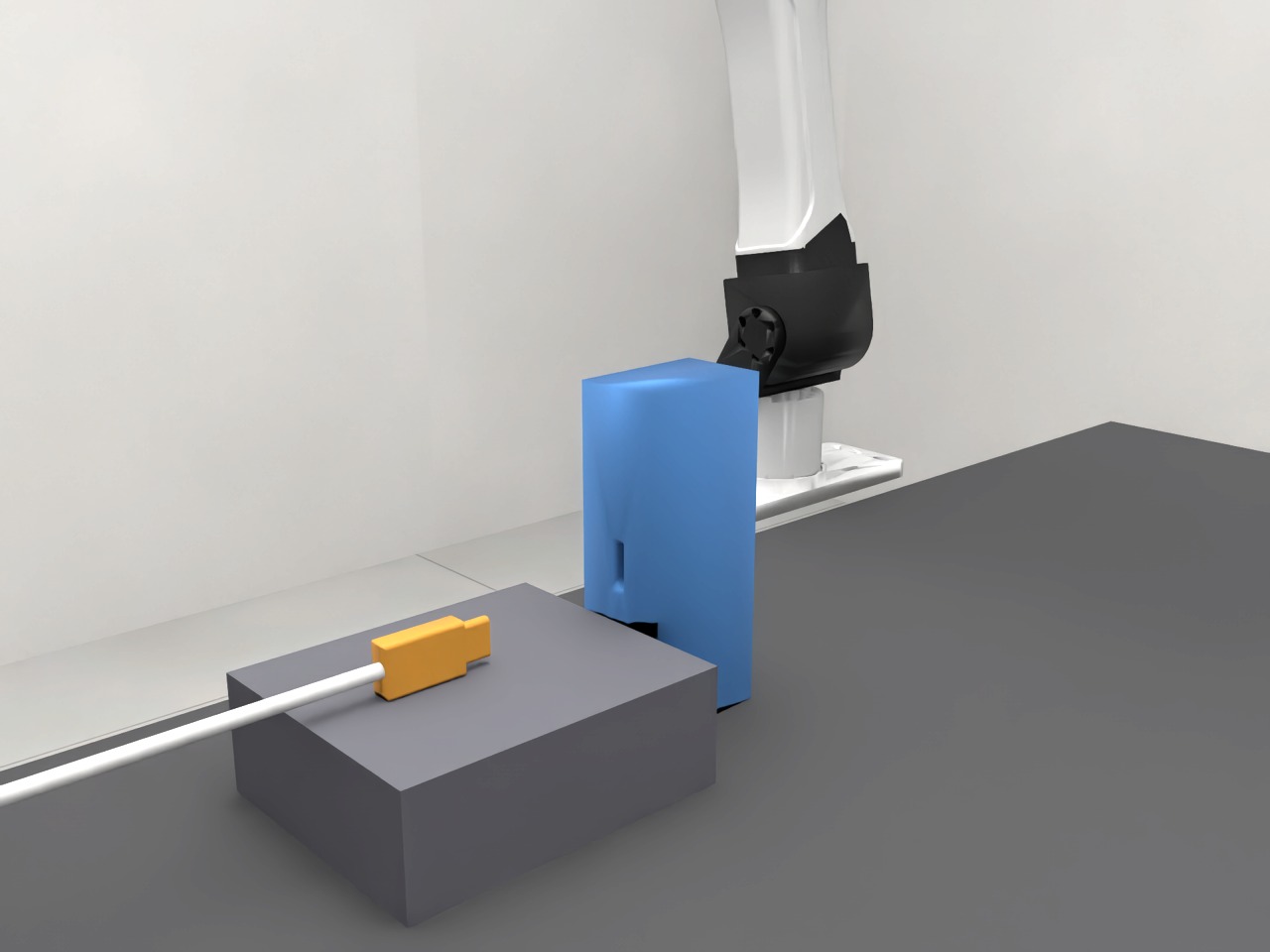}{USB-C Insertion}{Before}\tasksep{0.5pt}%
  \taskpanel{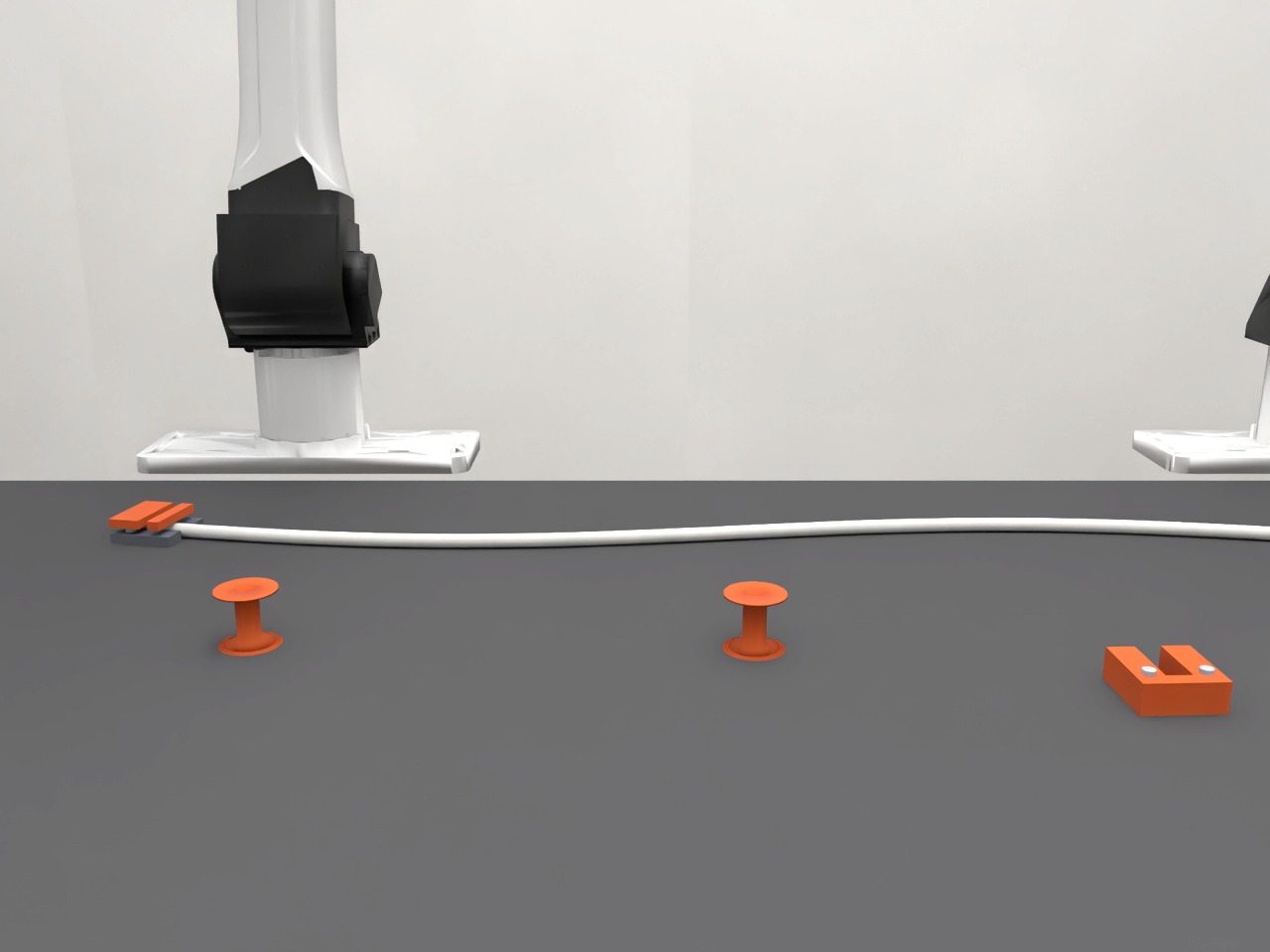}{Cable Routing}{Before}\benchsep{0.5pt}%
  \taskpanel{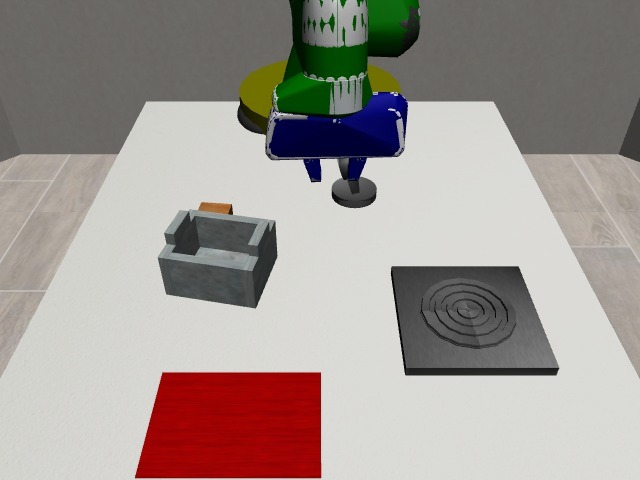}{Kitchen}{}\tasksep{0.5pt}%
  \taskpanel{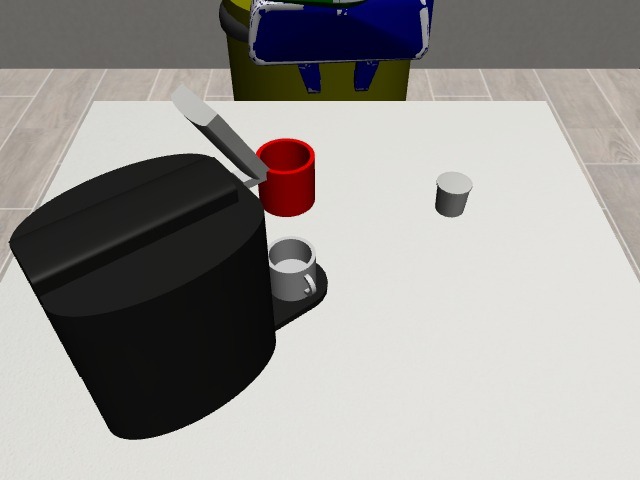}{Coffee}{}\\[0.5pt]
  \taskpanel{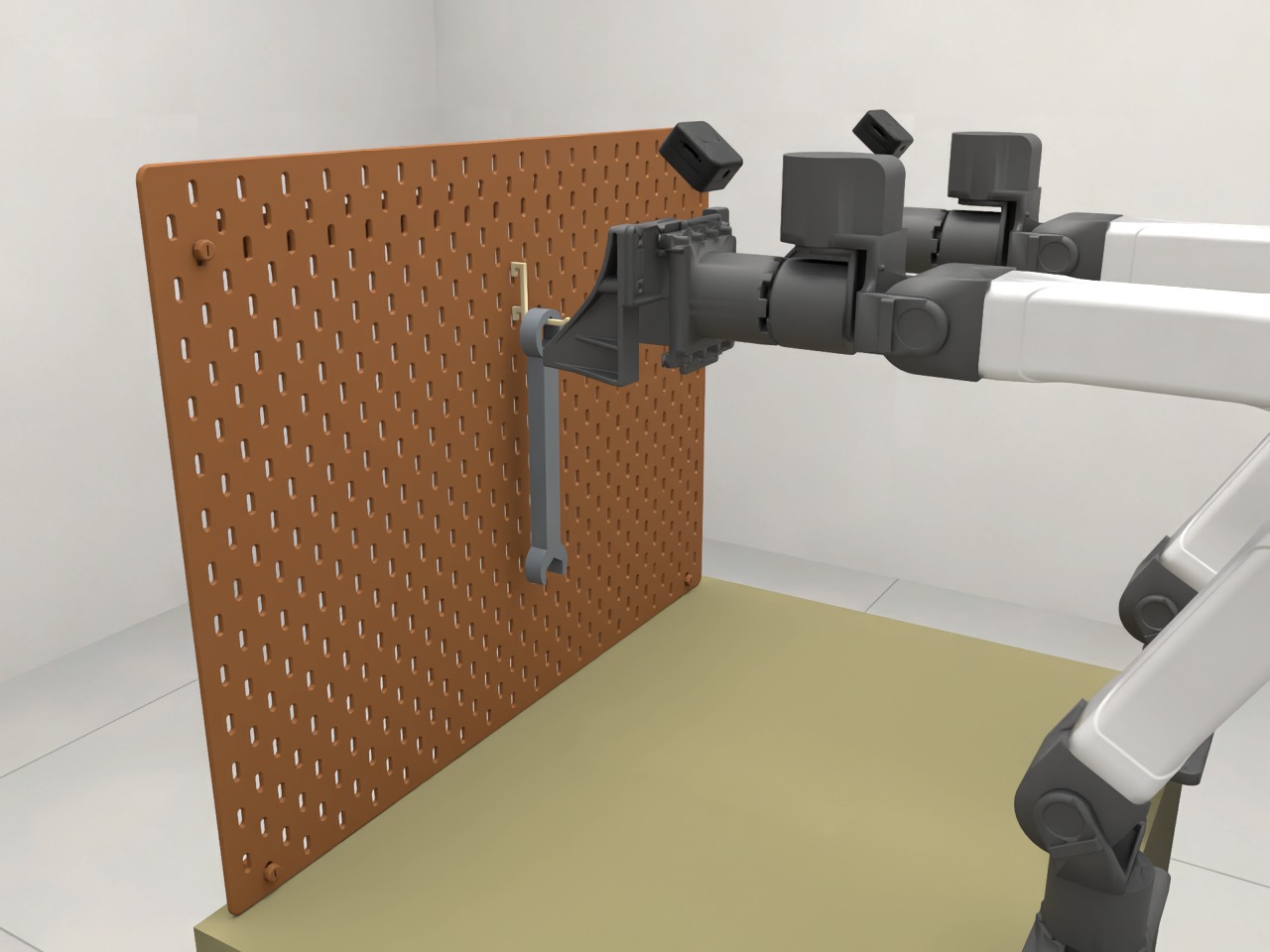}{}{After}\tasksep{0pt}%
  \taskpanel{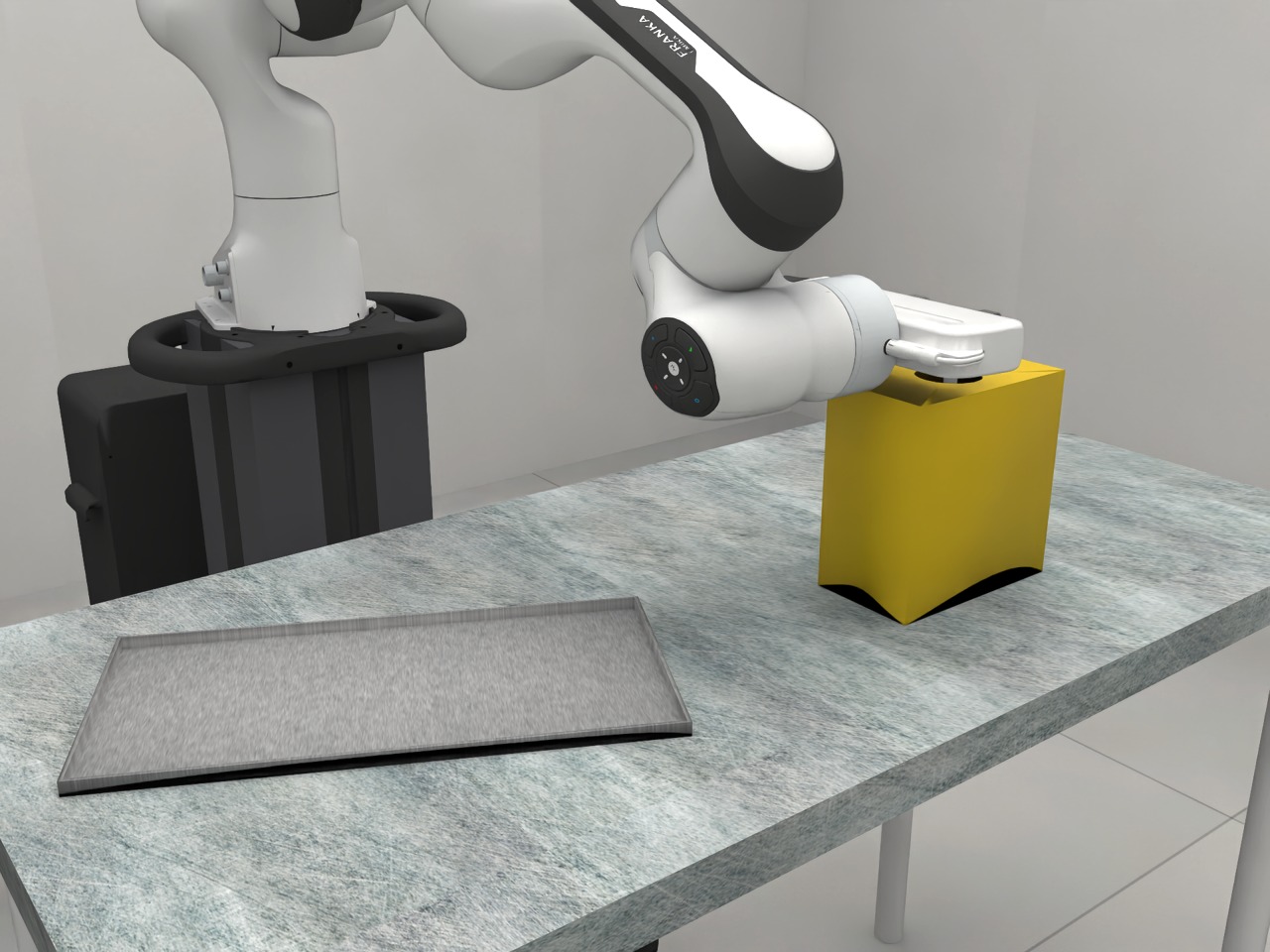}{}{After}\tasksep{0pt}%
  \taskpanel{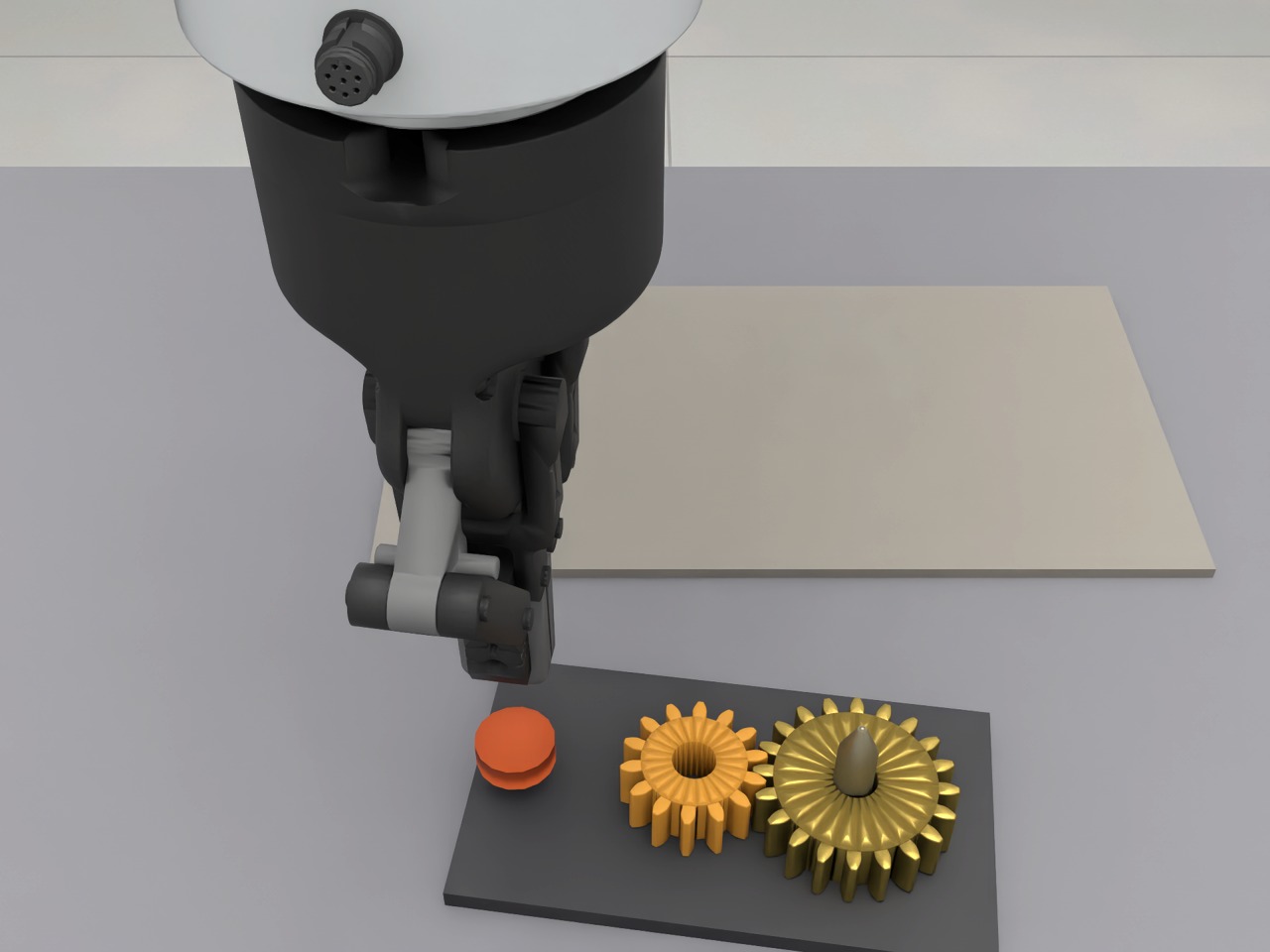}{}{After}\tasksep{0pt}%
  \taskpanel{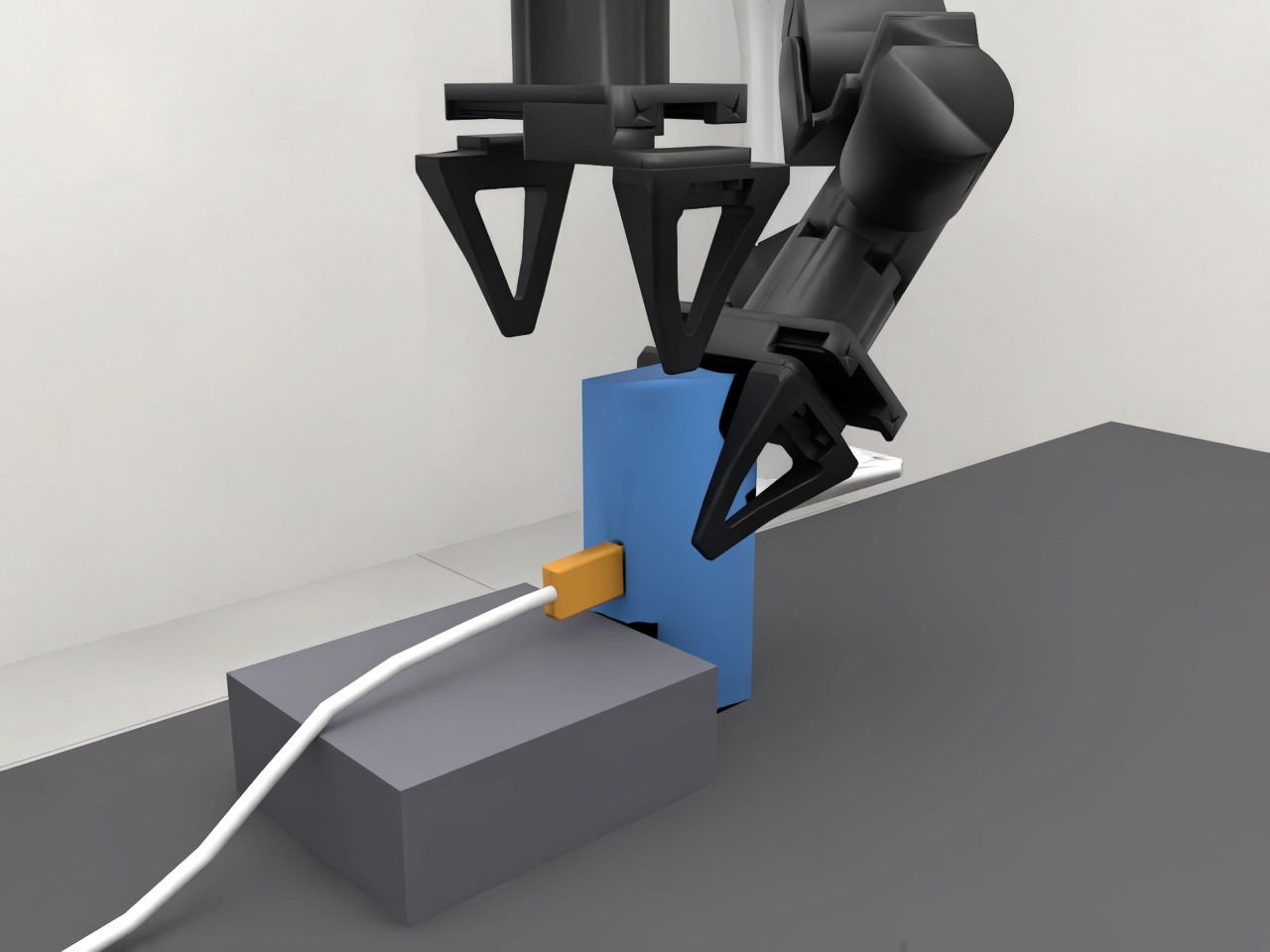}{}{After}\tasksep{0pt}%
  \taskpanel{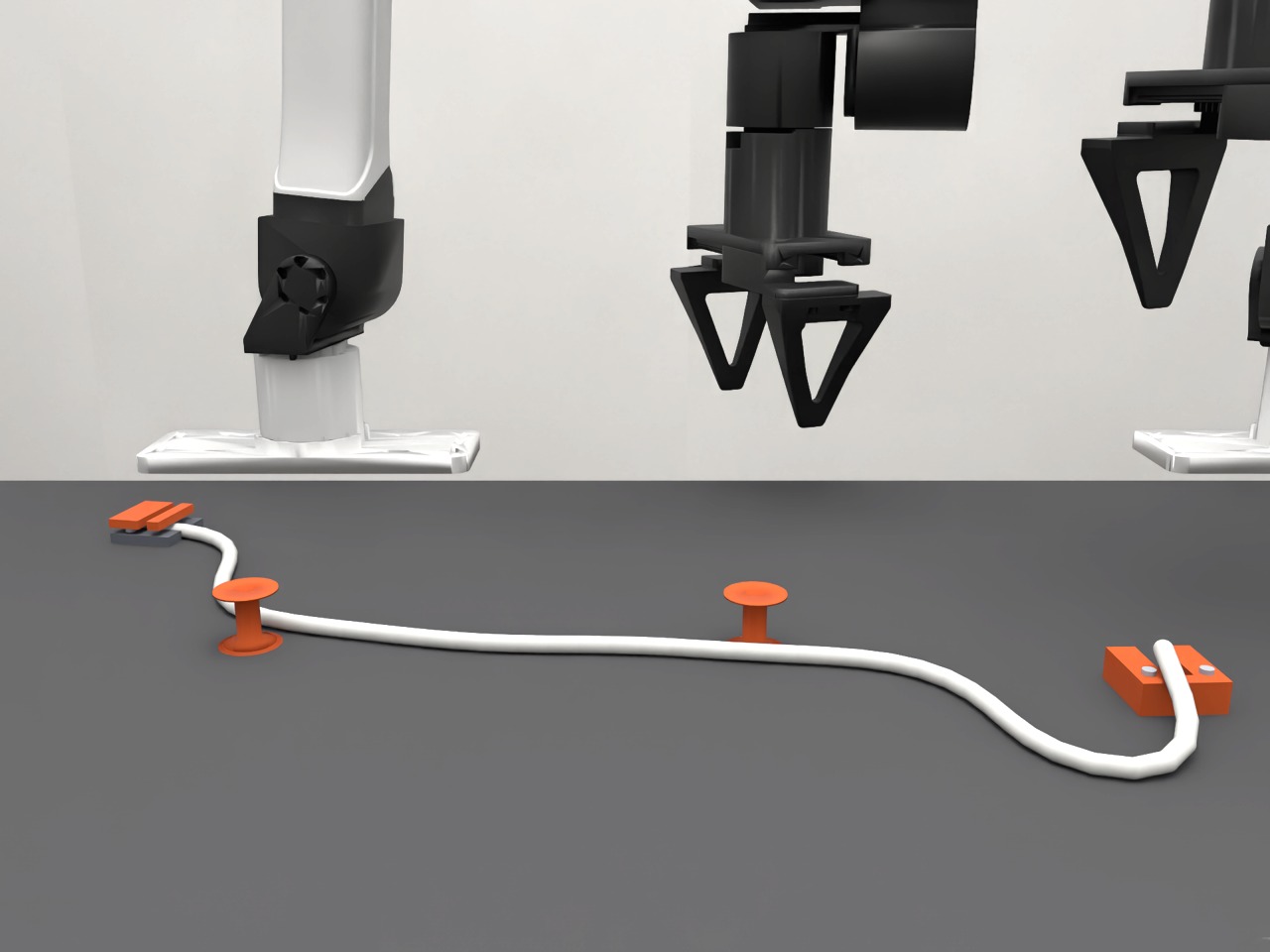}{}{After}\benchsep{0pt}%
  \taskpanel{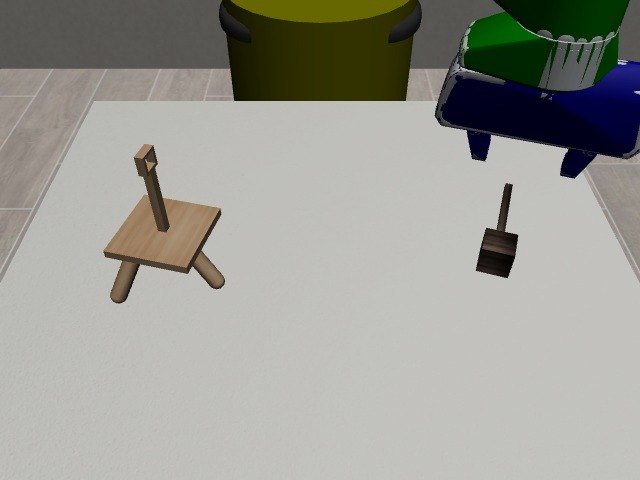}{Threading}{}\tasksep{0pt}%
  \taskpanel{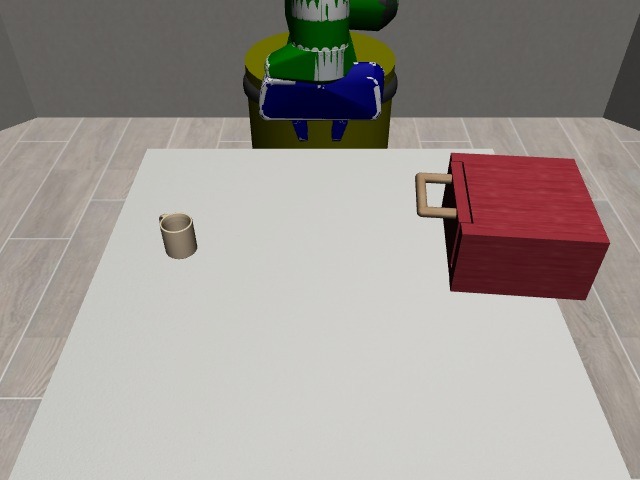}{Mug Cleanup}{}%
  \caption{Five VABAR tasks (Left) and four MimicGen tasks  (Right) used for this study.}
  \label{fig:tasks}\label{fig:mimicgen_tasks}
  \vspace{-10pt}
\end{figure}
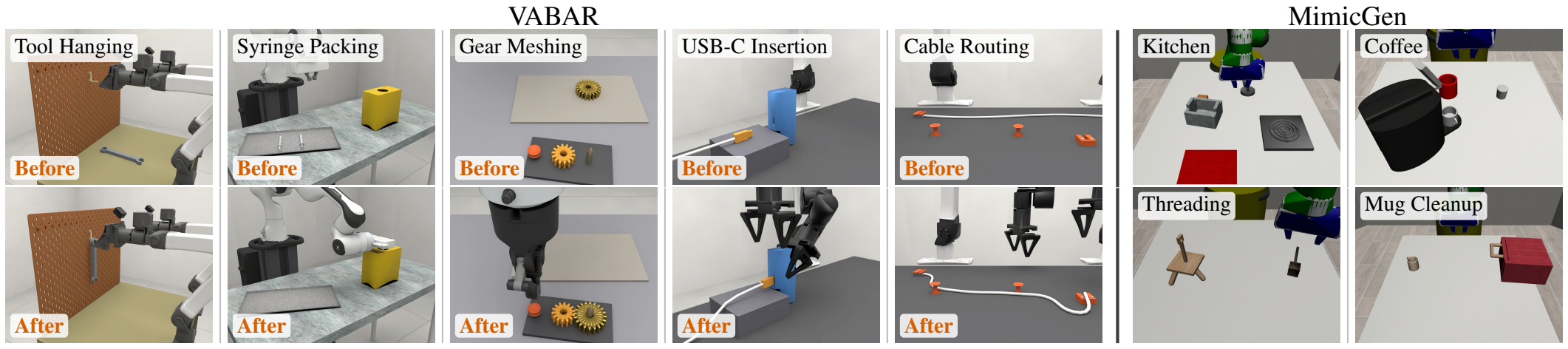

\paragraph{Task Setup.}
We evaluate five simulated manipulation tasks from the {Variational Automation Benchmark for Agentic Robotics (VABAR)}~\cite{vabar}: 
(1)~\emph{Tool Hanging}, (2)~\emph{Syringe Packing}, (3)~\emph{Gear Meshing}, (4)~\emph{USB-C Insertion}, and (5)~\emph{Cable Routing}. For the curriculum study, we evaluate four multi-step tasks from MimicGen~\cite{mandlekar2023mimicgen} using the Medium (D1) layouts: (1)~\emph{Kitchen}, (2)~\emph{Coffee}, (3)~\emph{Threading}, and (4)~\emph{Mug Cleanup} (Figure~\ref{fig:tasks}).

\paragraph{Skill Setup.}
We use  {open-robot-skills} provided by ~\cite{chen2026gap} as  task-agnostic skill library $\mathcal{S}$. The library contains a set of 30 skills in the Agent Skills format, each with a natural-language \texttt{SKILL.md} describing when the skill applies and conditions. The library covers perception, grasping and motion planning primitives.
The full list of skills can be found in Appendix~\ref{app:skills}.

\textbf{Metrics and Evaluation scale}.
In evaluation, every agent setting runs the same closed-loop episodes on the same training, validation and testing task instances with repeated runs. We report the success rate of the hold-out testing task instance that the scene configuration is different from the training and testing task instances.
We run each of the ten agent settings on the five tasks with five independent runs per task with two rounds of feedback. In each round the current program runs on 4 training, 5 validation and 10 held-out instances. We use full spatial and temporal feedback and no curriculum for the skill composition study and spatial temporal skill expansion study. We use Opus~5 xhigh as default for the feedback and skill-library studies. Including pilot studies, we use 17,900 GPU-hours on NVIDIA RTX PRO 6000 Blackwell GPUs and 2,700 GPU-hours on NVIDIA A100 with in total 27,600 coding agent rollouts with an estimation of \$70,000 computing cost. The study takes 2,500 coding agents editing sessions with \$36,500 for LLM inference cost.


\section{Experiments}

\subsection{Agent Model}

\begin{wrapfigure}{r}{0.36\linewidth}
    \centering
    \includegraphics[width=\linewidth]
    {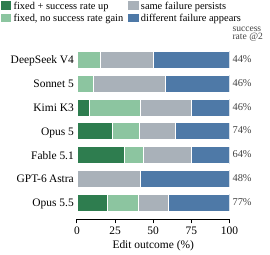}
    \caption{Outcomes of edits targeting a reported failing skill.}
    \label{fig:feedback_outcome}
\end{wrapfigure}
We study how the choice of coding model affects both initial policy generation
and subsequent self-learning.
Each agent is given the shared general skill library, and the task-specific expert skills provided with VABAR(~\cite{vabar}). The skill library is sufficiently expressive for successful task execution.
Each agent generates a zero-shot policy ($@0$) and then revises its code from
execution feedback for two rounds ($@1$ and $@2$).
We evaluate Claude Opus~5~\citep{anthropic2026opus5},
Claude Opus~5.5~\citep{anthropic2026opus55},
Claude Sonnet~5~\citep{anthropic2026sonnet5},
Claude Fable~5.1~\citep{anthropic2026fable51},
GPT-6 Astra~\citep{openai2026astra},
GPT-6 Sol~\citep{openai2026sol},
DeepSeek~V4~\citep{deepseek2026v4},
and Kimi~K3~\citep{kimiteam2026k3}.
Table~\ref{tab:models} reports held-out success after each round, and
Figure~\ref{fig:zs_composition} analyzes how models construct their initial policies from provided and  generated skills.

\begin{table}[t]
  \centering
  \setlength{\tabcolsep}{3.5pt}
  \setlength{\aboverulesep}{0pt}
  \setlength{\belowrulesep}{0pt}
  \renewcommand{\arraystretch}{1.15}
  \resizebox{\linewidth}{!}{%
  \begin{tabular}{l|ccc|cc|ccc|ccc|ccc|ccc|ccc}
    \toprule
    Model & \multicolumn{3}{c|}{Overall} & \multicolumn{2}{c|}{$\Delta$ Overall} & \multicolumn{3}{c|}{Tool Hanging} & \multicolumn{3}{c|}{Syringe Packing} & \multicolumn{3}{c|}{Gear Meshing} & \multicolumn{3}{c|}{USB-C Insertion} & \multicolumn{3}{c}{Cable Routing} \\
    \cmidrule(lr){2-4} \cmidrule(lr){5-6} \cmidrule(lr){7-9} \cmidrule(lr){10-12} \cmidrule(lr){13-15} \cmidrule(lr){16-18} \cmidrule(lr){19-21}
    & @0 & @1 & @2 & 1 & 2 & @0 & @1 & @2 & @0 & @1 & @2 & @0 & @1 & @2 & @0 & @1 & @2 & @0 & @1 & @2 \\
    \midrule
    Human & \multicolumn{3}{c|}{82} & \multicolumn{2}{c|}{--} & \multicolumn{3}{c|}{87} & \multicolumn{3}{c|}{90} & \multicolumn{3}{c|}{100} & \multicolumn{3}{c|}{87} & \multicolumn{3}{c}{47} \\
    \midrule
    Opus 5, xhigh & 45 & 64 & 74 & $+19$ & $+10$ & 22 & 62 & 60 & 28 & 38 & 35 & 100 & 100 & 100 & 0 & 50 & 100 & 73 & 72 & 73 \\
    Opus 5, high & 41 & 62 & \textbf{81} & $\mathbf{+21}$ & $\mathbf{+19}$ & 25 & \textbf{72} & \textbf{97} & 40 & 30 & 45 & 67 & 100 & 100 & 0 & 33 & 87 & 75 & 76 & 76 \\
    Opus 5, low & 57 & 67 & 74 & $+10$ & $+7$ & 0 & 30 & 57 & \textbf{51} & 50 & \textbf{60} & 100 & 100 & 100 & 60 & 80 & 80 & 73 & 76 & 76 \\
    \midrule
    DeepSeek V4 & 38 & 43 & 44 & $+5$ & $+1$ & 0 & 0 & 0 & 20 & 29 & 30 & 100 & 100 & 100 & 0 & 10 & 12 & 70 & 77 & 77 \\
    Sonnet 5 & 45 & 46 & 46 & $+1$ & $0$ & 0 & 0 & 0 & 32 & 34 & 34 & 100 & 100 & 100 & 25 & 25 & 25 & 67 & 72 & 72 \\
    Kimi K3 & 39 & 43 & 46 & $+4$ & $+3$ & 0 & 0 & 2 & 30 & 47 & 55 & 97 & 97 & 97 & 0 & 0 & 0 & 69 & 71 & 73 \\
    Fable 5.1 & 44 & 57 & 64 & $+13$ & $+7$ & 24 & 32 & 32 & 49 & 49 & 56 & 67 & 100 & 100 & 0 & 25 & 50 & 79 & 79 & 79 \\
    GPT-6 Astra & 43 & 47 & 48 & $+4$ & $+1$ & 0 & 0 & 0 & 20 & 25 & 30 & 100 & 100 & 100 & 15 & 30 & 30 & 80 & 80 & 80 \\
    Opus 5.5 & \textbf{67} & \textbf{74} & 77 & $+7$ & $+3$ & 25 & 40 & 55 & 46 & 50 & 50 & 100 & 100 & 100 & \textbf{85} & \textbf{100} & 100 & 79 & 79 & 79 \\
    GPT-6 Sol & 33 & 41 & 53 & $+8$ & $+12$ & 0 & 0 & 0 & 30 & 30 & 30 & 50 & 50 & 50 & 0 & 40 & 100 & \textbf{83} & \textbf{83} & \textbf{83} \\
    \bottomrule
  \end{tabular}}
  \caption{Self-learning on VABAR by model. Held-out per-object success (\%) reached within 0, 1, and 2 rounds of feedback (@$k$) reported by mean over the runs scored at each round. $\Delta_k$ denotes the gain in overall success from the $k$-th round of feedback, i.e.\ @$k$ $-$ @$(k{-}1)$.}
  \label{tab:models}
  \vspace{-10pt}
\end{table}

Table~\ref{tab:models} shows zero-shot performance is not sufficient to evaluate if the model can self-improve. Several models begin from nearly identical initial success rates but diverge after receiving the same execution feedback. For example, Opus~5 xhigh, Fable~5.1, GPT-6 Astra, and Sonnet~5 start at 43--45\% overall success, yet after two rounds Opus~5 xhigh and Fable~5.1 improve to 74\% and 64\%, respectively, while GPT-6 Astra and Sonnet~5 remain at 48\% and 46\%. The same distinction appears across individual tasks: models that initially fail Tool Hanging or USB-C Insertion can sometimes repair these policies substantially, whereas others remain near their initial performance despite observing the same failures. Opus~5.5 is strong at both stages, increasing from 67\% zero-shot to 77\% after feedback. These results suggest that generating a plausible initial manipulation policy and using execution evidence to diagnose and repair that policy are distinct model capabilities (Figure \ref{fig:feedback_outcome}).

\medskip
\noindent
\fbox{%
\parbox{\dimexpr\linewidth-2\fboxsep-2\fboxrule\relax}{%
\textbf{Finding 1.} Differences in model capability are less apparent at policy initialization but become pronounced during self-learning. Models with similar zero-shot performance can diverge substantially after receiving the same execution feedback, suggesting that policy generation and policy repair are distinct capabilities.}}

\subsection{Skill}

\begin{wrapfigure}{r}{0.4\linewidth}
    \centering
    \includegraphics[width=\linewidth]
    {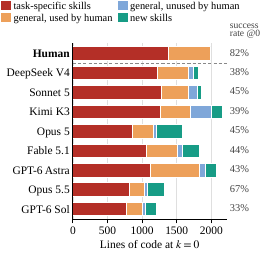}
    \caption{Zero-shot policy composition and attribution by code source.}
    \label{fig:zs_composition}
\end{wrapfigure}

We next study how agents use the skills they are given. The skill library provides an important prior: the expert skills encode task-specific manipulation strategies, while the general library provides
reusable perception, grasping, and motion-planning routines. Such a library could help the agent by providing useful building blocks, or
could instead constrain it to behaviors already represented in the library. To distinguish these cases, we trace the source of each line in the zero-shot policy ($@0$). After normalizing each line, we match it verbatim against the expert task-specific skills, the general-library skills also used by the human policy, and the remaining general-library skills; unmatched lines are attributed to
new code written by the agent.

Figure~\ref{fig:zs_composition} reports the mean number of lines from each source, averaged over runs per task and then over tasks, and relates it to success after feedback in Table~\ref{tab:models}.
As shown in Figure~\ref{fig:zs_composition}, models draw on the same skills in different ways.
Models that improve most with feedback also write a larger share of their policy as new task-specific code: new code makes up 24\% of the lines for Opus~5 xhigh, 18\% for Opus~5.5, and 12\% for Fable~5.1.
They use the expert skills as a basis but add or modify task-specific logic where the provided implementations are insufficient.
In contrast, Sonnet~5, DeepSeek~V4, Kimi~K3, and GPT-6 Astra write only 4--7\% new code and rely more heavily on behavior already represented in the expert and general skills.
These models also improve the least, ending at 39--46\% overall success after two rounds.
Using more of the library is not the same as adapting beyond it: Kimi~K3 uses the largest share of general skills not used by the human policy (13\% of its lines), yet still writes little new code and gains only 7 points.
Since several of these models start from similar zero-shot success, the difference is not simply in the quality of the first policy, but in whether the agent can modify and extend the provided skills when the existing behavior is insufficient.



\medskip
\noindent
\fbox{%
\parbox{\dimexpr\linewidth-2\fboxsep-2\fboxrule\relax}{%
\textbf{Finding 2.}
Expert skills are a boundary for weaker models but a scaffold for stronger ones. Weaker models largely compose behaviors already represented in the skill library, while stronger models selectively reuse and depart from human-provided skills to create new task-specific approaches.
}}


\subsection{Feedback}

\begin{wrapfigure}{r}{0.4\linewidth}
\vspace{-8pt}
\centering
\includegraphics[width=\linewidth]{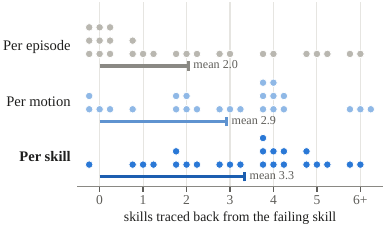}
\caption{Number of skills traced back from the observed failing skill before each post-failure policy edit. Finer-grained and semantic feedback leads agents to trace failures further upstream.}
\label{fig:failure_attribution_distance}
\vspace{-8pt}
\end{wrapfigure}

\textbf{Temporal Failure Attribution}.
We next study whether self-learning agents can attribute an observed robot
failure to the earlier policy decision that caused it.
A manipulation episode consists of multiple perception, planning, and
execution steps, so the point at which a failure becomes observable need not
be the point at which the underlying error was introduced.
As shown in Figure~\ref{fig:failure_attribution_distance}, the policy skill
selected for repair can occur throughout the preceding trajectory,
sometimes substantially before the point at which the episode finally halts.
We therefore ask whether exposing intermediate execution outcomes helps the
agent identify which earlier policy step should be revised.

\begin{table*}[t]
\centering
\caption{
Effect of temporal feedback structure on held-out success (\%).
\emph{Per motion} reports the outcome of each robot motion  that is not aligned with the policy's semantic skills, \emph{Per skill} reports the outcome of each semantic skills.
}
\label{tab:temporal_feedback}
\small
\resizebox{0.8\textwidth}{!}{
\begin{tabular}{lcccccc}
\toprule
Feedback
& Overall
& Tool Hanging
& Syringe Packing
& Gear Meshing
& USB-C Insertion
& Cable Routing \\
\midrule
Initial policy
& 3
& 0
& 0
& 0
& 0
& 17 \\
\midrule
Per episode
& 25
& 7
& \textbf{90}
& 10
& 0
& 17 \\
Per motion
& 36
& 45
& 55
& \textbf{80}
& 0
& 0 \\
Per skill
& \textbf{52}
& \textbf{90}
& 50
& 70
& \textbf{20}
& \textbf{27} \\
\bottomrule
\end{tabular}
}
\vspace{-20pt}
\end{table*}

Table~\ref{tab:temporal_feedback} shows that the localized
temporal feedback depends on where the underlying error occurs
relative to where the failure becomes observable.
Reporting every motion primitive can expose an earlier error once it
manifests as incorrect motion, reaching 45\% on Tool Hanging and 80\% on
Gear Meshing.
However, this finer-grained feedback does not expose non-motion operations such
as perception.
The useful unit of feedback is therefore the intermediate outcome associated
with a meaningful policy decision, rather than the finest available time
step.
In Tool Hanging, the per-episode agent repeatedly attributed failure to the
final reaching stage and modified that part of the policy, reaching only
7\% held-out success.
The observed miss, however, originated from an earlier incorrect estimate of
the wrench ring.
Skill-level feedback records the state changes and return values around each
semantic step, exposing this earlier error and increasing held-out success
to 90\%.
Gear Meshing shows the same pattern, improving from 10\% to 70\%.
Conversely, localized temporal feedback is less useful when the cause of
failure is already apparent in the final state.
In Syringe Packing, a syringe left outside or across the
container opening directly exposes the relevant geometric failure, and
per-episode feedback reaches 90\% held-out success.

\medskip
\noindent
\fbox{%
\parbox{\dimexpr\linewidth-2\fboxsep-2\fboxrule\relax}{%
\textbf{Finding 3.}
Feedback can be more useful when reported at the level of semantically meaningful policy skills, not simply at the finest temporal resolution. In multi-stage manipulation, the observed failure can occur well after the decision that caused it. Skill-level feedback exposes the intermediate outcomes of meaningful policy decisions and improves failure attribution, while finer motion-level feedback misses non-motion decisions such as perception.
}}

\paragraph{Spatial and Geometric Representation}

\begin{table*}[t]
\centering
\caption{
Effect of spatial representation on held-out success (\%).
World-frame and object-frame reports describe the same underlying scene state at different levels of relational organization, while part annotations
add task-relevant functional geometry.
}
\label{tab:spatial_feedback}
\small
\resizebox{0.9\textwidth}{!}{
\begin{tabular}{lcccccc}
\toprule
Spatial representation
& Overall
& Tool Hanging
& Syringe Packing
& Gear Meshing
& USB-C Insertion
& Cable Routing \\
\midrule
Outcome only
& 19
& 15
& 45
& 0
& 0
& 33 \\
\midrule
State + relations, world frame
& 30
& 25
& \textbf{60}
& 50
& 0
& 13 \\
State + relations, object frames
& 40
& {63}
& 50
& 60
& 0
& \textbf{27} \\
+ part annotations
& \textbf{52}
& \textbf{90}
& 50
& \textbf{70}
& \textbf{20}
& \textbf{27} \\
\bottomrule
\end{tabular}
}
\vspace{-10pt}
\end{table*}

\begin{wrapfigure}{r}{0.6\linewidth}
\centering
\includegraphics[width=\linewidth]{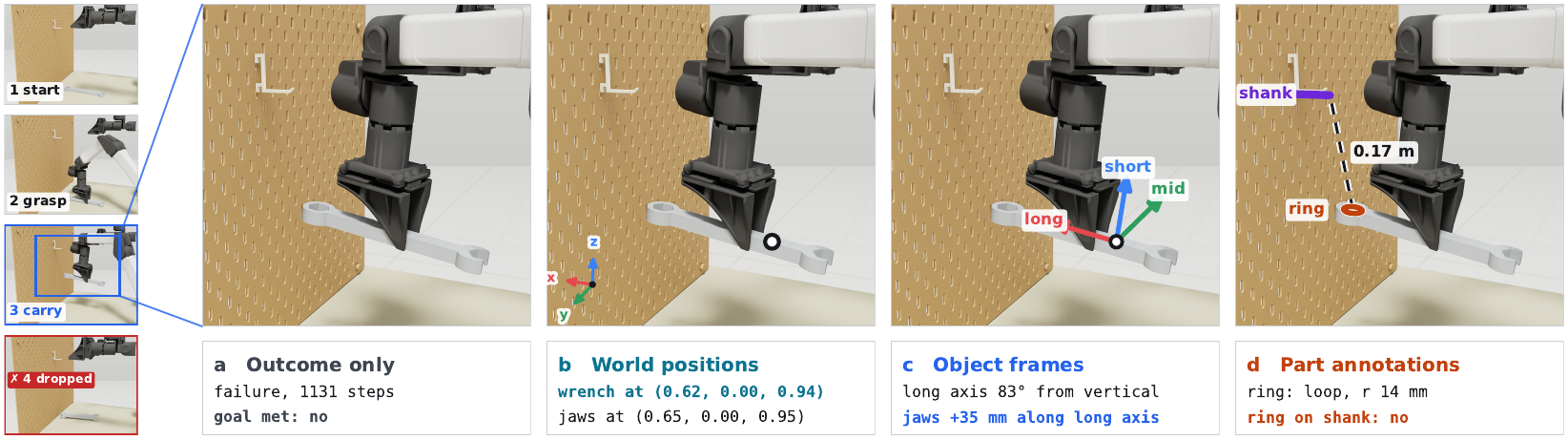}
\caption{ Spatial representations of the feedback report on one failed. The same execution state expressed at different levels of spatial
  organization.}
\label{fig:feedback_repr}
\end{wrapfigure}
We next study whether the representation of execution state affects an
agent's ability to reason about spatial failures.
A conventional world-frame representation describes where objects are, but leaves the agent to reconstruct the object-relative quantities needed for manipulation.
As suggests in Table~\ref{tab:spatial_feedback}, the world-frame state is sufficient in principle to recover the manipulation-relevant quantities, the agent often fails to reliably compose the required coordinate transformations and connect them to a policy decision.
Object-frame feedback makes these same relationships explicit, such as grasp offsets and relative orientations, and substantially improves performance on Tool Hanging and, to a lesser extent, Gear Meshing. This suggests that information completeness alone is insufficient, and agents can possess the necessary coordinates yet still fail to derive the spatial relationships needed for manipulation.
The same limitation extends beyond coordinate frames to object geometry itself. Providing accurate object state and geometry does not mean the agent understands how that geometry constrains action, because inferring   the affordance features of the wrench and hanger from whole-object poses is challenging for agents. Part annotations further improve feedback by exposing the functional structures that matter for interaction. Part-annotated
feedback reaches 52\% overall success, compared with 40\% for object frames and 30\% for world-frame state and relations.

\medskip
\noindent
\fbox{%
\parbox{\dimexpr\linewidth-2\fboxsep-2\fboxrule\relax}{%
\textbf{Finding 4.}
Recovering task-relevant spatial and geometric relationships from raw state information remains a bottleneck for agentic robotics. Although world-frame and object-frame feedback describe the same underlying scene, agents perform substantially better when object-relative geometry and functional spatial relationships are explicitly available, rather than requiring them to derive them.
}}

\paragraph{Feedback Reasoning.}

\begin{wrapfigure}{r}{0.34\linewidth}
\vspace{-8pt}
\centering
\includegraphics[width=\linewidth]
{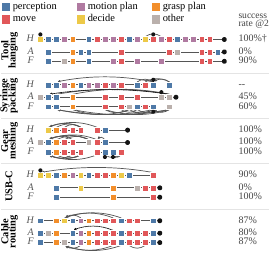}
\caption{
Policy structure by human(H), Astra(A), and
Fable(F). 
}
\label{fig:program_structure}
\vspace{-8pt}
\end{wrapfigure}
We next study how self-learning agents translate spatial understanding into executable robot motion. We compare how human-written and agent-learnt policies structure perception, motion planning, and recovery. Figure~\ref{fig:program_structure} shows that human policies frequently make geometric structure explicit through repeated perception and re-registration, intermediate waypoints, and dedicated motion-planning stages. These intermediate poses encode geometric priors such as approaching from a safer direction, maintaining clearance, or moving through configurations from which a feasible motion is easier to find.
In contrast, self-learning agents more often use verification, retry, and recovery around coarser manipulation steps.
They can reuse motions containing useful geometric structure, but are less
consistent at reconstructing why that structure is necessary when generating
or modifying the policy themselves.

Syringe Packing provides a concrete example.
The robot must carry a 150\,mm syringe through a 55\,mm opening while
reasoning about the syringe as part of the moving system, surrounding
obstacles, reorientation clearance, and insertion orientation
(Figure~\ref{fig:syringe_geometry}).
The human policy represents these dependencies explicitly with a collision
world, held-object registration, collision-aware waypoints, and
orientation-constrained insertion.
\begin{wrapfigure}{r}{0.3\linewidth}
\centering
\includegraphics[width=\linewidth]
{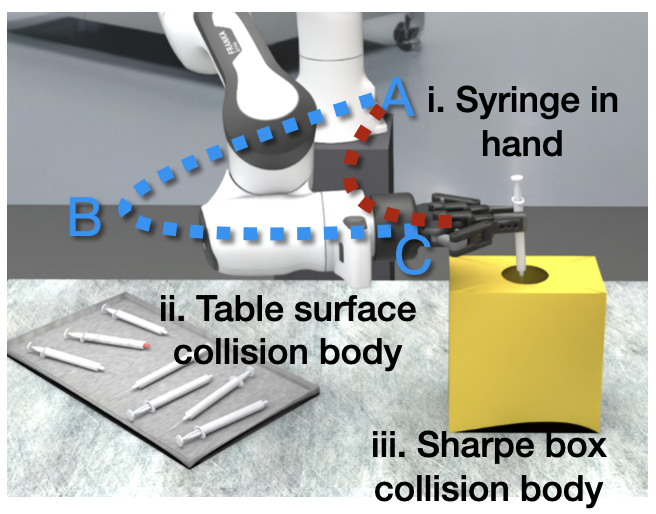}
\caption{
Human motion plan (blue) considers a collision free spot (B) to reorientate gripper, while agent plan (red) has limited room to reorientate.
}
\label{fig:syringe_geometry}
\vspace{-68pt}
\end{wrapfigure}
In contrast, none of the agent runs without expert skills independently
constructed an equivalent collision representation, and 17 of 19 initial
self-written policies carried or released the syringe flat over the opening.
Feedback often allowed agents to recognize the resulting errors and eventually
adopt heuristics such as reorienting above the lid and descending vertically,
but they still tended to encode these fixes as clearance constants, staging
heights, and retries rather than as a general collision-aware model.

\medskip
\noindent
\fbox{%
\parbox{\dimexpr\linewidth-2\fboxsep-2\fboxrule\relax}{%
\textbf{Finding 5.}
Reasoning over collision geometry and spatial relationships remains a key bottleneck for agentic robot motion planning. Agents can recognize local geometric errors, but struggle to integrate object pose, held-object geometry, obstacles, and clearance into a coherent collision-aware motion plan.

}}

\subsection{Curriculum Design}
\label{sec:curriculum-design}
\begin{figure}[t]
  \centering
  \includegraphics[width=0.9\linewidth]
  {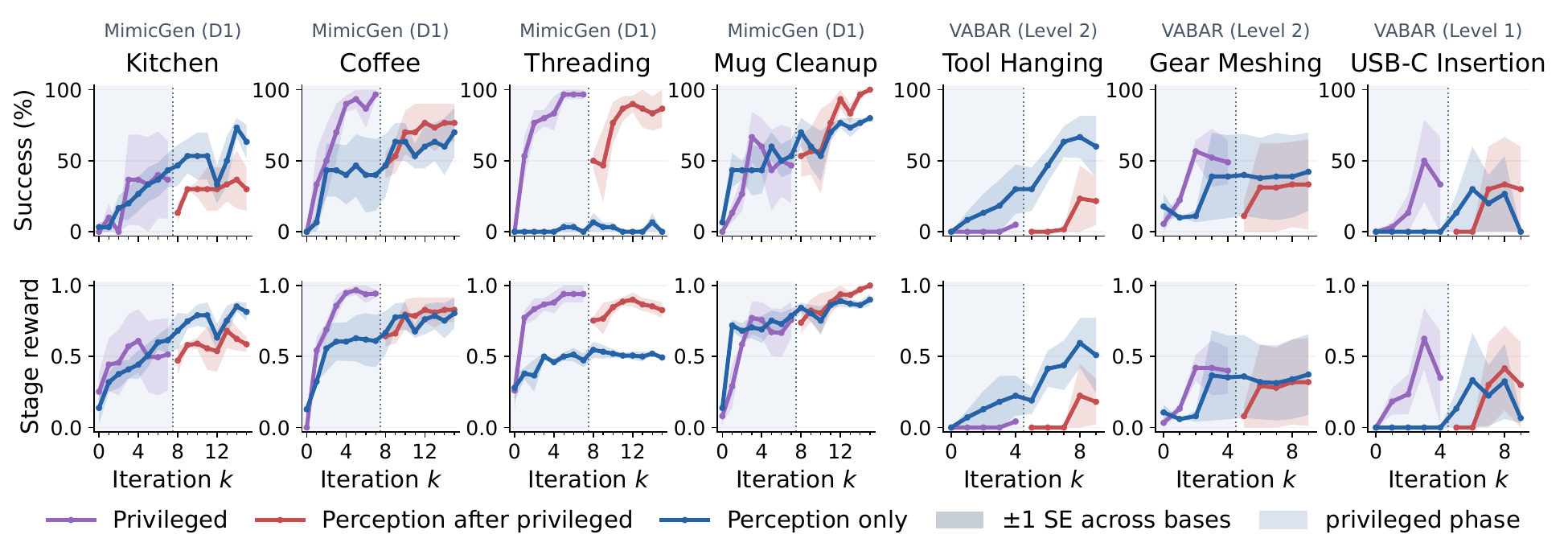}
  \caption{Held-out success (top) and stage reward (bottom) for the privileged-to-perception curriculum and perception-only control on MimicGen (left) and VABAR tasks (right), evaluated on 10 fixed held-out layouts per task. Lines show means over three seed replicates and bands show $\pm1$ standard error. The dotted line marks transfer to perception; shading indicates the privileged phase, during which only the curriculum experiment has access to privileged geometry.
  }
  \vspace{-10pt}
  \label{fig:mimicgen_pv}
\end{figure}

We study a representative \emph{privileged-to-perception curriculum}, a common design in reinforcement learning and used in prior works such as~\cite{pinto2017asymmetric,chen2020learning}.
As a focused case study of curriculum transfer in agentic robotics, the agent first learns with privileged simulation states with exact simulator geometry and is then transferred to non-privileged simulation states such as RGB-D perception.
We evaluate this curriculum on four MimicGen tasks using their Medium (D1) layouts with  8 privileged and 8 perception iterations and Claude Fable~5.1. We also use three VABAR tasks: Tool Hanging (Level 2), Gear Meshing (Level 2), and USB-C Insertion (Level 1) with 5 privileged iterations and 5 perception iterations with Claude Opus~5. Figure~\ref{fig:mimicgen_pv} reports results on four MimicGen and three VABAR tasks. Exact task configurations, training and held-out layouts, iteration budgets, probe settings, and other protocol details are provided in Appendix~\ref{sec:app-curriculum}.
\begin{wraptable}{r}{0.47\textwidth}
\centering
\small
\setlength{\tabcolsep}{3pt}
\resizebox{\linewidth}{!}{%
\begin{tabular}{lrrrr}
\toprule
& \multicolumn{2}{c}{\textbf{\# Skills}}
& \multicolumn{2}{c}{\textbf{Success (\%)}} \\
\cmidrule(lr){2-3}\cmidrule(lr){4-5}
Task
& w/ curr.
& w/o curr.
& w/ curr.
& w/o curr. \\
\midrule
Tool Hanging    & 19.3 & 28.0 & 16 & \textbf{63} \\
Gear Meshing    & 18.3 & 21.7 & 33 & \textbf{40} \\
USB-C Insertion & 18.0 & 10.7 & \textbf{31} & 16 \\
Kitchen         & 13.0 & 13.3 & 33 & \textbf{62} \\
Coffee          & 5.0  & 11.7 & \textbf{76} & 64 \\
Threading       & 8.7  & 18.3 & \textbf{86} & 2 \\
Mug Cleanup     & 6.3  & 9.3  & \textbf{93} & 77 \\
\midrule
Overall         & 12.7 & 16.1 & \textbf{52} & 46 \\
\bottomrule
\end{tabular}%
}
\caption{
Policy size (\# Skills, final policy) and held-out success (mean over the last three iterations) with (w/ curr.) and without (w/o curr.) the privileged-to-perception curriculum; mean over three runs.
}
\label{tab:curriculum-comparison}
\vspace{-1em}
\end{wraptable}

Figure~\ref{fig:mimicgen_pv} suggests that privileged-to-perception learning does not uniformly improve over learning directly from perception. Privileged information removes perception uncertainty during early learning, but, as suggested by Findings 4 and 5, accurate geometry alone does not guarantee that the agent discovers the right spatial representation or manipulation strategy.
The privileged phase can learn useful task structure, as in Threading, Coffee, and Mug Cleanup, but it can also sometimes commit early to brittle or non-generalizable choices.
For example, Kitchen develops a grasp strategy that depends on geometric precision unavailable after transfer, while Tool Hanging remains below 10\% held-out success during privileged learning despite access to exact wrench geometry. Thus, privileged information can simplify the learning problem without necessarily producing a policy that remains appropriate under perception. We further visualize and analyze representative failure cases in Appendix~\ref{sec:curriculum-transfer}.

In addition, Table~\ref{tab:curriculum-comparison} further suggests that once a policy structure is established under privileged
state, subsequent perception-based learning tends to preserve and patch that structure rather than fundamentally reorganize it (which corresponds to Finding 1).
Across tasks, the privileged-to-perception policies change little in size after transfer, from 12.0 skills at the end of the privileged phase to 12.7 after perception learning on average, while the final perception-only policies contain 16.1 skills.
The resulting path dependence can be beneficial when the early structure captures a manipulation strategy that remains valid under vision, but this can be harmful when the structure assumes geometric precision unavailable from perception. In such cases, the curriculum does not merely provide an initialization; it can constrain subsequent self-learning.

\medskip
\noindent
\fbox{%
\parbox{\dimexpr\linewidth-2\fboxsep-2\fboxrule\relax}{%
\textbf{Finding 6.}
Privileged-to-perception curriculum learning can either accelerate or constrain subsequent perception-based learning. Privileged agents learn policies that rely on exact state, usually with less verification and recovery, and often retain this structure after transfer. This helps when the learned manipulation strategy remains valid, allowing perception learning to focus on perception errors, but hurts when the policy depends on state precision unavailable under perception or lacks mechanisms for handling perceptual uncertainty.
}}

\section{Discussion and Limitations}

\begin{wrapfigure}{r}{0.4\linewidth}
\centering
\includegraphics[width=\linewidth]{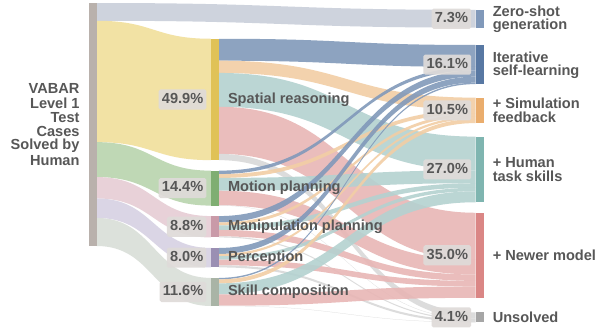}
\caption{
\textbf{What closes the gap to human-written policies?}
For VABAR Level-1 test cases solved by the human policy with Opus 5, improvements come progressively with the earlier findings,  while a small fraction unsolved.
}
\label{fig:syringe_geometry}
\vspace{-8pt}
\end{wrapfigure}
Our results return to the central question of this study: what determines whether an agent can interpret physical interaction and convert it into effective robot-policy improvement?
Figure~\ref{fig:syringe_geometry} provides a complementary view across the experiments.
Only 7.3\% of the human-solvable Level-1 test cases are solved by zero-shot policy generation alone, but the results suggest that improving agentic robotics requires the combination of iterative self-learning with richer simulation feedback, human task skills, and more powerful models.
At the same time, this decomposition shows that 4.1\% of test cases solved by a human-written policy under the same robot and simulation setup remain unsolved by the agents. Moreover, this study primarily considers Level-1 VABAR tasks with uncluttered environments and relatively simple motions.
The remaining gap is concentrated in agents' ability to interpret physical interaction into task-relevant reasoning and policy changes, particularly for spatial reasoning, manipulation planning, and collision-aware motion planning.


\textbf{Limitations.}
This study does not cover the full range of conditions that may affect agentic robot self-learning, such as prompting strategies and harness configurations. Given the large number of agent rollouts required for controlled evaluation, we leave a broader study of these factors to future work. Our model comparison is also necessarily a snapshot in time, as new agent models are released frequently; we plan to update the evaluation as new models become available. We focus on simulation to enable repeatable evaluation, leaving sim-to-real transfer and adaptation to real-world feedback as important next steps; real-to-sim toolkits such as RoboSimStudio~\citep{robosimstudio} may make it easier to construct matching simulations of real tasks for such studies. Finally, we do not consider human-in-the-loop self-learning, where human feedback may complement autonomous simulation feedback during policy improvement.

\bibliography{iclr2027_conference}
\bibliographystyle{iclr2027_conference}

\appendix
%

\newif\ifAppAspirational
\AppAspirationaltrue
\newcommand{\AppRunsPerCell}{five}
\newcommand{\AppTableOneRuns}{250}
\newcommand{\AppFeedbackRuns}{200}
\newcommand{\AppSkillRuns}{50}

\clearpage
\startcontents[appendix]
{\centering\Large\bfseries Appendix\par}
\vspace{0.5em}

\noindent The appendix documents what is needed to read, check and reproduce the study, in three parts.
\textbf{Experimental setup} (Appendices~\ref{app:tasks}--\ref{app:cost}) specifies the tasks, how they are scored and how
their instances are drawn; the skill library and what each experiment ships; the self-learning harness, its complete
prompts and feedback reports, and its revert, stopping and selection rules; and the models, their inference settings and
the compute and cost of the study.
\textbf{Additional results} (Appendix~\ref{app:failures}) locates where the programs fail.
\textbf{Curriculum study} (Appendix~\ref{sec:app-curriculum}) details the privileged-to-perception experiments of
Section~\ref{sec:curriculum-design}.

\vspace{0.4em}
\begin{multicols}{2}
{\small\raggedright\setlength{\parskip}{0pt}\printcontents[appendix]{}{0}{\setcounter{tocdepth}{2}}}
\end{multicols}

\clearpage
\renewcommand{\topfraction}{0.9}\renewcommand{\bottomfraction}{0.6}\renewcommand{\textfraction}{0.1}
\renewcommand{\floatpagefraction}{0.8}\setcounter{topnumber}{3}\setcounter{totalnumber}{5}
\makeatletter\setlength{\@fptop}{0pt}\makeatother

\appendixpart{Experimental Setup}
\section{Tasks, Scoring and Instances}
\label{app:tasks}

This appendix specifies the five VABAR tasks (Figure~\ref{fig:tasks}) at the level needed to reproduce a score:
what each task demands (Appendix~\ref{app:tasks_spec}), how an episode becomes per-object success, the metric of
every table (Appendix~\ref{app:tasks_scoring}), and how the 4 training, 5 validation and 10 held-out instances of
each task are drawn (Appendix~\ref{app:tasks_splits}). Appendix~\ref{app:tasks_difficulty} then relates held-out
success on each instance to the quantities it was drawn with.

\subsection{Task Specifications}
\label{app:tasks_spec}

The five tasks are the Easy (Level-1) tier of VABAR~\cite{vabar}. Every task instance is one RoboSimStudio task
document that, with its suite, fixes the robot, the objects, the start layout, a step cap, a goal predicate and a
reward (stated, or the share of goal clauses met). Table~\ref{tab:app_tasks} summarizes the five tasks.

\begin{table}[ht]
  \centering
  \scriptsize
  \setlength{\tabcolsep}{3pt}
  \renewcommand{\arraystretch}{1.12}
  \begin{tabularx}{\linewidth}{@{}>{\raggedright\arraybackslash}p{1.45cm}*{5}{>{\raggedright\arraybackslash}X}@{}}
    \toprule
    & Tool Hanging & Syringe Packing & Gear Meshing & USB-C Insertion & Cable Routing \\
    \midrule
    Instruction
      & \emph{hang the wrench on the hook}
      & \emph{pick up each syringe from the instrument tray, one at a time, and drop it through the hole in the top of the yellow sharps disposal box, until both are in the box}
      & \emph{put the gear on its peg so it meshes with the pinion, then press the red button at the near end of the board to start it}
      & \emph{insert the USB-C plug into the receptacle}
      & \emph{weave the cable through the two pegs and route its free end into the port} \\
    \addlinespace
    Robot
      & bimanual I2RT YAM; prompted with one arm
      & Franka Panda
      & Franka Panda with Robotiq 2F-85
      & bimanual I2RT YAM; prompted with one arm
      & bimanual I2RT YAM; prompted with both arms \\
    \addlinespace
    Objects
      & wrench (260\,mm); hook on a pegboard
      & two identical syringes 90\,mm apart in an instrument tray; sharps box with a top aperture
      & spur gear (16, 20 or 24 teeth) on a mat; board with a peg, a driven pinion and a start button
      & USB-C plug ($2\times$ scale) on an attached 240\,mm cable; receptacle bolted to the bench, 0.15\,mm clearance per side
      & 600\,mm cable, one end welded in an anchor bracket; two pegs; a port \\
    \addlinespace
    Success trials per episode
      & 1: the goal
      & 2: each syringe in the box
      & 1: the goal
      & 1: the goal
      & 3: seated at peg 1, at peg 2, in the port with its free end \\
    \addlinespace
    Partial-credit reward (floor)
      & share of the 3 goal clauses (1/3: a wrench at rest meets the speed clause)
      & share of the 2 syringes in the box (0)
      & turning fraction of its one gear, so 0 or 1 (0)
      & insertion depth / 10.4\,mm, capped at 1; 0 while the lateral offset exceeds 8.8\,mm (0)
      & lower of the shares of the 4 containment and the 2 release clauses (0) \\
    \addlinespace
    Step cap & 8{,}500 & 22{,}800 & 6{,}000 & 9{,}000 & 13{,}000 \\
    Motion stack & cuRobo & cuRobo & OSC & OSC & OSC \\
    Segmentation & SAM3 & SAM3 & SAM3 & color palette & color palette \\
    \bottomrule
  \end{tabularx}
  \caption{The five tasks. The instruction is the task document's, verbatim, and is the task description the agent
  receives. Success is scored per trial (object); the reward is partial credit, logged but never reported as success.
  Step caps are in control steps.}
  \label{tab:app_tasks}
\end{table}

\paragraph{What each task demands.}
Each task stresses a different part of a policy. Tool Hanging moves only the wrench, so its difficulty lies in a
grasp that keeps the ring free and a collision-free carry onto the hook. Syringe Packing must turn each 150\,mm
syringe upright to pass the 55\,mm aperture, twice, among obstacles; it is the longest task (a median episode of
11{,}345 control steps, and 22\% of episodes reach the step cap against 2--5\% elsewhere) and the collision-geometry
example of Finding~5. Gear Meshing is the only task with a temporal goal: the gear must turn with the pinion for
1\,s of simulated time after the button press, unheld. USB-C Insertion is the precision task, with 0.15\,mm of
clearance per side and a cable dragging on the plug. Cable Routing manipulates a deformable object directly, with
three separately scored sub-goals.

\paragraph{Instruction, embodiment and perception.}
The agent receives the document's instruction verbatim; it is identical across a task's 19 documents, so instances
differ only in the drawn scene. Only Cable Routing is prompted as bimanual (Table~\ref{tab:app_prompt_slots}). The
motion stack and segmentation backend are fixed per task and identical in every run, so no comparison in the main
text mixes back ends.

\subsection{Goal Predicates and Scoring}
\label{app:tasks_scoring}

An episode ends when the policy returns (after a 60-control-step settle, 120 for Gear Meshing), reaches the step
cap or diverges, and a Syringe Packing or Gear Meshing episode also as soon as its goal holds. The simulator then
reads the document's goal, a predicate over object features, contacts and speeds; \texttt{once} and
\texttt{hold-for} clauses track the whole episode. Listing~\ref{lst:app_goals} gives the goal clauses.
\makeatletter\@ifundefined{c@applisting}{\newcounter{applisting}}{}\makeatother
\newcommand{\apptaskgoals}[3]{\VerbatimInput[fontsize=\scriptsize,breaklines,samepage,frame=#1,framesep=1.5mm,%
  firstline=#2,lastline=#3]{appendix/goals.txt}}
\newcommand{\apptaskbreak}{\par\vfil\penalty1000\vfilneg}
\par\medskip\noindent\begin{minipage}{\linewidth}
\refstepcounter{applisting}\label{lst:app_goals}Listing~\theapplisting:~Goal clauses of the five tasks' documents, in
Table~\ref{tab:app_tasks}'s order.\par\vspace{-1pt}
\apptaskgoals{topline}{1}{12}
\end{minipage}\apptaskbreak
\apptaskgoals{none}{14}{23}\apptaskbreak
\apptaskgoals{none}{25}{34}\apptaskbreak
\apptaskgoals{none}{36}{47}\apptaskbreak
\apptaskgoals{bottomline}{49}{64}

\paragraph{Per-object success.}
Success is scored per object with VABAR's scorer: \texttt{derive\_trial\_keys} strips the modal wrappers and makes
one trial per single-object goal, or one per (object, region) pair when a goal has several containment clauses
(Table~\ref{tab:app_tasks}); release clauses belong to no trial, and a diverged episode fails every trial. Scoring
objects rather than episodes keeps the partial progress of the two multi-object tasks: at the kept iterate, 43\% of
held-out Syringe Packing episodes place exactly one of the two syringes, and no Cable Routing program meets every
trial on all ten layouts. Some readings were scored before this rule was final; they count one Syringe Packing trial
per episode (part of the reference runs of the feedback and skill-library studies) or a seated gear as a met Gear
Meshing goal (some Gear Meshing reference runs). The instance analysis
(Appendix~\ref{app:tasks_difficulty}) and the failure analysis (Appendix~\ref{app:failures}) rescore every batch they use with the final rule.

\paragraph{Partial credit.}
Each document also defines a partial-credit reward in $[0,1]$ (Table~\ref{tab:app_tasks}), which the loop logs but
no reported number uses, because it adds little to per-object success: over every training and held-out episode of
the tables' finished runs, 80\%
of Tool Hanging episodes sit exactly at the $1/3$ floor, 3\% of USB-C Insertion rewards lie strictly between 0 and 1,
Gear Meshing's is binary and Syringe Packing's equals the share of syringes placed; only Cable Routing's quarters
add resolution.

\subsection{Scene Generation and Splits}
\label{app:tasks_splits}

\begin{figure}[b]
  \centering
  \includegraphics[width=\linewidth]{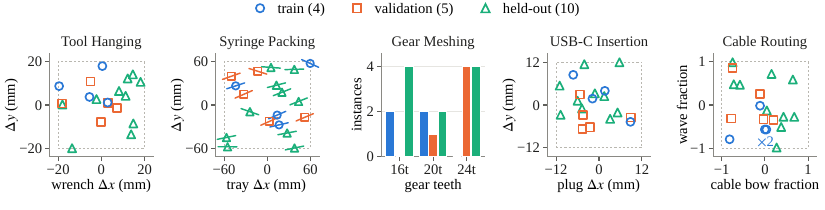}
  \caption{Main randomized quantities of every instance by split, recomputed from the samplers; the segment through
  each Syringe Packing point shows the tray's yaw, and dashed boxes are the sampling ranges
  (Table~\ref{tab:app_splits}).}
  \label{fig:app_splits}
\end{figure}

A Tool Hanging, USB-C Insertion or Cable Routing instance is its base document drawn once by the base's sampler at a
fixed seed, from NumPy's \texttt{default\_rng(seed + episode)} at episode 0. A Gear Meshing or Syringe Packing
instance records its draw as start poses: each gear board is drawn at its own seed, its tooth count from one
balanced draw over the 40 boards, and the 30 syringe layouts are drawn from one \texttt{default\_rng(0)} stream (tray
shifts at least 10\,mm apart). Either way a document reproduces its scene exactly, so every setting runs the same
episodes. Table~\ref{tab:app_splits} lists the samplers, the ranges of every randomized quantity, and the documents of
each split; Figure~\ref{fig:app_splits} shows each instance's draw. The agent reads feedback from the training
episodes only; validation decides when a run stops and which iterate is kept; held-out success is only reported
(Appendix~\ref{app:loop}): Table~\ref{tab:models}'s @$k$ is the best held-out reading within $k$ rounds, and the
other tables read the zero-shot program and the kept iterate.

\begin{table}[t]
  \centering
  \footnotesize
  \setlength{\tabcolsep}{3.5pt}
  \renewcommand{\arraystretch}{1.12}
  \begin{tabularx}{\linewidth}{@{}>{\raggedright\arraybackslash}p{1.3cm}>{\raggedright\arraybackslash}X>{\raggedright\arraybackslash}p{2.6cm}>{\raggedright\arraybackslash}p{1.75cm}@{}}
    \toprule
    Task & Base document, sampler and randomized quantities & Documents: 4 train, 5 val., 10 held-out & Seed \\
    \midrule
    Tool Hanging & \texttt{wrenches\_level1}, \texttt{layout:LEVEL1}: wrench $x,y$ each $\mathcal{U}(\pm20)$\,mm; hook fixed & train: 001, 012--014\newline val.: 015--019\newline held-out: 002--011 & doc $n$: seed $4n{-}3$ \\
    \addlinespace
    Syringe Packing & \texttt{tray\_pair\_both}, 30-trial layout sweep: tray arrangement yaw $\mathcal{U}(\pm25^\circ)$ about the first syringe, shift $x,y$ each $\mathcal{U}(\pm60)$\,mm, shifts $\geq$10\,mm apart; sharps box $x,y$ each $\mathcal{U}(\pm30)$\,mm & train: t00, t11--t13\newline val.: t14--t18\newline held-out: t01--t10 & trial $n$ of one sweep at seed 0 \\
    \addlinespace
    Gear Meshing & 40 drawn single-station boards, \texttt{draw\_layout}: tooth count 16/20/24 (balanced); plate center $x,y$ each $\mathcal{U}(\pm30)$\,mm; chain yaw $\mathcal{U}(\pm10^\circ)$; gear dropped uniformly on the mat ($\geq$15\,mm from the plate), clocked $\mathcal{U}[0,2\pi)$ & train: 01, 12--14\newline val.: 15--19\newline held-out: 02--11 & doc $n$: 31-bit SHA-256 of $(0, n)$; teeth: one balanced draw \\
    \addlinespace
    USB-C Insertion & \texttt{insert\_easy}, \texttt{staging:USBC\_EASY}: plug yaw $\mathcal{U}(\pm8^\circ)$, $x,y$ each $\mathcal{U}(\pm12)$\,mm; receptacle $\mathcal{U}(\pm5)$\,mm across, $\mathcal{U}(\pm1.5)$\,mm along, yaw $\mathcal{U}(\pm5^\circ)$, redrawn until it clears the bench & train: 01, 12--14\newline val.: 15--19\newline held-out: 02--11 & doc $n$: seed $9999{+}n$ \\
    \addlinespace
    Cable Routing & \texttt{port\_easy}, \texttt{staging:PORT\_BOARD}: cable bow $\pm20$\,mm, shift $\pm15$\,mm, yaw $\pm4^\circ$, wave $\pm30$\,mm (fractions $\mathcal{U}(-1,1)$), anchor follows the cable; port $x,y$ each $\mathcal{U}(\pm6)$\,mm, yaw $\mathcal{U}(\pm3^\circ)$; peg gap $\mathcal{U}(0.7, 1.3)\times$ the 240\,mm pitch within the anchor--port corridor, lateral wobble $\pm5$\,mm, board yaw $\mathcal{U}(\pm5.7^\circ)$, board position uniform in the corridor, rejected unless turning, wrap and route length stay near the declared board's values & train: 01, 12--14\newline val.: 15--19\newline held-out: 02--11 & doc $n$: seed $9999{+}n$ \\
    \bottomrule
  \end{tabularx}
  \caption{Scene generation and splits. Each instance is one task document, its base document drawn once as the last column states.}
  \label{tab:app_splits}
\end{table}

\paragraph{Coverage gaps.}
The splits are fixed document ids, not stratified, and nine training and validation draws leave parts of each
sampling range to the held-out split alone. No Gear Meshing training board has 24 teeth and no validation board has
16. Training wrenches are offset at most +2.9\,mm in $x$ (validation +7.4\,mm), while 7 of the 10 held-out wrenches
are offset 8--18\,mm. Training and validation trays are shifted at most 27.6\,mm toward $-y$, while 4 held-out trays
are shifted 39--60\,mm. Training and validation plugs are offset at most 7.2\,mm toward $-x$, while 2 held-out plugs
are offset 10.8--11.0\,mm. Training cables bow by a fraction of at most +0.03 and validation cables by at most +0.20,
while 6 held-out cables bow +0.15 to +0.68.

\subsection{Held-Out Instance Difficulty}
\label{app:tasks_difficulty}

\begin{figure}[t]
  \centering
  \includegraphics[width=\linewidth]{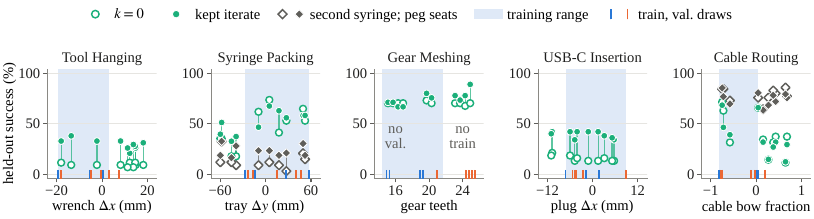}
  \caption{Held-out success per instance: the share of programs that succeed on each held-out object at $k{=}0$
  (hollow) and at the kept iterate (filled), pooled over the runs of all three tables, against one randomized
  quantity. Diamonds are the second syringe and the mean of the two peg seats; the band spans the training draws.}
  \label{fig:app_instance_difficulty}
\end{figure}

To see whether some held-out instances are systematically harder, we pool the runs of all three studies' tables and
compute, per held-out object, the share of programs that succeed on it, at $k{=}0$ (each distinct zero-shot program
once; seeded feedback arms share their reference's) and at the kept iterate, scoring every program with the final
per-object rule (Figure~\ref{fig:app_instance_difficulty}).

\paragraph{Success is mostly a property of the program.}
On Gear Meshing and USB-C Insertion a program usually solves every held-out instance or none: 68\% of zero-shot Gear
Meshing programs solve all ten boards, and even at the kept iterate 84\% of USB-C programs are all-or-nothing, with
per-layout shares of 34--42\% and no trend in the plug's offset, including the two plugs beyond every training draw.
Zero-shot Tool Hanging succeeds on 7--11\% of programs on every layout. On these tasks a held-out score mostly counts
the programs that work.

\paragraph{Layout effects present from the start.}
Where success does vary across layouts, the variation coincides with coverage gaps. On the four Syringe Packing
layouts whose tray lies beyond every training and validation tray in $-y$ (two of them also in $-x$), the first
syringe is placed by 26\% of zero-shot and 40\% of kept programs, against 58\% on the other six, and grasp misses (G
in Table~\ref{tab:app_failures}) make up 21\% of its kept-iterate failures there against 6\% elsewhere; the second
syringe is harder on every layout (23\% against 51\% at the kept iterate). On Cable Routing the peg seats stay at
56--93\% at the kept iterate, but the port trial varies with the cable's bow (12--68\%): on the six layouts bowing
+0.15 or more, 28\% of zero-shot and 24\% of kept programs reach the port, against 58\% and 55\% on the other four.
These six layouts also have the most negative port yaws, which ten layouts cannot separate from the bow.

\paragraph{Gaps that open with learning.}
On Gear Meshing and Tool Hanging the instances start out equal and learning separates them. Zero-shot Gear Meshing
success is flat across tooth counts (70--72\%), but the kept iterate fixes 26\% of the zero-shot failures on 16-tooth
boards against 46--52\% on the others, leaving them at 69\% against 78--79\%: the 24-tooth boards, which no training
board has, gain the most, and the 16-tooth boards, which no validation board has, the least. At the kept iterate the
seven Tool Hanging wrenches offset +8\,mm or more succeed for 28\% of programs against 34\% for the other three, with
reach failures (T) twice as common on them (22\% against 11\% of failures), though single layouts of the two groups
overlap (21--33\% against 33--38\%).

\paragraph{Implications.}
Because every setting meets the same instances, the comparisons of Tables~\ref{tab:models},
\ref{tab:temporal_feedback} and~\ref{tab:spatial_feedback} stay paired, but every task's absolute held-out level
includes instances beyond the training draws. These cost success from the start only on Syringe Packing and Cable
Routing, and learning does not close those gaps: it narrows Syringe Packing's from 31 to 18 points and leaves Cable
Routing's unchanged. Of the two gaps that open with learning, the larger (Gear Meshing, 10
points) lies where validation, which selects the kept iterate, has no instance, not where training has none.

\FloatBarrier
\section{Skill Library}
\label{app:skills}

The agent composes its policy over $\mathcal{S} = \mathcal{S}_0 \cup \mathcal{S}g \cup \mathcal{S}\tau \cup \mathcal{S}_A$ (Section~\ref{sec:protocol}), where $\mathcal{S}_A$ is the code it writes. We first describe how a program uses a skill, the three parts the harness provides, and what each experiment ships; the skill-library study qualifies Finding~2.

\subsection{Library Format and Use}
\label{app:skills_format}
The general skills and the tool bundles ship in the Agent Skills format: a directory with a \texttt{SKILL.md} (when the skill applies, its state flow, declared tools, required inputs and named exits) and scripts. Each subgraph of a program carries a \texttt{skill} label, the $\mathcal{S}_g$ skill it runs or \texttt{generated} for a stage the agent wrote. With the source, the agent copies the skill's scripts into its workflow; in the docs-callable arm a node names a script and the runtime resolves it from the named skill. Every run enforces its arm: the runtime loads only the shipped skills, the guard refuses other labels, and the library checkout is closed to the agent. All runs read open-robot-skills \texttt{ed6ab27}, except three early Table~\ref{tab:models} runs, which read \texttt{113261b} (the same skills, three scripts differ); the earliest reference runs read a \texttt{113261b} working tree whose skills equal \texttt{ed6ab27}'s.

\subsection{Primitive Tools (\texorpdfstring{$\mathcal{S}_0$}{S0})}
\label{app:skills_tools}
$\mathcal{S}_0$ holds 70 to 78 tools per task: robot primitives, the motion backend's planner, six tool bundles (cuRobo planning, curve fitting, point-cloud geometry, Grounding DINO detection, SAM3 segmentation and tracking, a VLM) and, on USB-C Insertion and Cable Routing, a few audited tools of the task's own (Table~\ref{tab:app_tools}). The agent gets \texttt{SKILLS.md}, which documents every tool, and the six bundle manuals; \texttt{sim.check\_success} and the privileged \texttt{sim.*} state readers are withheld from every run, since the goal is what the agent has to learn. $\mathcal{S}_0$ is thus not a set of bare primitives: detection, segmentation and grasp proposals are single calls.

\begin{table}[ht]
  \centering
  \footnotesize
  \setlength{\tabcolsep}{3.5pt}
  \setlength{\aboverulesep}{0pt}
  \setlength{\belowrulesep}{0pt}
  \renewcommand{\arraystretch}{1.15}
  \begin{tabularx}{\linewidth}{@{}l|>{\raggedright\arraybackslash\scriptsize}X@{}}
    \toprule
    Namespace & \footnotesize Tools \\
    \midrule
    \texttt{robot} (23) & \texttt{close\_gripper}, \texttt{describe\_arm}, \texttt{describe\_gripper}, \texttt{describe\_workspace}, \texttt{execute\_trajectory}, \texttt{forward\_kinematics}, \texttt{get\_camera\_pose}, \texttt{get\_ee\_pose}, \texttt{get\_gripper}, \texttt{get\_gripper\_pose}, \texttt{get\_observation}, \texttt{go\_home}, \texttt{go\_to\_pose}, \texttt{go\_to\_pose\_cartesian}, \texttt{grasp\_frame}, \texttt{instruction}, \texttt{move\_cartesian\_until\_contact}, \texttt{move\_to\_joints}, \texttt{open\_gripper}, \texttt{set\_grip}, \texttt{solve\_ik}, \texttt{stream\_dual}, \texttt{wait\_steps} \\
    \texttt{motion} (5/8) & \texttt{describe\_planner}, \texttt{get\_robot\_collision\_spheres}, \texttt{plan\_joint}, \texttt{plan\_linear}, \texttt{plan\_support\_disengagement}; with cuRobo also \texttt{build\_world\_tsdf}, \texttt{plan\_to\_pose}, \texttt{solve\_ik\_batch} \\
    \texttt{curobo} (9) & \texttt{cloud\_to\_attachment}, \texttt{plan\_directed\_linear}, \texttt{plan\_grasp\_motion}, \texttt{plan\_linear}, \texttt{plan\_to\_grasp\_poses}, \texttt{plan\_to\_pose}, \texttt{plan\_with\_grasped\_object}, \texttt{validate\_joint\_trajectory\_grasped}, \texttt{validate\_joint\_trajectory\_robot} \\
    \texttt{curve} (2) & \texttt{fit\_centerline}, \texttt{track\_centerline} \\
    \texttt{geometry} (22) & \texttt{build\_world\_config}, \texttt{cloud\_to\_attachment}, \texttt{compute\_drop\_position}, \texttt{compute\_feature\_mate}, \texttt{compute\_obb}, \texttt{compute\_xy\_distance}, \texttt{depth\_to\_point\_cloud}, \texttt{exclude\_robot\_points}, \texttt{filter\_and\_compute\_obb}, \texttt{filter\_noise}, \texttt{fit\_linear\_feature}, \texttt{fit\_planar\_feature}, \texttt{front\_grasp\_from\_obb}, \texttt{iou}, \texttt{mask\_to\_world\_points}, \texttt{pixel\_to\_world\_point}, \texttt{pose\_distance}, \texttt{rotate\_quat\_z90}, \texttt{select\_top\_down\_grasp}, \texttt{top\_down\_grasp\_candidates}, \texttt{top\_down\_grasp\_from\_obb}, \texttt{transform\_points} \\
    \texttt{grounding-dino} (1) & \texttt{detect} \\
    \texttt{sam3} (6) & \texttt{segment\_box}, \texttt{segment\_point}, \texttt{segment\_text}, \texttt{tracker\_close}, \texttt{tracker\_init}, \texttt{tracker\_update} \\
    \texttt{vlm} (2) & \texttt{query}, \texttt{query\_yes\_no} \\
    \texttt{usbc} (5) & \texttt{grip\_state}, \texttt{instruction}, \texttt{next\_pair}, \texttt{receptacle\_shape}, \texttt{set\_gripper} \\
    \texttt{cable} (8) & \texttt{carry\_with\_feed}, \texttt{curve\_memory}, \texttt{next\_station}, \texttt{plan\_station}, \texttt{plan\_support}, \texttt{set\_gripper}, \texttt{support\_memory}, \texttt{wait\_steps} \\
    \bottomrule
  \end{tabularx}
  \caption{The tool surface $\mathcal{S}_0$ (the \texttt{none} privilege tier every run uses), by namespace. Each task gets the 23 \texttt{robot} tools, the \texttt{motion} tools of its backend (8 with cuRobo on Tool Hanging and Syringe Packing, 5 with OSC on the other tasks) and the 42 tools of the six bundles that ship with open-robot-skills; USB-C Insertion and Cable Routing add 5 and 8 of their own tools, admitted after an audit found they read no privileged state. Totals from Tool Hanging to Cable Routing: 73, 73, 70, 75 and 78 tools.}
  \label{tab:app_tools}
\end{table}

\subsection{General Skills (\texorpdfstring{$\mathcal{S}_g$}{Sg})}
\label{app:skills_general}
open-robot-skills holds 30 skills with 72 scripts (Table~\ref{tab:app_skills_ors}), most of them for perception (11) and grasping (8); five skills compute, plan and execute the motion of a held object. A run admits only skills whose scripts call nothing outside its $\mathcal{S}_0$, so no program reaches simulator state through a skill: three skills call withheld simulator tools and never ship, and the two \texttt{executing-*} skills need cuRobo's \texttt{motion.plan\_to\_pose}. This leaves 27 callable skills on the cuRobo tasks and 25 on the OSC tasks.

\begin{table}[p]
  \centering
  \scriptsize
  \setlength{\tabcolsep}{3.5pt}
  \setlength{\aboverulesep}{0pt}
  \setlength{\belowrulesep}{0pt}
  \renewcommand{\arraystretch}{1.15}
  \begin{tabularx}{\linewidth}{@{}lccc>{\raggedright\arraybackslash}X@{}}
    \toprule
    Skill & Scripts & cuRobo & OSC & What it does \\
    \midrule
    \multicolumn{5}{@{}l}{\textit{Perception} (11)} \\
    \texttt{perceiving-deformable-linear-objects} & 1 & \checkmark & \checkmark & Ordered 3D cable centerline from RGB-D, kept current across a task \\
    \texttt{perceiving-functional-features} & 6 & \checkmark & \checkmark & Fits a described part (loop, shaft, tip, aperture) from RGB-D \\
    \texttt{perceiving-next-item} & 3 &  &  & Container and next remaining item from one view, or done \\
    \texttt{perceiving-object-parts} & 1 & \checkmark & \checkmark & Detects the parent object, then segments a named subpart in its crop \\
    \texttt{perceiving-objects} & 1 & \checkmark & \checkmark & DINO detection, VLM crop tournament, SAM3 mask, world-frame box \\
    \texttt{perceiving-objects-oneshot} & 1 & \checkmark & \checkmark & DINO plus one VLM set-of-marks pick, with a clean not-found exit \\
    \texttt{perceiving-relational-correspondences} & 2 & \checkmark & \checkmark & Language-conditioned matching of source and destination instances \\
    \texttt{perceiving-routing-fixtures} & 2 & \checkmark & \checkmark & Surveys a cable bench: stations, cable centerline, owed crossing sides \\
    \texttt{perceiving-sorting-pairs} & 3 & \checkmark & \checkmark & Labeled four-compartment destination and the next source with its region \\
    \texttt{reconstructing-collision-worlds} & 2 & \checkmark & \checkmark & Planner collision world from RGB-D, with a carved approach corridor \\
    \texttt{registering-held-objects} & 1 & \checkmark & \checkmark & Re-observes a held object from the wrist; feature in TCP frame \\
    \midrule
    \multicolumn{5}{@{}l}{\textit{Tracking} (1)} \\
    \texttt{tracking-objects} & 0 & \checkmark & \checkmark & Long-running SAM3 tracker publishing an object's mask and box \\
    \midrule
    \multicolumn{5}{@{}l}{\textit{Grasping} (8)} \\
    \texttt{grasping-concave-contour} & 1 & \checkmark & \checkmark & Grasps an awkward part across inward waists of its silhouette \\
    \texttt{grasping-direct-ik} & 4 & \checkmark & \checkmark & IK grasp that aligns above the target, then descends straight down \\
    \texttt{grasping-linear-feature} & 2 & \checkmark & \checkmark & Fits an elongated part's 3D axis; perpendicular, inclination-aware grasp \\
    \texttt{grasping-short-axis} & 7 & \checkmark & \checkmark & Deterministic top-down cuRobo grasp across the box's short axis \\
    \texttt{grasping-with-planner} & 6 & \checkmark & \checkmark & Axis-locked top-down linear descent with a cuRobo planner fallback \\
    \texttt{proposing-short-axis-grasps} & 1 & \checkmark & \checkmark & Proposes top-down grasps closing across an elongated object's short axis \\
    \texttt{proposing-side-grasps} & 1 & \checkmark & \checkmark & Proposes side-entry grasps at twelve azimuths, ranked by jaw margin \\
    \texttt{selecting-reachable-grasp} & 1 &  &  & Takes the first ranked grasp candidate the arm can reach \\
    \midrule
    \multicolumn{5}{@{}l}{\textit{Planning} (2)} \\
    \texttt{computing-feature-mating-poses} & 1 & \checkmark & \checkmark & Approach, engaged and mate TCP poses from held and fixture features \\
    \texttt{planning-held-object-motion} & 3 & \checkmark & \checkmark & Plans clearance lift, reorientation, transit and engagement for a held object \\
    \midrule
    \multicolumn{5}{@{}l}{\textit{Motion} (3)} \\
    \texttt{executing-feature-mating} & 2 & \checkmark &  & Executes a mating plan: tracked legs, servo, contact seating, release \\
    \texttt{executing-held-object-motion} & 2 & \checkmark &  & Collision-aware pose sequence for a held object, without releasing it \\
    \texttt{transporting-objects} & 11 & \checkmark & \checkmark & Carries the held object to a destination and releases it \\
    \midrule
    \multicolumn{5}{@{}l}{\textit{Manipulation} (2)} \\
    \texttt{placing-held-feature} & 1 &  &  & Moves a held object so its declared feature lands on a target \\
    \texttt{seating-a-cable-crossing} & 1 & \checkmark & \checkmark & Streams one cable crossing leg, then measures its seat by vision \\
    \midrule
    \multicolumn{5}{@{}l}{\textit{Verification} (3)} \\
    \texttt{verifying-a-cable-route} & 1 & \checkmark & \checkmark & Judges a finished cable route from a camera, with margins \\
    \texttt{verifying-grasps} & 3 & \checkmark & \checkmark & Checks arrival, jaw gap and lifted visibility of one grasp \\
    \texttt{verifying-placement} & 1 & \checkmark & \checkmark & Confirms from one RGB-D view that an object passed an aperture \\
    \midrule
    Total: 30 skills & 72 & 27 & 25 & \\
    \bottomrule
  \end{tabularx}
  \caption{The general skills $\mathcal{S}_g$ (open-robot-skills) by the category their \texttt{SKILL.md} declares, with their Python scripts (\texttt{tracking-objects} runs from a tool module) and whether each is callable on the cuRobo tasks (Tool Hanging, Syringe Packing) and the OSC tasks (Gear Meshing, USB-C Insertion, Cable Routing), i.e.\ whether that task's $\mathcal{S}_0$ holds every tool its scripts call.}
  \label{tab:app_skills_ors}
\end{table}

\subsection{Expert Skills (\texorpdfstring{$\mathcal{S}_\tau$}{Stau})}
\label{app:skills_expert}
For each task, $\mathcal{S}_\tau$ holds the scripts of the human-written VABAR vision policy (Table~\ref{tab:app_expert}): 11 bundles with 43 entry scripts, the script nodes of each human graph without its wiring. Expert scripts carry no exit contract, so the agent copies what it uses into its own scripts and labels those subgraphs \texttt{generated}. $\mathcal{S}_\tau$ ranges from 17 entry scripts on Syringe Packing to a single glue script on Tool Hanging, whose human policy is built from 15 scripts of 10 skills in $\mathcal{S}_g$.

\begin{table}[ht]
  \centering
  \footnotesize
  \setlength{\tabcolsep}{3.5pt}
  \setlength{\aboverulesep}{0pt}
  \setlength{\belowrulesep}{0pt}
  \renewcommand{\arraystretch}{1.15}
  \begin{tabularx}{\linewidth}{@{}l|l|>{\raggedright\arraybackslash}X@{}}
    \toprule
    Task, human graph & Bundle & Scripts \\
    \midrule
    Tool Hanging & glue (1) & \texttt{choose\_placement\_mode} \\
    \texttt{gap\_perception} & from $\mathcal{S}_g$ (15) & \texttt{compute\_mate}, \texttt{execute\_placement\_plan}, \texttt{release\_and\_retract}, \texttt{execute\_approach\_only}, \texttt{execute\_reorientation}, \texttt{compute\_align\_pose}, \texttt{execute\_grasp\_align}, \texttt{perceive\_fixture\_feature}, \texttt{perceive\_object\_feature}, \texttt{select\_next\_source}, \texttt{plan\_clearance\_motion}, \texttt{plan\_feature\_engagement}, \texttt{build\_collision\_world}, \texttt{register\_held}, \texttt{descend\_release\_linear} (of 10 skills) \\
    \midrule
    Syringe Packing & perception (8) & \texttt{build\_collision\_world}, \texttt{count\_remaining}, \texttt{map\_syringe\_to\_hand}, \texttt{perceive\_fixture}, \texttt{perceive\_target}, \texttt{verify\_grasp}, \texttt{verify\_placement}, \texttt{verify\_seated} \\
    \texttt{gap\_perception\_v2} & motion (9) & \texttt{compute\_grasp}, \texttt{execute\_insertion}, \texttt{execute\_reorientation}, \texttt{grasp\_checked}, \texttt{lift\_held}, \texttt{plan\_clearance\_motion}, \texttt{plan\_visual\_insert}, \texttt{release\_and\_retract}, \texttt{unjam\_aperture} \\
     & helpers & \texttt{motion\_checks}, \texttt{perception\_checks} \\
     & from $\mathcal{S}_g$ (1) & \texttt{compute\_mate} (of \texttt{computing-feature-mating-poses}) \\
    \midrule
    Gear Meshing & perception (1) & \texttt{watch} \\
    \texttt{gap\_perception} & motion (6) & \texttt{carry}, \texttt{clear}, \texttt{next\_station}, \texttt{pick}, \texttt{press}, \texttt{seat} \\
     & helpers & \texttt{gearboard}, \texttt{sight} \\
    \midrule
    USB-C Insertion & perception (4) & \texttt{check\_grasp}, \texttt{ground\_vision}, \texttt{inspect\_vision}, \texttt{relook\_vision} \\
    \texttt{mate} & motion (2) & \texttt{grip\_step}, \texttt{move\_stage} \\
     & glue (2) & \texttt{next\_pair}, \texttt{route\_perception} \\
     & helpers & \texttt{bore\_cv}, \texttt{held\_frame} \\
     & omitted (5) & \texttt{ground\_oracle}, \texttt{inspect\_oracle}, \texttt{relook\_oracle} (call \texttt{sim.perceive\_mate}); \texttt{hold\_plan} (calls \texttt{usbc.plan\_hold}); \texttt{plan\_stroke} (calls \texttt{usbc.plan\_mate}) \\
    \midrule
    Cable Routing & perception (3) & \texttt{perceive\_rod}, \texttt{perceive\_weave\_vision}, \texttt{verify\_weave\_vision} \\
    \texttt{weave\_terminated} & motion (6) & \texttt{anchor\_far\_end}, \texttt{plan\_crossing}, \texttt{plug\_port}, \texttt{restore\_seats}, \texttt{run\_legs}, \texttt{seat\_station} \\
     & glue (1) & \texttt{next\_station} \\
     & helpers & \texttt{fixtures\_cv}, \texttt{grasp\_vision}, \texttt{terminal\_vision} \\
    \bottomrule
  \end{tabularx}
  \caption{Expert skills $\mathcal{S}_\tau$: the entry scripts of each task's human VABAR vision policy, bundled by the tools they call (count in parentheses), without the graph that wires them. Helpers are modules the entry scripts import; scripts taken from $\mathcal{S}_g$ ship there, and scripts that call a tool outside $\mathcal{S}_0$ are omitted.}
  \label{tab:app_expert}
\end{table}

\subsection{Availability per Experiment}
\label{app:skills_access}
Table~\ref{tab:app_skill_access} lists what each experiment ships and how many runs it has; every setting of Table~\ref{tab:models} gets the whole library. The feedback study's reference omits $\mathcal{S}_\tau$ because early zero-shot programs copied the human scripts, and a reference built on human code would measure that code more than the feedback. The skill-library study removes one part of the reference at a time: docs callable withholds the source but keeps every skill documented and callable, separating a skill's interface from its code, and tools only removes $\mathcal{S}_g$.

\begin{table}[ht]
  \centering
  \footnotesize
  \setlength{\tabcolsep}{3.5pt}
  \setlength{\aboverulesep}{0pt}
  \setlength{\belowrulesep}{0pt}
  \renewcommand{\arraystretch}{1.15}
  \begin{tabularx}{\linewidth}{@{}l|*{4}{>{\centering\arraybackslash}X}@{}}
    \toprule
    & Table~\ref{tab:models}, every setting & Feedback study; full library & Tools only & Docs callable \\
    \midrule
    Library & $\mathcal{S}_0 + \mathcal{S}_g + \mathcal{S}_\tau$ & $\mathcal{S}_0 + \mathcal{S}_g$ & $\mathcal{S}_0$ & $\mathcal{S}_0$ + docs of $\mathcal{S}_g$ \\
    $\mathcal{S}_g$ skills callable & 27\,/\,25 & 27\,/\,25 & 0 & 27\,/\,25 \\
    $\mathcal{S}_g$ source readable & \checkmark & \checkmark & -- & -- \\
    $\mathcal{S}_\tau$ expert scripts & \checkmark & -- & -- & -- \\
    \midrule
    \texttt{SKILLS.md} & \checkmark & \checkmark & \checkmark & \checkmark \\
    \texttt{SKILL\_LIBRARY.md} & $\mathcal{S}_g$, $\mathcal{S}_\tau$ & $\mathcal{S}_g$ & notice only & $\mathcal{S}_g$ \\
    \texttt{skills/} & $\mathcal{S}_g$, $\mathcal{S}_\tau$ & $\mathcal{S}_g$ & -- & -- \\
    \texttt{WORKFLOW\_FORMAT.md} & \checkmark & \checkmark & \checkmark & \checkmark \\
    \midrule
    Runs (all tasks) & 250 & 200 & 25 & 25 \\
    \bottomrule
  \end{tabularx}
  \caption{Skill availability per experiment. Callable counts are for the cuRobo\,/\,OSC tasks; the middle rows are the workspace files that carry each part (\texttt{SKILLS.md} documents $\mathcal{S}_0$, \texttt{SKILL\_LIBRARY.md} lists the callable skills, \texttt{skills/} holds their source).}
  \label{tab:app_skill_access}
\end{table}

\FloatBarrier
\section{Agent Harness}
\label{app:harness}

Every run of Table~\ref{tab:models} of the feedback study (Tables~\ref{tab:temporal_feedback}
and~\ref{tab:spatial_feedback}) executes the
loop of Section~\ref{sec:protocol} in one harness; the runs differ only in the agent and its reasoning budget, the
shipped skills, the feedback renderer, the evaluation schedule and the time budgets. Each instance set has one role:
the agent sees only the training episodes, validation decides when a run stops and which iterate it keeps, and the
held-out set only measures. We follow one iteration of the loop (Appendix~\ref{app:loop}), then give each study's
evaluation schedule (Appendix~\ref{app:schedule}) and the workspace and the checks a proposal passes
(Appendix~\ref{app:guard}). Appendices~\ref{app:prompts} and~\ref{app:feedback} reproduce what the agent reads, and
Appendix~\ref{app:stopping} gives the revert, stopping and selection rules and relates the @$k$ of
Table~\ref{tab:models} to the validation-selected iterate.

\subsection{Self-Learning Loop}
\label{app:loop}

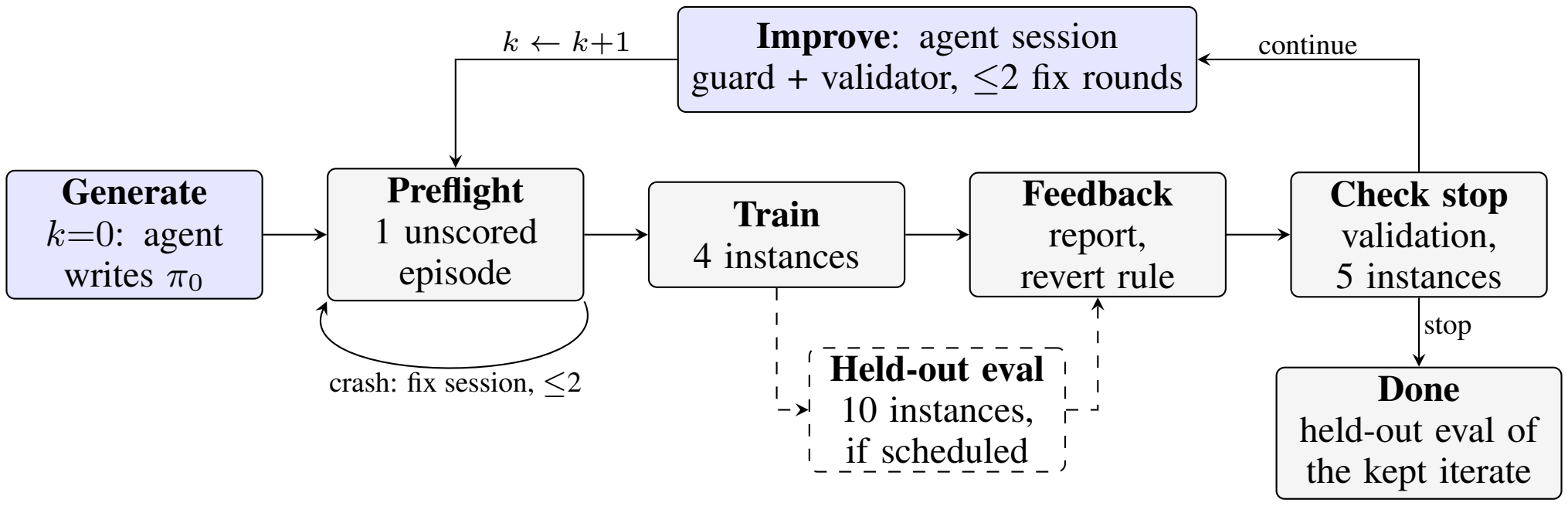
\begin{figure}[ht]
  \centering
  \begin{tikzpicture}[
      font=\footnotesize, >=stealth,
      st/.style={draw, rounded corners=2pt, align=center, text width=2.05cm, minimum height=0.9cm, inner sep=2pt, fill=black!4},
      ag/.style={st, fill=blue!10},
      opt/.style={st, dashed, fill=white},
      lab/.style={font=\scriptsize, inner sep=1pt}]
    \node[ag] (gen) at (0,0) {\textbf{Generate}\\$k{=}0$: agent writes $\pi_0$};
    \node[st] (pre) at (2.75,0) {\textbf{Preflight}\\1 unscored episode};
    \node[st] (tr) at (5.5,0) {\textbf{Train}\\4 instances};
    \node[st] (fb) at (8.25,0) {\textbf{Feedback}\\report, revert rule};
    \node[st] (cs) at (11,0) {\textbf{Check stop}\\validation, 5 instances};
    \node[ag, text width=4.3cm] (imp) at (6.875,1.5) {\textbf{Improve}: agent session\\guard + validator, $\le$2 fix rounds};
    \node[opt] (ev) at (6.875,-1.5) {\textbf{Held-out eval}\\10 instances, if scheduled};
    \node[st, text width=2.3cm] (done) at (11,-1.7) {\textbf{Done}\\held-out eval of the kept iterate};
    \draw[->] (gen) -- (pre);
    \draw[->] (pre) -- (tr);
    \draw[->] (tr) -- (fb);
    \draw[->] (fb) -- (cs);
    \draw[->, dashed] (tr.south) |- (ev.west);
    \draw[->, dashed] (ev.east) -| (fb.south);
    \draw[->] (cs) -- node[lab, right, pos=0.45] {stop} (done);
    \draw[->] (cs.north) |- node[lab, above, pos=0.75] {continue} (imp.east);
    \draw[->] (imp.west) -| node[lab, above, pos=0.25] {$k \leftarrow k{+}1$} (pre.north);
    \draw[->] (pre.south east) .. controls +(0.35,-0.75) and +(-0.35,-0.75) .. node[lab, below] {crash: fix session, $\le$2} (pre.south west);
  \end{tikzpicture}
  \caption{The self-learning loop. Blue states and the preflight's fix sessions run an agent session; the harness
  runs everything else. No validation or held-out reading reaches the agent.}
  \label{fig:app_loop}
\end{figure}

At $k{=}0$ the agent writes $\pi_0$ from the task brief; every iteration then runs one unscored preflight episode on
the first training instance, the four training episodes and, when scheduled, the held-out batch, and builds the
feedback report from the training episodes alone, after the revert rule has chosen the policy the next session edits
(Figure~\ref{fig:app_loop}). The five validation episodes run after the report is written, for the stopping and
selection rules (Appendix~\ref{app:stopping}); unless the run stops, a new agent session edits the policy into
$\pi_{k+1}$. The session keeps no conversation from earlier sessions, but its report summarizes every earlier iterate
(training result, edit, behavior and the previous session's own account, Appendix~\ref{app:feedback}) and
\texttt{history/} holds the exact diffs. A rate limit that outlasts the retries of Table~\ref{tab:app_inference}
defers the round without recording an iteration; any other provider error ends the session without an edit and
counts as an unchanged iteration (1\% of iterations).

\subsection{Evaluation Schedule}
\label{app:schedule}

Table~\ref{tab:app_schedule} gives the schedule of each study and Appendix~\ref{app:tasks} the instance splits.
Table~\ref{tab:models} reports a learning curve, so it scores the held-out set every round; the 6-iteration studies
report only $\pi_0$ and the kept iterate, and score only those two.

Two first-pass runs of the Opus~5, xhigh row of Table~\ref{tab:models} (Tool Hanging and Gear Meshing) depart from
the splits: they train on the first training instance repeated four times and have no validation set
(Appendix~\ref{app:first_passes}). That row reuses
the first three iterates of 6-iteration runs with the general and expert skills ($\mathcal{S}_g \cup
\mathcal{S}_\tau$); its @$k$ averages the runs with a held-out reading at round $k$: the $k{=}0$ reading, the kept
iterate's, one carried from an earlier stop, or a re-score of the $k{=}1$ or $k{=}2$ iterate on the same ten held-out
instances after the run.

\begin{table}[ht]
  \centering
  \small
  \setlength{\tabcolsep}{4pt}
  \setlength{\aboverulesep}{0pt}
  \setlength{\belowrulesep}{0pt}
  \renewcommand{\arraystretch}{1.2}
  \begin{tabularx}{\linewidth}{l|>{\raggedright\arraybackslash}X|>{\raggedright\arraybackslash}X}
    \toprule
    & Table~\ref{tab:models} (models) & Feedback and skill-library studies \\
    \midrule
    Iterations & 3 ($k{=}0,1,2$) & 6 ($k{=}0,\dots,5$) \\
    Training & 4 instances per iteration & 4 instances per iteration \\
    Validation & 5 instances per iteration & 5 instances per iteration \\
    Held-out & 10 instances per iteration & 10 instances at $k{=}0$ and on the kept iterate \\
    \midrule
    Initial policy $\pi_0$ & written by the agent & written by the agent; in the feedback study every arm reuses the reference run's $\pi_0$ with its training batch and held-out reading, and runs its own validation \\
    Kept iterate & \multicolumn{2}{l}{first iterate with the highest validation success} \\
    Reported & @$k$: best held-out success within $k$ rounds & initial policy: held-out success of $\pi_0$; other rows: held-out success of the kept iterate \\
    \bottomrule
  \end{tabularx}
  \caption{Evaluation schedule. Every instance is one episode, scored per object; an unscored preflight episode
  precedes each training batch. The Opus~5, xhigh row of Table~\ref{tab:models} runs the right column's schedule and
  reports @$k$ (Appendix~\ref{app:schedule}).}
  \label{tab:app_schedule}
\end{table}

\subsection{Workspace, Guard and Preflight}
\label{app:guard}

\paragraph{Workspace and sandbox.}
Each session runs in a sealed workspace whose files Listing~\ref{lst:app_mechanics} lists; \texttt{workflow/} holds
the policy and is the only directory the harness takes back. Claude Code runs with a deny list that blocks reading,
listing and searching the simulator, the harness, the skill sources and all run outputs (the whole home directory),
and with shell access limited to \texttt{./validate.sh} (plus \texttt{echo} and \texttt{mkdir}). The Codex CLI cannot
start its own sandbox on our machines, so its \texttt{AGENTS.md} states the same rules and off-limits paths, and the
harness audits every command and file change of each session for those paths afterwards; no command or file change
named one.

\paragraph{Guard.}
A static guard accepts an improve session's proposal only if every tool it names (tool nodes, recovery steps, literal
\texttt{ctx.tool("...")} calls) is on the tier's allowlist, no edited script computes a tool name at run time, every
subgraph label names a shipped skill or marks the subgraph as the agent's own, and \texttt{./validate.sh} passes.
$\pi_0$ and preflight repairs skip the guard and are checked only by the \texttt{./validate.sh} their session is asked
to run. A rejected proposal returns to the agent with the violations for up to two fix rounds; one still rejected, or
unchanged, carries the edited policy ($\pi_k$, or the incumbent after a revert) forward. The guard rejects 9\% of
first proposals, never for a withheld tool or an unshipped skill: every rejection is for a tool name computed at run
time, which no static check can clear, and every rejected proposal passes after one fix round.

\paragraph{Preflight.}
The preflight separates a policy that cannot run from one that runs and fails. Before anything is scored, an episode
goes back to a fix session (up to two) if its graph fails to load or an error message carries a Python error no
correct program raises (\texttt{AttributeError}, \texttt{ImportError}, \texttt{IndexError}, \texttt{KeyError},
\texttt{NameError}, \texttt{SyntaxError}, \texttt{TypeError}, \texttt{UnboundLocalError}, \texttt{ZeroDivisionError}),
a tool-argument error or tool refusal, or one of a list of interpreter, tool-refusal and executor-binding messages.
Any other error, such as a \texttt{RuntimeError} a policy raises to report a failure, counts as a failed episode.
Capped at 1{,}500 steps, the preflight costs a broken program one short episode instead of a scored batch of identical
failures. 3\% of preflights crash (5\% in the 3-iteration runs, 2\% in the 6-iteration runs), mostly on
\texttt{TypeError}, graphs that fail to load and tool-argument errors; fix sessions repair 88\% of them, and the rest,
all in 3-iteration runs, run their batch unrepaired.
Table~\ref{tab:app_failure_events} counts these interventions per 100 agent sessions; preflights crash most often
in the restricted skill-library arms (7.4 per 100 sessions), whose programs call raw tools or skills with hidden source.

\begin{table}[ht]
  \centering
  \footnotesize
  \setlength{\tabcolsep}{3.5pt}
  \setlength{\aboverulesep}{0pt}
  \setlength{\belowrulesep}{0pt}
  \renewcommand{\arraystretch}{1.15}
  \begin{tabular}{l|c|c|c|c}
    \toprule
    Event (per 100 agent sessions) & Table~\ref{tab:models} & Feedback & Skill library & All \\
    \midrule
    Guard rejections & 5.5 & 6.3 & 4.3 & 5.7 \\
    \quad non-literal tool name (\%) & 100 & 100 & 100 & 100 \\
    Crashed preflights & 3.6 & 0.5 & 7.4 & 2.7 \\
    \quad repaired by fix sessions (\%) & 82 & 100 & 100 & 88 \\
    \quad run unrepaired (\%) & 18 & 0 & 0 & 12 \\
    Reverts & 8.7 & 16.7 & 3.2 & 11.6 \\
    \quad training success rate fell (\%) & 37 & 43 & 67 & 42 \\
    \quad closest approach grew (\%) & 34 & 36 & 0 & 35 \\
    \quad every episode aborted (\%) & 27 & 17 & 33 & 21 \\
    \quad fewer goal reaches or carries (\%) & 2 & 3 & 0 & 3 \\
    Deferred rounds (rate limit) & 1.9 & 3.4 & 0.0 & 2.4 \\
    API errors recorded as unchanged & 0.2 & 0.0 & 3.2 & 0.4 \\
    Updates left unchanged by the agent & 1.3 & 0.7 & 0.0 & 0.9 \\
    \bottomrule
  \end{tabular}
  \caption{Harness events per 100 agent sessions (zero-shot generation, improvement and preflight-fix sessions), with the studies split as in Table~\ref{tab:app_cost}; indented rows give the share (\%) of the event above. The Opus~5, xhigh runs of Table~\ref{tab:models} count all six rounds.}
  \label{tab:app_failure_events}
\end{table}

\FloatBarrier
\makeatletter
\@ifundefined{c@applisting}{\newcounter{applisting}}{}
\@ifundefined{app@box}{\newsavebox{\app@box}}{}
\@ifundefined{ifapp@fresh}{\newif\ifapp@fresh}{}
\providecommand{\appneedspace}[1]{\par\begingroup\setlength{\dimen@}{#1}\vskip\z@\@plus\dimen@\penalty-100%
  \vskip\z@\@plus-\dimen@\vskip\dimen@\penalty9999\vskip-\dimen@\vskip\z@skip\endgroup}
\providecommand{\app@listing}[3]{\noindent\refstepcounter{applisting}\label{#1}%
  {Listing~\theapplisting:~#2}\par\nopagebreak\vspace{-1pt}%
  \if\relax\detokenize{#3}\relax\else{\@beginparpenalty\@M\fvset{breaksymbolleft={},breakindent=1em}\listingfile{#3}}\fi}
\providecommand{\appboxed}[2][\medskipamount]{\par\setbox\app@box\hbox{\begin{minipage}[t]{\linewidth}#2\end{minipage}}%
  \appneedspace{\dimexpr\ht\app@box+\dp\app@box+#1\relax}\vskip#1\box\app@box\par}
\providecommand{\applisting}[3]{%
  \if\relax\detokenize{#3}\relax\appneedspace{14\baselineskip}\global\app@freshtrue\else\appneedspace{7\baselineskip}\fi
  \medskip\app@listing{#1}{#2}{#3}}
\providecommand{\appboxlisting}[3]{\appboxed{\app@listing{#1}{#2}{#3}}}
\providecommand{\app@panel}[2]{{\footnotesize\textit{#1}}\par\nopagebreak\vspace{-3pt}%
  {\@beginparpenalty\@M\fvset{breaksymbolleft={},breakindent=1em}\listingfile{#2}}}
\providecommand{\appsublisting}[2]{%
  \ifapp@fresh\global\app@freshfalse\par\smallskip\noindent\begin{minipage}[t]{\linewidth}\app@panel{#1}{#2}\end{minipage}\par
  \else\appboxed[\smallskipamount]{\app@panel{#1}{#2}}\fi}
\makeatother

\section{Prompts}
\label{app:prompts}

Every coding-agent session starts in a sealed workspace, and the whole prompt is its instruction file \texttt{INSTRUCTIONS.md}: the session's command-line prompt (Table~\ref{tab:app_inference}) only points to that file, and we add no system prompt. \texttt{INSTRUCTIONS.md} is a task template with its \texttt{\{\{slots\}\}} filled, followed by a workspace block. We give the session templates, the workspace block, the slots whose values depend on the task, and what differs between the arms of a study. The templates are byte-identical in every harness snapshot the runs used.

\subsection{Session Templates}
\label{app:prompt_templates}

A generate session ($k{=}0$) reads Listing~\ref{lst:app_generate}; every improve session reads Listing~\ref{lst:app_improve}, with the feedback report in \texttt{FEEDBACK.md} (Appendix~\ref{app:feedback}). When the guard or the validator rejects a proposal, the next round edits the rejected proposal with Listing~\ref{lst:app_fixval} appended. When the unscored preflight episode raises a program error, a fix session runs the improve template with Listing~\ref{lst:app_fixpre} appended.

The templates hold only task-independent advice. The generate brief asks for stages with a routable failure exit each, so that a failure ends the episode at a named step the report can name instead of exhausting the step budget. The improve brief names every channel a report may carry, including the stills that only the stills arm renders (Table~\ref{tab:app_feedback_conditions}), so that it is the same text under every feedback condition.

\appboxlisting{lst:app_generate}{\texttt{generate\_brief.md}, the template of the generate session.}{appendix/prompts/generate_brief.txt}

\applisting{lst:app_improve}{\texttt{improve\_freeform.md}, the template of every improve session.}{appendix/prompts/improve_freeform.txt}

\appboxlisting{lst:app_fixval}{\texttt{fix\_validation.md}, appended after a rejected proposal.}{appendix/prompts/fix_validation.txt}

\appboxlisting{lst:app_fixpre}{\texttt{fix\_preflight.md}, appended after a preflight crash.}{appendix/prompts/fix_preflight.txt}

\subsection{Workspace Instructions}
\label{app:prompt_workspace}

Listing~\ref{lst:app_mechanics} is appended to every brief. It names the workspace files and the one shell command the agent needs, and asks for a closing summary that the loop records as the iteration's rationale; the next report quotes it as unverified (Listing~\ref{lst:app_fb_loop}), next to the harness's own account of the edit. In the workspace of a Fable~5.1 Tool Hanging run of Table~\ref{tab:models}, the files the block names take 126\,KB: the workflow-format primer \texttt{WORKFLOW\_FORMAT.md} (6.7\,KB), the tool contract \texttt{SKILLS.md} (31\,KB), the skill catalog \texttt{SKILL\_LIBRARY.md} (21\,KB), six tool-bundle manuals (30\,KB) and the validator's surface \texttt{skills.lock.json} (37\,KB), next to the 28 shipped skill bundles. Codex sessions also find \texttt{AGENTS.md} (Listing~\ref{lst:app_agents}), which states the rules Claude Code enforces with its deny list (Appendix~\ref{app:guard}).

\applisting{lst:app_mechanics}{The workspace block (\texttt{claude\_code.\_mechanics}); slots as in Table~\ref{tab:app_prompt_slots}.}{appendix/prompts/mechanics.txt}

\appboxlisting{lst:app_agents}{\texttt{AGENTS.md} of a Codex session (GPT-6 Sol, Gear Meshing); absolute paths replaced by their role.}{appendix/prompts/agents_md.txt}

\subsection{Task-Specific Slots}
\label{app:prompt_slots}

Table~\ref{tab:app_prompt_slots} lists every slot and its value. Across tasks, only the instruction, the shipped skills, the Cable Routing bimanual note and the USB-C Insertion and Cable Routing guidance (Listings~\ref{lst:app_extra_usbc} and~\ref{lst:app_extra_cable}) differ. The layout note is in every brief because the reports print the training layouts' world coordinates, which fit no held-out layout if copied into a policy. The bimanual note is set only for Cable Routing, whose expert policy anchors the cable's far end with the support arm, while every motion verb drives arm~0 unless told otherwise.

\begin{table}[p]
  \centering
  \footnotesize
  \renewcommand{\arraystretch}{1.12}
  \begin{tabularx}{\linewidth}{@{}l l >{\raggedright\arraybackslash}X@{}}
    \toprule
    Slot & In & Value in the study \\
    \midrule
    \texttt{\{\{instruction\}\}} & G, I & the task's instruction, verbatim (Table~\ref{tab:app_tasks}) \\
    \texttt{\{\{perception\}\}} & G, I & \texttt{vision} \\
    \texttt{\{\{privileged\}\}} & G, I & \texttt{none} \\
    \texttt{\{\{update\_mode\}\}} & G & \texttt{freeform} \\
    \texttt{\{\{layout\_note\}\}} & G, I & every task: ``- Each training episode is a DIFFERENT layout of this task, and the policy is scored afterwards on ten further layouts it never sees, where the objects (and possibly the fixtures) sit elsewhere. Locate everything that can move with perception at run time: a coordinate read off this feedback fits the one layout it was measured in, and a policy built on such coordinates fails the rest.''\newline The Opus~5, xhigh runs of Table~\ref{tab:models} that train on one layout (Appendix~\ref{app:schedule}): ``- Every training episode replays ONE fixed layout, but the policy is scored afterwards on layouts it never sees, where the objects (and possibly the fixtures) sit elsewhere. Locate everything that can move with perception at run time: a coordinate read off this feedback fits the training layout only, and a policy built on one fails every held-out layout.'' \\
    \texttt{\{\{bimanual\_note\}\}} & G, I & Cable Routing: ``This robot has 2 arms. Every robot.*/motion.* verb takes \texttt{arm\_id} (0 is the work arm, 1 the support arm); an omitted arm\_id drives arm 0. Sequence the arms deliberately — SKILLS.md states each tool's arm semantics.'' Other tasks: empty. \\
    \texttt{\{\{extra\_guidance\}\}} & G, I & USB-C Insertion and Cable Routing: which queries the color-palette matcher behind \texttt{sam3.segment\_text} accepts, and the task's goal clause in words, including that an episode ending with the object still held scores zero (Listings~\ref{lst:app_extra_usbc} and~\ref{lst:app_extra_cable}). Other tasks: empty. \\
    \texttt{\{\{guard\_feedback\}\}} & V & the guard's and validator's violations, e.g.\ ``non-literal ctx.tool(...) call in scripts/grasp/locked\_reach.py:268'' \\
    \texttt{\{\{preflight\_feedback\}\}} & P & what the unscored preflight episode raised, e.g.\ ``- Node 'clear\_sg.clear' failed: TypeError: int() argument must be a string, a bytes-like object or a real number, not 'builtin\_function\_or\_method'\,'' \\
    \midrule
    \texttt{\{\{action\}\}} & M & generate: ``Author the policy into \texttt{workflow/} (create \texttt{workflow/workflow.json} and \texttt{workflow/scripts/...}).'' Improve: ``Edit the policy in \texttt{workflow/} in place. Read \texttt{FEEDBACK.md} first.'' \\
    \texttt{\{\{tool\_bundle\_manuals\}\}} & M & every task: \texttt{curobo.md}, \texttt{curve.md}, \texttt{geometry.md}, \texttt{grounding-dino.md}, \texttt{sam3.md}, \texttt{vlm.md} \\
    \texttt{\{\{shipped\_skills\}\}} & M & full library: the shipped skills' names, i.e.\ the callable general skills $\mathcal{S}_g$ (Table~\ref{tab:app_skills_ors}) and, in Table~\ref{tab:models}, the task's 1--3 expert skills $\mathcal{S}_\tau$ (Appendix~\ref{app:skills}). Tools only and docs callable: this line and the \texttt{SKILL\_LIBRARY.md} line above it become ``\texttt{SKILL\_LIBRARY.md} — what the skill library holds for this run (it may say that it holds nothing).'' \\
    \midrule
    \multicolumn{2}{@{}l}{\texttt{WORKFLOW\_FORMAT.md}, Skills} & full library: as written (Listing~\ref{lst:app_primer_full}); docs callable and tools only: replaced (Listings~\ref{lst:app_primer_listed} and~\ref{lst:app_primer_none}) \\
    \bottomrule
  \end{tabularx}
  \caption{Template slots and their values. In: the template that uses the slot (G generate, I improve, V fix-validation, P fix-preflight, M workspace block). Quoted values are verbatim; the workspace block and the primer are those of every snapshot after the first, whose earlier wording the text quotes.}
  \label{tab:app_prompt_slots}
\end{table}

The extra guidance states two facts no report shows. On these two tasks \texttt{sam3.segment\_text} runs a color matcher that \texttt{sam3.md} does not describe, and its error for other queries blames missing SAM3 weights, which no edit can fix. And reports describe the scene but never the goal clause: in an early USB-C Insertion run, a report showed the plug 12.8\,mm past the bore's mouth, 0.2\,mm off its axis and ``seated in \texttt{port.bore}: yes'' in an episode that scored zero, because the fingers still held it; once the brief stated the goal clause, the next training batch succeeded in all four episodes.

\appboxlisting{lst:app_extra_usbc}{\texttt{\{\{extra\_guidance\}\}}, USB-C Insertion.}{appendix/prompts/extra_guidance_usbc.txt}

\appboxlisting{lst:app_extra_cable}{\texttt{\{\{extra\_guidance\}\}}, Cable Routing.}{appendix/prompts/extra_guidance_cable.txt}

\subsection{Differences Between Study Arms}
\label{app:prompt_arms}

Within a task, every feedback condition receives the same improve brief, byte for byte up to the earlier wording below, so the report is the only input that differs between the rows of Tables~\ref{tab:temporal_feedback} and~\ref{tab:spatial_feedback}. Between the skill-library arms, the brief differs in the shipped-skill lines of the workspace block and in the Skills section of the workflow primer (Listings~\ref{lst:app_primer_full}--\ref{lst:app_primer_none}); the skills themselves are in Appendix~\ref{app:skills}.

\paragraph{First harness snapshot.}
Runs launched on the first harness snapshot read an earlier wording of two passages; these are most runs of the reference configuration (the feedback reference and the full-library arm) and the Opus~5, xhigh runs of Table~\ref{tab:models} that train on one layout. Their workspace block lacks the sentence ``Run it exactly as written, from this directory---an absolute path to the same script is a different command and is refused, which costs a turn.'', and their primer ends the second paragraph of Listing~\ref{lst:app_primer_full} with ``\ldots\ and copy its \texttt{canonical\_scripts} into your \texttt{scripts/} directory.'' The rest of both texts is identical.

\appboxlisting{lst:app_primer_full}{Skills section of \texttt{WORKFLOW\_FORMAT.md}, full library.}{appendix/prompts/primer_skills_full.txt}

\appboxlisting{lst:app_primer_listed}{Skills section of \texttt{WORKFLOW\_FORMAT.md}, docs callable (source withheld).}{appendix/prompts/primer_skills_listed.txt}

\appboxlisting{lst:app_primer_none}{Skills section of \texttt{WORKFLOW\_FORMAT.md}, tools only.}{appendix/prompts/primer_skills_none.txt}

\FloatBarrier
\makeatletter
\@ifundefined{c@applisting}{\newcounter{applisting}}{}
\@ifundefined{app@box}{\newsavebox{\app@box}}{}
\@ifundefined{ifapp@fresh}{\newif\ifapp@fresh}{}
\providecommand{\appneedspace}[1]{\par\begingroup\setlength{\dimen@}{#1}\vskip\z@\@plus\dimen@\penalty-100%
  \vskip\z@\@plus-\dimen@\vskip\dimen@\penalty9999\vskip-\dimen@\vskip\z@skip\endgroup}
\providecommand{\app@listing}[3]{\noindent\refstepcounter{applisting}\label{#1}%
  {Listing~\theapplisting:~#2}\par\nopagebreak\vspace{-1pt}%
  \if\relax\detokenize{#3}\relax\else{\@beginparpenalty\@M\fvset{breaksymbolleft={},breakindent=1em}\listingfile{#3}}\fi}
\providecommand{\appboxed}[2][\medskipamount]{\par\setbox\app@box\hbox{\begin{minipage}[t]{\linewidth}#2\end{minipage}}%
  \appneedspace{\dimexpr\ht\app@box+\dp\app@box+#1\relax}\vskip#1\box\app@box\par}
\providecommand{\applisting}[3]{%
  \if\relax\detokenize{#3}\relax\appneedspace{14\baselineskip}\global\app@freshtrue\else\appneedspace{7\baselineskip}\fi
  \medskip\app@listing{#1}{#2}{#3}}
\providecommand{\appboxlisting}[3]{\appboxed{\app@listing{#1}{#2}{#3}}}
\providecommand{\app@panel}[2]{{\footnotesize\textit{#1}}\par\nopagebreak\vspace{-3pt}%
  {\@beginparpenalty\@M\fvset{breaksymbolleft={},breakindent=1em}\listingfile{#2}}}
\providecommand{\appsublisting}[2]{%
  \ifapp@fresh\global\app@freshfalse\par\smallskip\noindent\begin{minipage}[t]{\linewidth}\app@panel{#1}{#2}\end{minipage}\par
  \else\appboxed[\smallskipamount]{\app@panel{#1}{#2}}\fi}
\makeatother

\section{Feedback Reports}
\label{app:feedback}
\suppressfloats[t]

One renderer turns the recorded training batch into \texttt{FEEDBACK.md}, the report $f_k$ of Section~\ref{sec:protocol} and the only feedback an improve session reads. We define the feedback conditions as settings of that renderer, list what each renders, walk through one report, compare one batch under every condition, and measure report length. The reference condition is the feedback of Table~\ref{tab:models} and of the \emph{Per skill} and \emph{+ part annotations} rows of Tables~\ref{tab:temporal_feedback} and~\ref{tab:spatial_feedback}, which are the same runs.

\subsection{Conditions}
\label{app:feedback_conditions}

Each condition sets three axes of the renderer: the form of the report, its spatial representation and its temporal unit (Table~\ref{tab:app_feedback_conditions}). Every non-reference condition changes one axis of the reference, starts from the reference run's initial policy, and first re-renders the reference's own $k{=}0$ training batch, so that differences between rows reflect the report rather than the initial policy.

\begin{table}[ht]
  \centering
  \footnotesize
  \setlength{\tabcolsep}{3pt}
  \setlength{\aboverulesep}{0pt}
  \setlength{\belowrulesep}{0pt}
  \renewcommand{\arraystretch}{1.15}
  \begin{tabularx}{\linewidth}{@{}>{\raggedright\arraybackslash\hangindent=1em\hangafter=1}X|l|lll@{}}
    \toprule
    Condition & Run arm & Form & Spatial & Granularity \\
    \midrule
    Reference (Table~\ref{tab:models}; \emph{Per skill}, \emph{+ part annotations}) & \texttt{reference} & \texttt{state, relations} & \texttt{parts} & \texttt{semantic} \\
    \midrule
    \multicolumn{1}{@{}l|}{\textit{Spatial representation (Table~\ref{tab:spatial_feedback})}} & & & & \\
    \quad Outcome only & \texttt{success\_only} & \texttt{--} & \texttt{(parts)} & \texttt{(semantic)} \\
    \quad State + relations, world frame & \texttt{positions} & \texttt{state, relations} & \texttt{positions} & \texttt{semantic} \\
    \quad State + relations, object frames & \texttt{frames} & \texttt{state, relations} & \texttt{frames} & \texttt{semantic} \\
    \midrule
    \multicolumn{1}{@{}l|}{\textit{Temporal resolution (Table~\ref{tab:temporal_feedback})}} & & & & \\
    \quad Per episode & \texttt{episode} & \texttt{state, relations} & \texttt{parts} & \texttt{episode} \\
    \quad Per motion & \texttt{motion} & \texttt{state, relations} & \texttt{parts} & \texttt{motion} \\
    \midrule
    \multicolumn{1}{@{}l|}{\textit{Single-form reports}} & & & & \\
    \quad Outcome + stills & \texttt{visual} & \texttt{visual} & \texttt{(parts)} & \texttt{semantic} \\
    \quad Outcome + object state & \texttt{state} & \texttt{state} & \texttt{parts} & \texttt{semantic} \\
    \quad Outcome + relations & \texttt{relations} & \texttt{relations} & \texttt{parts} & \texttt{semantic} \\
    \bottomrule
  \end{tabularx}
  \caption{The feedback conditions as configuration (\texttt{feedback.design}), grouped by the table that reports them. Every non-reference condition changes one axis of the reference and continues from the reference run's initial policy. Parenthesized values select nothing: the spatial representation applies to state and relations, the granularity to any form beyond the outcome. Run arm: the condition's name in the released configurations.}
  \label{tab:app_feedback_conditions}
\end{table}

\subsection{What Each Condition Renders}
\label{app:feedback_gating}

Every condition renders the outcome block: per-episode and per-object success, steps, how each episode ended, and the errors the policy raised. Table~\ref{tab:app_feedback_gating} lists what each condition adds. Three consequences matter for reading the main tables. Outcome only is not a scalar reward: it names the node where each aborted episode raised and quotes the error, but not why a finished episode missed. A per-motion report has no unit for a step that moves nothing, so what a perception step returns never appears, the gap Finding~3 attributes to motion-level feedback. Syringe Packing declares no parts, so there the reference and object-frame reports differ only in wording, and the two rows end at the same 50\% in Table~\ref{tab:spatial_feedback}. On Cable Routing the contact record omits the deformable cable, so no condition reports contact with the cable or a grasp or carry of it, and no close is reported as meeting the cable.

\begin{table}[ht]
  \centering
  \footnotesize
  \setlength{\tabcolsep}{2.5pt}
  \setlength{\aboverulesep}{0pt}
  \setlength{\belowrulesep}{0pt}
  \renewcommand{\arraystretch}{1.15}
  \begin{tabularx}{\linewidth}{@{}>{\raggedright\arraybackslash}X|c|cccc|cc|cc@{}}
    \toprule
    & & \multicolumn{4}{c|}{Form} & \multicolumn{2}{c|}{Spatial} & \multicolumn{2}{c}{Temporal} \\
    Rendered & {\scriptsize\shortstack{Refer-\\ence}} & {\scriptsize\shortstack{Outcome\\only}} & {\scriptsize\shortstack{Outcome\\+ stills}} & {\scriptsize\shortstack{Outcome\\+ state}} & {\scriptsize\shortstack{Outcome\\+ relations}} & {\scriptsize\shortstack{World\\frame}} & {\scriptsize\shortstack{Object\\frames}} & {\scriptsize\shortstack{Per\\episode}} & {\scriptsize\shortstack{Per\\motion}} \\
    \midrule
    Outcome: episode table, per-object goals, abort, budget and stage banners, raised errors & \checkmark & \checkmark & \checkmark & \checkmark & \checkmark & \checkmark & \checkmark & \checkmark & \checkmark \\
    Node names (one unit per node, with status and time) & \checkmark & $\circ^a$ & $\circ^b$ & \checkmark & \checkmark & \checkmark & \checkmark & $\circ^a$ & $\circ^a$ \\
    Up to 16 stills (two cameras) around the nodes nearest the first failure & -- & -- & \checkmark & -- & -- & -- & -- & -- & -- \\
    Tool names, commanded vs.\ achieved targets & \checkmark & -- & -- & \checkmark & -- & \checkmark & \checkmark & -- & \checkmark \\
    Node outputs (\texttt{returned key=value}) & \checkmark & -- & -- & \checkmark & -- & $\circ^c$ & \checkmark & -- & -- \\
    World positions of objects and hands, moves & \checkmark & -- & -- & \checkmark & -- & \checkmark & \checkmark & \checkmark & \checkmark \\
    Orientations, own-axes offsets, poses in the target's frame & \checkmark & -- & -- & \checkmark & -- & -- & \checkmark & \checkmark & \checkmark \\
    Declared parts and part geometry & \checkmark & -- & -- & \checkmark & $\circ^d$ & -- & -- & \checkmark & \checkmark \\
    Relation predicates (rests on, lies flat, encloses: yes/no, goal met) & \checkmark & -- & -- & -- & \checkmark & \checkmark & \checkmark & \checkmark & \checkmark \\
    Contact and grasp events (came into contact, let go, what the close met) & \checkmark & -- & -- & -- & \checkmark & \checkmark & \checkmark & $\circ^e$ & \checkmark \\
    History ($k\geq1$): goal distances (median at the end, closest) & \checkmark & -- & -- & \checkmark & -- & $\circ^f$ & $\circ^f$ & $\circ^g$ & \checkmark \\
    History ($k\geq1$): reached-then-lost, held at the end, grasp counts & \checkmark & -- & -- & -- & \checkmark & \checkmark & \checkmark & $\circ^h$ & \checkmark \\
    \bottomrule
  \end{tabularx}
  \caption{What each feedback condition renders (\checkmark\ whenever the batch has the item, $\circ$ partly, -- never): the reference, then the conditions that change its form, its spatial representation or its temporal granularity (Table~\ref{tab:app_feedback_conditions}). $^a$only the node the graph raised at, in the outcome block; $^b$only the nodes that carry stills; $^c$orientation-valued outputs dropped; $^d$part names and kinds only, no geometry; $^e$events between start and end only, not what the close met; $^f$object-level (containment) goals only; $^g$median distance at the end only; $^h$held at the end only. The stills arm's first report has none: the reference batch it re-renders was recorded without video. Syringe Packing declares no parts, so there the reference differs from object frames only in its legend and goal lines.}
  \label{tab:app_feedback_gating}
\end{table}

\subsection{Report Anatomy}
\label{app:feedback_anatomy}

Listing~\ref{lst:app_fb_skeleton} is the opening of the reference report for the four training episodes that the first Tool Hanging reference run scored at $k{=}0$: the task line, the legend naming the condition, the training result, the space conventions and declared parts, the episode table and the banners. One representative episode follows per failure signature (the first failing node, how the episode ended, the goals met), at most six, each followed by the other episodes that failed the same way, so a batch that fails one way costs one episode's text.

\applisting{lst:app_fb_skeleton}{Opening of the reference report, up to the first representative episode.}{appendix/feedback/skeleton.txt}

\paragraph{Loop notice and history.}
From $k{=}1$ the report summarizes every earlier iterate: its training result, its edit, what its policy did, derived from its own recorded run, and the previous session's own account of its change, marked as unverified. When the training batch regresses under the revert rule (Appendix~\ref{app:revert}), the loop restores the earlier iterate's policy and the report opens with a notice, because the policy the agent now edits is no longer the one the report describes (Listing~\ref{lst:app_fb_loop}). Here the reverted iterate brought the wrench closer to the goal (median 0.067\,m), but every episode aborted when the planner refused the waypoint engaging the hook (Finding~5), and the rule always counts such a batch as a regression.

\appboxlisting{lst:app_fb_loop}{Loop notice and history block of the reference report at $k{=}1$; the quoted account is cut after its first line.}{appendix/feedback/loop_notice.txt}

\subsection{One Batch Under Every Condition}
\label{app:feedback_compare}

Since every condition re-renders the same $k{=}0$ batch, the reports compare line by line. Listing~\ref{lst:app_fb_close} shows the node \texttt{grasp\_sg.close} of Episode~2 under each condition that renders it, Listing~\ref{lst:app_fb_end} the end of Episode~0 under the three spatial representations, and Listing~\ref{lst:app_fb_episode} the per-episode report's single unit for that episode.

\applisting{lst:app_fb_close}{Node \texttt{grasp\_sg.close}, Episode~2, by condition.}{}
\appsublisting{(a) Reference (state + relations, parts, per node)}{appendix/feedback/close_reference.txt}
\appsublisting{(b) Outcome + object state}{appendix/feedback/close_state.txt}
\appsublisting{(c) Outcome + relations}{appendix/feedback/close_relations.txt}
\appsublisting{(d) State + relations, world frame}{appendix/feedback/close_positions.txt}
\appsublisting{(e) State + relations, object frames}{appendix/feedback/close_frames.txt}
\appsublisting{(f) Per motion: the same close as a motion unit}{appendix/feedback/close_motion.txt}

\applisting{lst:app_fb_end}{End of Episode~0 by spatial representation.}{}
\appsublisting{(a) World frame}{appendix/feedback/end_ep0_positions.txt}
\appsublisting{(b) Object frames}{appendix/feedback/end_ep0_frames.txt}
\appsublisting{(c) Part annotations (reference)}{appendix/feedback/end_ep0_reference.txt}

\appboxlisting{lst:app_fb_episode}{Episode~0 in the per-episode report.}{appendix/feedback/episode_ep0.txt}

\paragraph{Reading the listings side by side.}
Every report with relations says that the wrench never moved with the hand after the close; the state-only report gives the numbers without that verdict. The world-frame report gives the jaws' offset from the wrench's center as a world vector (0.123\,m), which is an offset along the handle only because this wrench lies along world $x$, a fact that report never states; object frames give the offset directly (+122\,mm along the long axis), and part annotations add that the jaws closed 48\,mm from the ring, the part the goal is written in. At the end of Episode~0, likewise, only part annotations relate the ring to the shank. The world-frame report holds the information but leaves the transform to the agent, the step Finding~4 finds unreliable. The per-episode report has no unit for the close, so a failure caused there is attributed to what the end state shows, and the per-motion report files the close under Motion~4, \texttt{robot.close\_gripper}, which the agent must map back to its own graph step (Finding~3).

\subsection{Report Size and Truncation}
\label{app:feedback_size}

\begin{figure}[ht]
  \centering
  \begin{minipage}[c]{0.56\linewidth}
    \includegraphics[width=\linewidth]{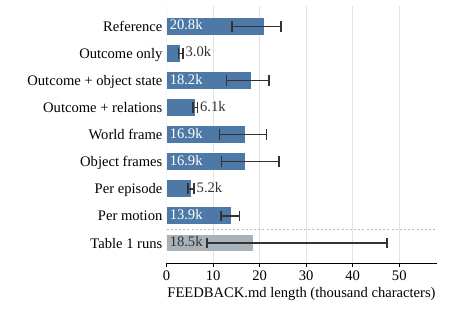}
  \end{minipage}\hfill
  \begin{minipage}[c]{0.41\linewidth}
    \caption{Length of the feedback report per condition, in characters: median (bar) and interquartile range (whiskers) over the distinct reports the improve sessions read. The gray bar pools every run of Table~\ref{tab:models}, all of which use the reference condition.}
    \label{fig:app_packet_size}
  \end{minipage}
\end{figure}

A report shows at most 12 outputs per node, each value cut to 60 characters, and at most 40 units per episode (the first and last four and those around the failing unit). Node errors are cut to 300 characters, a graph traceback to its last 1{,}500 characters, and the previous session's account to 800 characters. The unit cap almost never binds (fewer than 1\% of the reports behind Figure~\ref{fig:app_packet_size}, all of them per-motion reports). Figure~\ref{fig:app_packet_size} shows how long the reports are: outcome-only reports stay near 3k characters, while state numbers and part geometry bring the reference to 21k, and Syringe Packing reports of Table~\ref{tab:models} run to a median of 70k.

\paragraph{More text is not better.}
Length does not predict learning. World-frame and object-frame reports have the same median length (16.9k characters), yet reach 30\% and 40\% overall held-out success (25\% and 63\% on Tool Hanging). The state-only report is nearly as long as the reference (18.2k against 20.8k) yet ends below outcome only at 3.0k, 11\% against 19\% overall. The per-motion report is a third shorter than the reference yet reaches 90\% against 45\% on Tool Hanging. Across the eight conditions, the rank correlation between median length and overall held-out success is 0.24: what helps is which quantities a report makes explicit and at which unit (Findings~3 and~4), not how much it says.

\FloatBarrier
\section{Revert, Stopping and Selection Rules}
\label{app:stopping}

Three rules decide which policy the loop carries and which one a run reports: the revert rule chooses the policy the
next session edits, the stopping conditions end the run, and the selection rule picks the kept iterate. We give each
rule and what it did in the runs, then relate the @$k$ of Table~\ref{tab:models} to the kept iterate.
Table~\ref{tab:app_stopping} holds the thresholds and the per-task rates.

\subsection{Revert Rule}
\label{app:revert}
The revert rule compares each training batch with the incumbent, the best iterate so far, lexicographically on
per-object success rate, failed episodes that reached the goal geometry, median closest approach to the goal, and
grasps that carried the object, each with a hysteresis margin (Table~\ref{tab:app_stopping}; 3 episodes, 20\,mm,
3 carries); a lower key never overrules the direction of the success rate, and a batch in which every episode
aborted is always a regression. On a regression the next session edits the incumbent instead of the latest iterate,
and its report opens with a notice naming the undone edits (Listing~\ref{lst:app_fb_loop}). We add the rule because
in pilot runs sessions told that their last edit had regressed still built on it. The rule reads only training
episodes. It fires after 18\% of edited iterates (42\% of these on
success rate, 35\% on closest approach, 21\% on all-abort batches), most often where the margin is smallest relative
to the batch: after 37\% of Syringe Packing and 19\% of Cable Routing edits (margins of 2/8 and 3/12 objects), against
5--15\% on the one-object tasks.

\begin{table}[ht]
  \centering
  \setlength{\tabcolsep}{3.5pt}
  \setlength{\aboverulesep}{0pt}
  \setlength{\belowrulesep}{0pt}
  \renewcommand{\arraystretch}{1.15}
  \resizebox{\linewidth}{!}{%
  \begin{tabular}{l|ccccc}
    \toprule
    & Tool Hanging & Syringe Packing & Gear Meshing & USB-C Insertion & Cable Routing \\
    \midrule
    Validation target (success) & 1 & 0.83 & 1 & 1 & 0.67 \\
    \quad in objects & 5/5 & 9/10 & 5/5 & 5/5 & 11/15 \\
    Revert margin (train success) & 0.5 & 0.25 & 0.5 & 0.5 & 0.25 \\
    \quad in objects & 2/4 & 2/8 & 2/4 & 2/4 & 3/12 \\
    Episode wall budget (s) & 1500\,+\,4200 & 1500\,+\,12000 & 1500\,+\,7200 & 1500\,+\,7200 & 1500\,+\,14000 \\
    \midrule
    Unchanged streak & \multicolumn{5}{c}{3 iterations without an accepted edit} \\
    Spend per run & \multicolumn{5}{c}{\$60} \\
    Wall time per run & \multicolumn{5}{c}{120\,h (most $K{=}6$ runs: 30\,h)} \\
    Fix rounds & \multicolumn{5}{c}{2 after a guard rejection, 2 after a preflight crash} \\
    Preflight & \multicolumn{5}{c}{1 unscored episode, 1{,}500 steps} \\
    \midrule
    \multicolumn{6}{l}{\textit{Stop reason (\% of runs)}} \\
    \quad $K{=}3$, iteration cap & 84 & 100 & 8 & 69 & 4 \\ 
    \quad $K{=}3$, validation target & 16 & 0 & 92 & 31 & 96 \\ 
    \quad $K{=}6$, iteration cap & 81 & 100 & 40 & 81 & 79 \\ 
    \quad $K{=}6$, validation target & 15 & 0 & 55 & 19 & 21 \\ 
    \quad $K{=}6$, perfect training streak & 4 & 0 & 5 & 0 & 0 \\ 
    \multicolumn{6}{l}{\textit{Kept iterate is $\pi_0$ (\% of runs)}} \\
    \quad $K{=}3$ & 68 & 54 & 88 & 62 & 81 \\ 
    \quad $K{=}6$ & 26 & 29 & 30 & 62 & 21 \\ 
    \textit{Reverts (\% of edited iterates)} & 15 & 37 & 9 & 5 & 19 \\ 
    \bottomrule
  \end{tabular}}
  \caption{Stopping conditions per task and common to all tasks, and what they did: how the runs stopped, how often they kept the zero-shot program $\pi_0$, and how often the revert rule undid an edit. Success rates are per object: the validation target over the five validation instances, the revert margin over the four training instances.}
  \label{tab:app_stopping}
\end{table}

\subsection{Stopping Conditions}
\label{app:stop_conditions}
After each validation batch the harness checks, in order, the iteration cap, the validation target, the
perfect-training streak (only in the runs without a validation set, Appendix~\ref{app:schedule}), the unchanged streak,
the spend and the wall time, and stops at the first condition that holds (Table~\ref{tab:app_stopping}); a target met
in the last iteration is recorded under the cap. The validation target replaces the first passes' stop on two perfect
training batches (Appendix~\ref{app:first_passes}), since the agent edits against the training instances. The spend and wall time are runaway ceilings, not
budgets; the unchanged streak and the spend never end a run. Most 6-iteration runs ran with a 30\,h wall-time limit,
and part of them with shorter episode budgets (1{,}400, 6{,}000, 2{,}400, 1{,}200--2{,}400 and 9{,}000\,s in task
order).

\paragraph{How the runs stop.}
Of the 3-iteration runs, 54\% stop at the cap and 46\% on the target, 36\% of all runs already at $k{=}0$. Nearly
every Gear Meshing and Cable Routing run meets its target (92\% and 96\%), mostly with $\pi_0$, and carries that
reading forward, so these columns of Table~\ref{tab:models} largely measure the zero-shot program. Syringe Packing
never meets its target, so every Syringe run uses all its rounds. The 6-iteration runs, most without the expert
skills, meet the target less often and later: 76\% stop at the cap, 22\% on the target (5\% at $k{=}0$) and 2\% after
two perfect training batches.

\subsection{Kept Iterate}
\label{app:kept}
The kept iterate is the first iterate with the highest validation success, so an edit is kept only if it raises
validation success. It is $\pi_0$ in 70\% of the 3-iteration runs, for three reasons that split by task
(Table~\ref{tab:app_stopping}). Half of these runs meet the target with $\pi_0$, mostly on Gear Meshing and Cable
Routing. A third never succeed on a validation instance, so every iterate ties at zero: on Tool Hanging no run of
DeepSeek~V4, Kimi~K3 or either GPT-6 setting ever validates, nor any Kimi~K3 run on USB-C Insertion, the repair
failure of Finding~1 at its sharpest. In the remaining 15\%, almost all on Syringe Packing, $\pi_0$ validates at least
as well as both edits. The 6-iteration runs keep $\pi_0$ in 34\% of runs, as they rarely stop at $k{=}0$ and have five edits to improve on it.
The runs without a validation set keep the revert rule's incumbent, and one skill-library run, resumed to its iteration
cap after a wall-time stop, keeps the iterate chosen at that stop.

\subsection{Superseded First Passes}
\label{app:first_passes}
Three passes preceded the protocol of Appendix~\ref{app:harness}. The first stopped after two perfect training batches and kept the iterate that trained best, and its first launch shipped the expert scripts with the reference library, so the agents copied the human solution; two of its runs enter the Opus~5, xhigh row of Table~\ref{tab:models} (Appendix~\ref{app:schedule}), and its Tool Hanging run stopped with 70\% held-out success. The second trained on four distinct instances but still stopped and selected on them; a brief third validated on three. Since the report describes exactly the training episodes, selecting on them rewards edits fitted to those layouts, so we moved stopping and selection to five validation instances the agent never sees and, in the feedback and skill-library studies, ship $\mathcal{S}_\tau$ only to the $+\,\mathcal{S}_\tau$ arm.

\FloatBarrier
\section{Models and Inference}
\label{app:models}

This appendix documents the ten agent settings of Table~\ref{tab:models}: models and access
(Appendix~\ref{app:models_access}), call parameters (Appendix~\ref{app:models_inference}), reasoning controls and
realized thinking (Appendix~\ref{app:models_reasoning}), and session statistics (Appendix~\ref{app:models_sessions}),
the last two bearing on Finding~1. Appendix~\ref{app:cost} gives the compute and cost of all three studies.

\subsection{Models and Access}
\label{app:models_access}

Table~\ref{tab:app_models} identifies each setting by its model id, the date the id was first listed, and the
dates its sessions ran.
The Claude models run in the Claude Code CLI on a subscription login. Kimi~K3 and DeepSeek~V4 run in the same CLI
against OpenRouter's Anthropic-compatible endpoint, which keeps the prompt, tools, allowed commands and turn cap of
the Claude settings and changes only the model. GPT-6 Astra and Sol run in OpenAI's Codex CLI~0.155.1, which reads
the same instruction file plus an \texttt{AGENTS.md} stating the workspace rules (Appendix~\ref{app:prompts}). The feedback and skill-library studies use the command and model of Opus~5, xhigh.

\begin{table}[ht]
  \centering
  \footnotesize
  \setlength{\tabcolsep}{3pt}
  \setlength{\aboverulesep}{0pt}
  \setlength{\belowrulesep}{0pt}
  \renewcommand{\arraystretch}{1.15}
  \begin{tabular}{l|>{\raggedright\arraybackslash}p{12.2em}|c|c|l|c|cc}
    \toprule
    & & & Sessions & Harness & Effort & \multicolumn{2}{c}{Thinking tokens} \\
    Setting & Model id & Listed & (UTC) & control & xhigh & Generate & Improve \\
    \midrule
    Opus 5, xhigh & \texttt{claude-opus-5} & 07-24 & 09-18--22 & -- & 64\% & 30.8k & 16.5k \\
    Opus 5, high & \texttt{claude-opus-5} & 07-24 & 09-22--24 & budget 32k & 13\% & 17.8k & 21.2k \\
    Opus 5, low & \texttt{claude-opus-5} & 07-24 & 09-22--25 & budget 4k & 12\% & 18.6k & 15.4k \\
    \midrule
    DeepSeek V4 & \texttt{deepseek/}\allowbreak\texttt{deepseek-v4-pro-0813} & 08-12 & 09-22--25 & -- & 22\% & 75.1k & 69.2k \\
    Sonnet 5 & \texttt{claude-sonnet-5} & 06-30 & 09-22--25 & -- & 21\% & 46.2k & 20.2k \\
    Kimi K3 & \texttt{moonshotai/}\allowbreak\texttt{kimi-k3} & 07-16 & 09-22--25 & -- & 12\% & 30.4k & 18.8k \\
    Fable 5.1 & \texttt{claude-fable-5-1} & 09-01 & 09-22--25 & -- & 7\% & 22.9k & 27.1k \\
    GPT-6 Astra & \texttt{gpt-6-astra} & 09-04 & 09-23--25 & effort xhigh & 100\% & 3.4k & 1.6k \\
    Opus 5.5 & \texttt{claude-opus-5-5} & 09-22 & 09-22--25 & -- & 21\% & 10.3k & 6.0k \\
    GPT-6 Sol & \texttt{gpt-6-sol} & 09-22 & 09-23--25 & effort xhigh & 100\% & 5.1k & 3.2k \\
    \bottomrule
  \end{tabular}
  \caption{Agent settings of Table~\ref{tab:models} (Opus~5, xhigh: rounds $k\le2$). Listed: first OpenRouter listing of the model id in 2026 (the DeepSeek~V4 family: 04-24). Sessions: first and last session date. Harness control: the reasoning setting the harness passes (Table~\ref{tab:app_inference}). Effort xhigh: share of sessions whose CLI logged effort level xhigh. Thinking tokens: median per session as the CLI reports them.}
  \label{tab:app_models}
\end{table}

\subsection{Inference Parameters}
\label{app:models_inference}

Table~\ref{tab:app_inference} gives the exact call of each CLI and every parameter the harness controls. We pass
only what the CLIs expose: sampling stays at the provider defaults, every session has a 2-hour wall clock, and Claude
Code caps sessions at 80 turns. From 2026-09-22, a rate-limited Claude session is rerun at once with the same model
through OpenRouter, so a rate limit can change a session's route but never its model or prompt.

\begin{table}[ht]
  \centering
  \footnotesize
  \renewcommand{\arraystretch}{1.15}
  \begin{tabular}{@{}>{\raggedright\arraybackslash}p{0.17\linewidth}>{\raggedright\arraybackslash}p{0.79\linewidth}@{}}
    \toprule
    Claude Code call & \Verb[breaklines,breakafter={,},breakaftersymbolpre={}]|claude -p "Follow INSTRUCTIONS.md in this directory." --output-format json --max-turns 80 --allowedTools "Read,Edit,Write,Grep,Glob,Bash(./validate.sh:*),Bash(echo:*),Bash(mkdir:*)" --model <id>|; Kimi~K3 and DeepSeek~V4 set \texttt{ANTHROPIC\_BASE\_URL} to OpenRouter's Anthropic-compatible endpoint. \\
    \midrule
    Codex call & \Verb[breaklines,breakafter={,},breakaftersymbolpre={}]|codex exec --json --skip-git-repo-check -s danger-full-access -C <workspace> -o <transcript>/last_message.txt -m <id> -c model_reasoning_effort="xhigh" "Follow INSTRUCTIONS.md in this directory. AGENTS.md states the rules."| \\
    \midrule
    Reasoning & Opus~5, high and low add \texttt{MAX\_THINKING\_TOKENS}=32000/4000 to the environment; Codex gets \texttt{model\_reasoning\_effort} in the call. Claude Code's effort level is not set: the local machine's CLI settings pin \texttt{effortLevel} to xhigh, and the remote machines did not record it. \\
    \midrule
    Session limits & 2\,h wall clock per session; turn cap \texttt{--max-turns 80} for Claude Code, none for Codex. \\
    \midrule
    Sampling & temperature and top-$p$ not set (provider default). \\
    \midrule
    Context and output & as the CLI reports them. Context: 1M tokens for the Claude models, 200k for Kimi~K3, DeepSeek~V4 and the sessions rerun through OpenRouter (the CLI's default for a model outside its catalog). Maximum output per response: 64k or 128k (Opus~5, Sonnet~5), 128k (Opus~5.5), 64k (Fable~5.1, the reruns), 32k (Kimi~K3, DeepSeek~V4). Codex reports neither. \\
    \midrule
    Rate limits & from 2026-09-22, a Claude session refused with HTTP 429 on the subscription login is rerun at once with the same model through OpenRouter; before that, and for Kimi~K3 and DeepSeek~V4, it is retried after the limit resets when that is within 1\,h (at most 4 attempts). A later reset, or a Codex usage limit, defers the round, and the run resumes it. \\
    \bottomrule
  \end{tabular}
  \caption{Inference parameters of every agent session of the VABAR studies (the curriculum study's are in Appendix~\ref{sec:app-curriculum}). \texttt{<id>} is the model id of Table~\ref{tab:app_models}; the working directory is the session's workspace.}
  \label{tab:app_inference}
\end{table}

\subsection{Reasoning Budget}
\label{app:models_reasoning}

\paragraph{Controls and logged effort.}
The harness sets a reasoning control only for Opus~5, high and low (thinking budgets of 32k and 4k tokens per
request) and the two GPT-6 settings (Codex effort xhigh; Table~\ref{tab:app_inference}). Claude Code's effort level
is left to each machine's CLI settings. Where the logs record it, it is xhigh, except for 13\% of the Opus~5, xhigh
sessions (high, all on 2026-09-18) and most locally run Fable~5.1 sessions (medium); the remote machines record no
level. Table~\ref{tab:app_models} reports the sessions logged at xhigh per setting. In the feedback and
skill-library studies these are the 26\% of sessions run on the local machine; the rest ran on the remote machines.

\paragraph{Realized thinking.}
The 32k and 4k budgets yield indistinguishable sessions (Table~\ref{tab:app_models}): median thinking of 17.8k and
18.6k tokens for initial programs and 21.2k and 15.4k for feedback rounds.
The budget caps a single request, which rarely approaches it: Opus~5 thinks a median of 420 tokens per turn under
the 32k budget and 387 under the 4k one. We therefore read the three Opus~5 rows as three samples of one model, not a
sweep of reasoning effort (the xhigh row sets no budget and comes from 6-iteration runs, Appendix~\ref{app:schedule}).
Read this way, they support Finding~1: the three rows start further apart (41--57\%) than the six settings of
DeepSeek~V4, Sonnet~5, Kimi~K3, Fable~5.1 and GPT-6 (33--45\%), yet all three reach 74--81\% at @2, far above the
44--48\% at which DeepSeek~V4, Sonnet~5, Kimi~K3 and GPT-6 Astra end.

\subsection{Agent Sessions}
\label{app:models_sessions}

Table~\ref{tab:app_sessions} summarizes the agent sessions. The median session lasts 3.0 to 11.8 minutes
(DeepSeek~V4: 29.0), far below the 2-hour limit, and takes 31--47 turns in Claude Code (Codex: 22 and 28 tool
actions).

\begin{table}[ht]
  \centering
  \footnotesize
  \setlength{\tabcolsep}{4pt}
  \setlength{\aboverulesep}{0pt}
  \setlength{\belowrulesep}{0pt}
  \renewcommand{\arraystretch}{1.15}
  \begin{tabular}{l|c|cc|c|c}
    \toprule
    & Fix & \multicolumn{2}{c|}{Per session} & Lines & Gain \\
    Setting & sessions & Turns & Minutes & per edit & @0$\to$@2 \\
    \midrule
    Opus 5, xhigh & 7\% & 37 & 8.1 & 113 & $+29$ \\  
    Opus 5, high & 10\% & 42 & 8.1 & 156 & $+40$ \\  
    Opus 5, low & 14\% & 38 & 8.5 & 150 & $+17$ \\  
    \midrule
    DeepSeek V4 & 4\% & 47 & 29.0 & 30 & $+6$ \\  
    Sonnet 5 & 6\% & 31 & 8.7 & 27 & $+1$ \\  
    Kimi K3 & 21\% & 36 & 11.8 & 88 & $+7$ \\  
    Fable 5.1 & 6\% & 44 & 11.8 & 196 & $+20$ \\  
    GPT-6 Astra & 0\% & 22 & 4.7 & 63 & $+5$ \\  
    Opus 5.5 & 14\% & 37 & 6.4 & 54 & $+10$ \\  
    GPT-6 Sol & 32\% & 28 & 3.0 & 35 & $+20$ \\  
    \bottomrule
  \end{tabular}
  \caption{Agent sessions of the settings of Table~\ref{tab:models} (Opus~5, xhigh: rounds $k\le2$): the share of fix sessions (repairing a rejected proposal or a crashing preflight), median turns (Codex: commands and file edits), minutes and lines added plus removed per accepted edit, and the change in Overall success from @0 to @2 in Table~\ref{tab:models}.}
  \label{tab:app_sessions}
\end{table}

\paragraph{Turns and thinking do not track repair.}
DeepSeek~V4 takes the most turns, the longest sessions and the most thinking (69.2k tokens per feedback round), yet
gains 6 points, while GPT-6 Sol gains 20 in 3-minute sessions that think 3.2k. Across the ten settings, the gain's
rank correlation with median turns ($\rho=0.37$; 0.35 over the eight Claude Code settings alone), minutes
($\rho=-0.22$) and thinking per feedback round ($\rho=0.09$) is weak; edit size tracks it ($\rho=0.69$). The
Opus~5 rows and Fable~5.1 change 113--196 lines per accepted edit and gain 17--40 points; Sonnet~5 and DeepSeek~V4
change 27--30 lines and gain 1--6. Repair is thus not bought with more turns or thinking (Finding~1), and the models
that repair little adjust their programs only slightly (Finding~2). Opus~5.5 edits little but starts highest (67\% at @0). The exception is Sol, whose whole gain is USB-C
Insertion (0\% to 100\%), reached with edits of 6--29 lines.

\paragraph{Fix sessions.}
Most settings spend at most 14\% of their sessions repairing a proposal the guard or validator rejected or a program
that crashed in preflight (Appendix~\ref{app:guard}). Kimi~K3 (21\%) and GPT-6 Sol (32\%) spend the most; in both,
the largest single source is one Gear Meshing run whose graph aborts before its first node runs. Kimi's is repaired
in its zero-shot round and scores 90\%, Sol's never passes the preflight and scores 0\% in every round. Producing a
program that loads is thus a hurdle separate from the behavioral repair that the gain measures.

\FloatBarrier
\section{Compute and Cost}
\label{app:cost}

This appendix accounts for the study's compute: the coding-agent sessions that write and edit the policies
(Appendix~\ref{app:cost_agent}), the simulation that executes them (Appendix~\ref{app:cost_sim}), and the total LLM
spend with the pilot and curriculum studies (Appendix~\ref{app:cost_total}).
Table~\ref{tab:app_cost} counts the runs, agent sessions, tokens, cost and simulation of every setting of
Table~\ref{tab:models} and of each study, with superseded passes and dropped arms in a row of their own. Neither
Table~\ref{tab:app_cost} nor the study and pilot rows of Table~\ref{tab:app_spend} include the API calls that
perception tools make inside the policies.

\begin{table}[ht]
  \centering
  \footnotesize
  \setlength{\tabcolsep}{3.5pt}
  \setlength{\aboverulesep}{0pt}
  \setlength{\belowrulesep}{0pt}
  \renewcommand{\arraystretch}{1.15}
  \begin{tabular}{l|r|r|rr|r|rr}
    \toprule
    & & Agent & \multicolumn{2}{c|}{Tokens (M)} & Cost & Simulator & Policy \\
    Setting & Runs & sessions & Input & Output & (USD) & hours & episodes \\
    \midrule
    Opus 5, xhigh & 25 & 62 & 194 & 3.4 & 292 & 669 & 1{,}000 \\  
    Opus 5, high & 25 & 47 & 138 & 2.3 & 193 & 520 & 718 \\  
    Opus 5, low & 25 & 49 & 148 & 2.3 & 199 & 641 & 714 \\  
    \midrule
    DeepSeek V4 & 25 & 56 & 165 & 6.0 & 33 & 599 & 996 \\  
    Sonnet 5 & 25 & 56 & 199 & 3.7 & 112 & 806 & 978 \\  
    Kimi K3 & 25 & 78 & 197 & 3.4 & 198 & 828 & 1{,}083 \\  
    Fable 5.1 & 25 & 59 & 97.9 & 3.6 & 417 & 1{,}045 & 1{,}049 \\  
    GPT-6 Astra & 25 & 55 & 53.1 & 0.4 & 115 & 550 & 998 \\  
    Opus 5.5 & 25 & 49 & 107 & 1.8 & 108 & 932 & 772 \\  
    GPT-6 Sol & 25 & 88 & 71.1 & 0.7 & 33 & 562 & 1{,}140 \\  
    \midrule
    Table~\ref{tab:models} & 250 & 600 & 1{,}369 & 27.8 & 1{,}700 & 7{,}152 & 9{,}447 \\  
    Feedback study & 200 & 957 & 2{,}519 & 36.0 & 3{,}187 & 7{,}166 & 10{,}098 \\  
    Skill library & 50 & 325 & 428 & 10.2 & 747 & 1{,}563 & 3{,}212 \\  
    Superseded passes and dropped arms & 131 & 576 & 1{,}459 & 25.7 & 1{,}923 & 2{,}030 & 4{,}862 \\  
    \midrule
    Total & 631 & 2{,}459 & 5{,}775 & 99.7 & 7{,}557 & 17{,}910 & 27{,}620 \\
    \bottomrule
  \end{tabular}
  \caption{Compute and cost per setting of Table~\ref{tab:models} and per study. Agent sessions are the coding-agent calls (initial program, feedback rounds, preflight repairs); input tokens include cache reads and writes; cost is repriced at the rates of Table~\ref{tab:app_prices}; simulator hours sum the wall time of every train, validation and held-out episode. Opus~5, xhigh counts rounds $k\le2$ of its six-round runs and the row of superseded passes and dropped arms its later rounds; the skill library's full-library arm is the feedback study's reference.}
  \label{tab:app_cost}
\end{table}

\subsection{Agent Cost}
\label{app:cost_agent}

\paragraph{Pricing.}
Cost is repriced from each session's tokens at the rates of Table~\ref{tab:app_prices}, since Claude Code records
Kimi~K3 and DeepSeek~V4 at Claude rates and the Codex CLI records no cost.

\paragraph{What drives cost.}
Input tokens, mostly cache reads of the session's growing context, outnumber output tokens more than twenty-fold in
every setting, and cost follows the rates more than the token counts: Fable~5.1 reads the fewest input tokens of the
Claude Code settings but costs the most, at twice Opus~5's input, cache-write and output rates, while DeepSeek~V4
writes the most output tokens but is the cheapest Claude Code setting, at under a tenth of Opus~5's rates.
Repairs of rejected proposals and crashing preflights add sessions but no feedback rounds; Kimi~K3 and GPT-6~Sol
spend 21\% and 32\% of their sessions on them (Table~\ref{tab:app_sessions}).
Spend does not track success: Opus~5, high and Opus~5.5 each cost less than half as much as Fable~5.1 and end above
it at @2 in Table~\ref{tab:models} (81\% and 77\% against 64\%).

\begin{table}[ht]
  \centering
  \footnotesize
  \setlength{\tabcolsep}{3.5pt}
  \setlength{\aboverulesep}{0pt}
  \setlength{\belowrulesep}{0pt}
  \renewcommand{\arraystretch}{1.15}
  \begin{tabular}{l|rrrr}
    \toprule
    Model & Input & Cache read & Cache write & Output \\
    \midrule
    Opus 5 & 5 & 0.5 & 10 & 25 \\
    Opus 5.5 & 4 & 0.2 & 8 & 20 \\
    Sonnet 5 & 2 & 0.2 & 4 & 10 \\
    Fable 5.1 & 10 & 0.25 & 20 & 50 \\
    Haiku 4.5 (CLI background model) & 1 & 0.1 & 1.25 & 5 \\
    \midrule
    Kimi K3 & 3 & 0.3 & -- & 15 \\
    DeepSeek V4 & 0.462 & 0.0154 & -- & 1.386 \\
    GPT-6 Astra & 10 & 1 & 12.5 & 50 \\
    GPT-6 Sol & 2 & 0.2 & 2.5 & 10 \\
    \bottomrule
  \end{tabular}
  \caption{Token prices (USD per million tokens) used to reprice every agent session: the list rates at which Claude Code prices the Claude models, and OpenRouter's list rates of 2026-09-24, during the runs, for the others (DeepSeek~V4's fell to 0.264/0.0088/0.792 by 2026-09-26). Claude Code writes a one-hour prompt cache at twice the input rate; the background model and the sessions rerun through OpenRouter write a five-minute cache at 1.25$\times$; Kimi~K3 and DeepSeek~V4 wrote no cache.}
  \label{tab:app_prices}
\end{table}

\subsection{Simulation}
\label{app:cost_sim}

\paragraph{By task.}
In each of the three studies, Syringe Packing and Cable Routing take most of the simulator time although they
contribute fewer than half of the episodes: their episodes run the most control steps (their step caps are the two
largest, Table~\ref{tab:app_tasks}), and Cable Routing is also the slowest task to simulate per step.

\paragraph{By split.}
Across splits, the evaluation schedule (Table~\ref{tab:app_schedule}) decides where the time goes. A setting of
Table~\ref{tab:models} evaluates ten held-out instances every round, so held-out episodes take the largest share of
its simulator time; the 6-iteration runs of the other two studies evaluate them only at $k{=}0$ and on the kept
iterate, so validation takes the largest share there.

\subsection{Total LLM Spend}
\label{app:cost_total}

Table~\ref{tab:app_spend} totals the LLM spend of the paper: the four study rows of Table~\ref{tab:app_cost}, the
pilot study and the curriculum study. The pilot study covers the other Claude Code and Codex sessions on our
development workstation, which also ran part of the study: those in its session logs, measured with
\texttt{ccusage}, and older harness sessions, at the cost their transcripts record; older interactive sessions left
no record, and the other study machines kept no logs of their sessions outside Table~\ref{tab:app_cost}. The
curriculum study's spend is the figure its authors report. Nothing is counted twice: the study-run sessions of
Table~\ref{tab:app_cost} are removed from the pilot study by session id, and the curriculum study ran on its
authors' machines.

\begin{table}[ht]
  \centering
  \footnotesize
  \setlength{\tabcolsep}{3.5pt}
  \setlength{\aboverulesep}{0pt}
  \setlength{\belowrulesep}{0pt}
  \renewcommand{\arraystretch}{1.15}
  \begin{tabular}{l|l|r}
    \toprule
    Component & Source & Cost (USD) \\
    \midrule
    Table~\ref{tab:models} & Table~\ref{tab:app_cost} & 1{,}700.03 \\
    Feedback study & Table~\ref{tab:app_cost} & 3{,}186.60 \\
    Skill library & Table~\ref{tab:app_cost} & 747.42 \\
    Superseded passes and dropped arms & Table~\ref{tab:app_cost} & 1{,}923.12 \\
    \midrule
    Pilot study & \texttt{ccusage}, harness transcripts & 22{,}462.13 \\  
    Curriculum study & reported by its authors & 6{,}436.83 \\
    \midrule
    Total & & 36{,}456.13 \\
    \bottomrule
  \end{tabular}
  \caption{Total LLM spend of the paper's coding agents. The first four rows are the study rows of Table~\ref{tab:app_cost}. The pilot study covers the other Claude Code and Codex sessions on our development workstation from 2026-08-18 to 2026-09-26 (Appendix~\ref{app:cost_total}), priced by \texttt{ccusage} from the session logs (models it does not price at the rates of Table~\ref{tab:app_prices}) or, for harness sessions whose logs had been deleted, at the cost their transcripts record. The curriculum study's spend is reported by its authors.}
  \label{tab:app_spend}
\end{table}

\FloatBarrier
\appendixpart{Additional Results}

\FloatBarrier

\FloatBarrier
\section{Failure Analysis}
\label{app:failures}

This appendix asks where the programs behind the paper's tables fail. We define ten failure categories
(Appendix~\ref{app:fail_taxonomy}) and compare them across tasks (Appendix~\ref{app:fail_tasks}) and across models
and feedback rounds (Appendix~\ref{app:fail_models}). A category marks where a failure becomes observable, possibly
several skills after the decision that caused it (Finding~3). The harness's own interventions (guard rejections,
preflight crashes, reverts) are reported with the harness and its rules (Appendices~\ref{app:harness}
and~\ref{app:stopping}).

\subsection{Taxonomy and Assignment}
\label{app:fail_taxonomy}

\paragraph{Unit.}
The unit is the failed held-out object (two per Syringe Packing episode, three per Cable Routing episode, otherwise
one; Table~\ref{tab:app_tasks}) of the zero-shot and @2 programs of every Table~\ref{tab:models} run and of the kept
programs of the feedback and restricted skill-library runs, scored with the final per-object rule of
Appendix~\ref{app:tasks_scoring}. Pooled columns count each held-out batch once, so an initial program kept by several
seeded arms counts once.

\paragraph{Precedence.}
An episode's failed objects take its category from the first rule that applies: (1)~the outcome, step budget (U) or
a diverged simulation or failed machine (I); (2)~the message of the first node that raised, matched by keyword rules
to P, G, K, T, B, or A for a placement check that raises; (3)~a self-check abort (S) when the program routes itself to
a failure end; (4)~for a finished program the evaluator fails, A when the placement clause is unmet and R when only a
rest, release or turning clause is. An error outranks the goal clause because a node that raises stops the program
and leaves the placement unmet as a consequence. Of the final programs' failed objects, 47\% are assigned by the goal
clause, 36\% by an error message and 18\% by the outcome or end node. Table~\ref{tab:app_failures} gives each rule with a
verbatim example; the keyword rules were checked by eye on each category's commonest messages. Held-out episodes store
no trace, so a self-check abort cannot be traced to the check that fired.

\begin{table}[ht]
  \centering
  \footnotesize
  \setlength{\tabcolsep}{3.5pt}
  \renewcommand{\arraystretch}{1.15}
  \begin{tabularx}{\linewidth}{@{}l>{\raggedright\arraybackslash}p{0.17\linewidth}>{\raggedright\arraybackslash}X>{\raggedright\arraybackslash}p{0.4\linewidth}@{}}
    \toprule
    & Category & Rule & Example message \\
    \midrule
    P & Perception & A perception node raises: the target, part or region is not found or not localized. & \texttt{SAM3 could not localize \textquotesingle{}ring\textquotesingle{}} \\
    G & Grasp miss/\allowbreak{}loss & The jaws close on nothing, or grasp verification finds the object gone from the hand. & \texttt{the jaws closed on nothing (gap 8.00 mm, lag 0.00 mm)} \\
    K & Planning/\allowbreak{}IK refusal & The planner or IK returns no feasible grasp, orientation, path or approach. & \texttt{plan\_\allowbreak{}grasp: 0/\allowbreak{}1 candidate grasps had a feasible CuRobo plan \dots} \\
    T & Reach/\allowbreak{}tracking/\allowbreak{}transport & The hand stops short of, or tracks off, a commanded pose. & \texttt{the hand stopped 23 mm from the grasp pose} \\
    A & Placement/\allowbreak{}insertion misalignment & The graph finishes with the object's placement clause unmet, or a placement or insertion check raises. & \texttt{encloses/\allowbreak{}wrench\_\allowbreak{}1.ring/\allowbreak{}wrench\_\allowbreak{}hook\_\allowbreak{}1.shank: False} \\
    R & Release/\allowbreak{}goal-clause & The graph finishes with placement met but a rest, release or turning clause unmet. & \texttt{speed/\allowbreak{}wrench\_\allowbreak{}1: False} \\
    S & Self-check abort & The program routes itself to a failure end node; no node raises. & \texttt{workflow terminated at end node \textquotesingle{}unfinished\textquotesingle{} (failure)} \\
    B & Program bug/\allowbreak{}tool misuse & The program's own code raises, or it calls a tool outside the tool's contract. & \texttt{TypeError: run() missing 1 required positional argument: \textquotesingle{}fixture\_\allowbreak{}feature\textquotesingle{}} \\
    U & Step budget & The episode reaches its step limit. & \texttt{close\_\allowbreak{}kind: truncated} \\
    I & Infrastructure/\allowbreak{}simulator & The physics diverges or the machine fails. & \texttt{OSError: [Errno 28] No space left on device} \\
    \bottomrule
  \end{tabularx}
  \caption{Failure categories of held-out objects. A failed object takes its episode's category, assigned by outcome first, then by the error text of the node that raised, then by the goal clause left unmet; examples are verbatim from held-out episode records: the error message, or for A and R an unmet goal predicate and for U the close kind.}
  \label{tab:app_failures}
\end{table}

\subsection{Failures by Task}
\label{app:fail_tasks}

Table~\ref{tab:app_failure_shares} gives the shares per task (left) and per study and reading (right). Placement is
49\% of all failures and the largest category everywhere except Tool Hanging; 93\% of it comes from programs that
finish without an error while the object is misplaced.

\begin{table}[ht]
  \centering
  \setlength{\tabcolsep}{3.5pt}
  \setlength{\aboverulesep}{0pt}
  \setlength{\belowrulesep}{0pt}
  \renewcommand{\arraystretch}{1.15}
  \resizebox{\linewidth}{!}{%
  \begin{tabular}{l|ccccc|c|cc|c|c}
    \toprule
    & \multicolumn{5}{c|}{Final programs, by task} & & \multicolumn{2}{c|}{Table~\ref{tab:models}} & & \\
    \cmidrule(lr){2-6} \cmidrule(lr){8-9}
    Category & \begin{tabular}[b]{@{}c@{}}Tool\\Hanging\end{tabular} & \begin{tabular}[b]{@{}c@{}}Syringe\\Packing\end{tabular} & \begin{tabular}[b]{@{}c@{}}Gear\\Meshing\end{tabular} & \begin{tabular}[b]{@{}c@{}}USB-C\\Insertion\end{tabular} & \begin{tabular}[b]{@{}c@{}}Cable\\Routing\end{tabular} & All & @0 & @2 & Feedback & \begin{tabular}[b]{@{}c@{}}Skill\\library\end{tabular} \\
    \midrule
    P Perception & 1 & 15 & 17 & 1 & 7 & 8 & 11 & 7 & 6 & 25 \\
    G Grasp & $<$1 & 14 & 0 & 17 & 0 & 6 & 13 & 11 & 0 & 1 \\
    K Planning/IK & \textbf{45} & 1 & 0 & 18 & 0 & 13 & 16 & 16 & 10 & 0 \\
    T Reach/tracking & 13 & 6 & 0 & 2 & 0 & 5 & 9 & 5 & 5 & 4 \\
    A Placement & 36 & \textbf{43} & \textbf{67} & \textbf{36} & \textbf{69} & \textbf{49} & \textbf{31} & \textbf{39} & \textbf{63} & \textbf{69} \\
    R Release/clause & 1 & 0 & 4 & 0 & 0 & $<$1 & 0 & 0 & 1 & 0 \\
    S Self-check abort & 0 & 1 & 9 & 9 & 20 & 7 & 9 & 8 & 9 & 0 \\
    B Program bug & 0 & 0 & 0 & 5 & 0 & 1 & 3 & 0 & 2 & 0 \\
    U Step budget & 4 & 20 & 4 & 9 & 4 & 10 & 8 & 14 & 4 & 1 \\
    I Infra./sim. & $<$1 & 0 & 0 & 4 & 1 & 1 & $<$1 & 1 & 0 & 0 \\
    \bottomrule
  \end{tabular}}
  \caption{Share (\%) of the failed held-out objects in each category: per task and in All over the final programs of the three studies (Table~\ref{tab:models} @2, and the kept programs of every feedback arm, including the \emph{Outcome + object state} and \emph{Outcome + relations} rows of Table~\ref{tab:app_feedback_conditions}, and of the two restricted skill-library arms), each held-out batch once; then per study, split as in Table~\ref{tab:app_cost}, with Table~\ref{tab:models} at its zero-shot (@0) and @2 programs. Bold marks each column's largest category.}
  \label{tab:app_failure_shares}
\end{table}

\paragraph{Per task.} The categories beyond placement set the tasks apart.
\begin{itemize}[nosep, leftmargin=*]
  \item \emph{Tool Hanging} fails mostly before the hook: 45\% are planning refusals, three in five of them the planner
  rejecting every candidate grasp (\texttt{plan\_grasp: 0/1 candidate grasps had a feasible CuRobo plan \dots}).
  \item \emph{Syringe Packing} adds step budget (20\%), perception (15\%, mostly \texttt{could not perceive
  \textquotesingle{}medical syringe\textquotesingle{}}) and grasp failures (14\%, mostly the program's own in-hand check
  seeing no syringe follow the lifted hand) to placement (43\%).
  \item \emph{Gear Meshing}, which most programs solve, fails mostly with no gear seated (67\%); most of its
  perception failures (17\%) are programs that find no red start button (\texttt{nothing red enough to be the button is visible}).
  \item \emph{USB-C Insertion} spreads over placement (36\%, nearly all a plug short of the insertion depth), planning
  refusals (18\%, e.g.\ \texttt{no reachable transit above the plug}) and grasp misses (17\%); its program bugs (5\%)
  all come from the seeded arms' inherited initial policy, which calls \texttt{usbc.receptacle\_shape} on a port the
  tool does not describe.
  \item \emph{Cable Routing} adds self-check aborts (20\%, more than half to an end node named \texttt{unfinished}) to
  placement (69\%).
\end{itemize}

\subsection{Failures by Model and Feedback Round}
\label{app:fail_models}

\begin{figure}[ht]
  \centering
  \includegraphics[width=\linewidth]{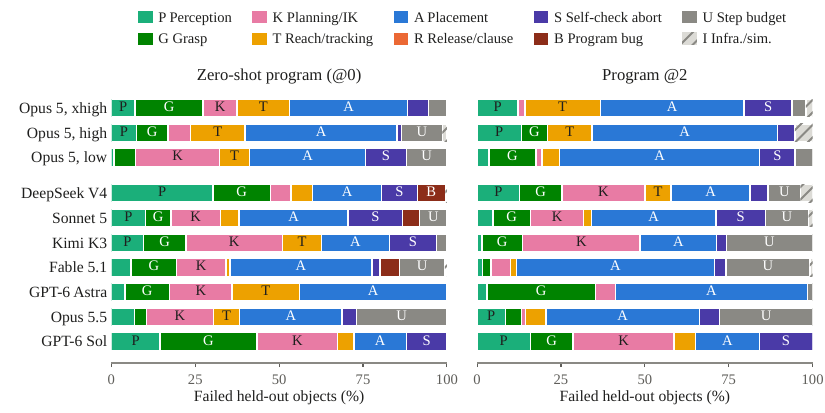}
  \caption{Category shares (\%) of the failed held-out objects per Table~\ref{tab:models} setting, for the zero-shot
  program (left) and the program @2 reads (right). Letters follow Table~\ref{tab:app_failures}; rows follow
  Table~\ref{tab:models}.}
  \label{fig:app_failures}
\end{figure}

Figure~\ref{fig:app_failures} splits the @0 and @2 columns of Table~\ref{tab:app_failure_shares} by model.

\paragraph{Pooled.}
From @0 to @2, perception (11$\rightarrow$7\%), grasp (13$\rightarrow$11\%), reach (9$\rightarrow$5\%) and program-bug
failures (3$\rightarrow$0\%) shrink while placement (31$\rightarrow$39\%) and step budget (8$\rightarrow$14\%) grow:
improved programs fail later in the task. The program bugs that still reach held-out episodes escape the single
preflight episode, and none survives to @2. Both growing categories mostly carry no error message, so their cause
must be traced back from intermediate outcomes, the attribution problem of Finding~3.

\paragraph{By model.}
Planning refusals, flat at 16\% pooled, separate the models. The three Opus~5 rows, Fable~5.1 and Opus~5.5 cut them to
at most 6\% at @2, leaving placement as their largest category (43--60\%); on Tool Hanging, Opus~5, low, Fable~5.1 and
Opus~5.5 move from failing mainly at the planner to failing mainly at the hook. GPT-6 Astra cuts them as well (to
6\%) but trades them for grasp failures (13$\rightarrow$32\%). DeepSeek~V4, Sonnet~5, Kimi~K3 and GPT-6 Sol have
16--35\% at @2; on Tool Hanging, which they leave at 0--2\%, planning refusals are each model's largest category. For
DeepSeek~V4, feedback repaired the perception behind 70\% of its @0 Tool Hanging failures only to expose planning
refusals behind all of its @2 ones. Models that start alike thus stall at different stages (Finding~1), and four of
the five models that stall on Tool Hanging stall where a collision-aware planner refuses their grasp or path
(Finding~5); GPT-6 Astra stalls at the hook (85\% of its @2 Tool Hanging failures are placement).

\paragraph{Standard Deviation}
\begin{table}[ht]
  \centering
  \footnotesize
  \setlength{\tabcolsep}{3.5pt}
  \setlength{\aboverulesep}{0pt}
  \setlength{\belowrulesep}{0pt}
  \renewcommand{\arraystretch}{1.15}
  \resizebox{\ifdim\width>\linewidth\linewidth\else\width\fi}{!}{%
  \begin{tabular}{l|ccc|ccc|ccc}
    \toprule
    Model & \multicolumn{3}{c|}{Overall} & \multicolumn{3}{c|}{Tool Hanging} & \multicolumn{3}{c}{Syringe Packing} \\
    \cmidrule(lr){2-4} \cmidrule(lr){5-7} \cmidrule(lr){8-10}
    & @0 & @1 & @2 & @0 & @1 & @2 & @0 & @1 & @2 \\
    \midrule
    Human & \multicolumn{3}{c|}{82} & \multicolumn{3}{c|}{87} & \multicolumn{3}{c}{90} \\
    \midrule
    Opus 5, xhigh & 45{\scriptsize$\pm$7} & 64{\scriptsize$\pm$9} & 74{\scriptsize$\pm$4} & 22{\scriptsize$\pm$33} & 62{\scriptsize$\pm$26} & 60{\scriptsize$\pm$19} & 28{\scriptsize$\pm$4} & 38{\scriptsize$\pm$2} & 35{\scriptsize$\pm$0} \\
    Opus 5, high & 41{\scriptsize$\pm$12} & 62{\scriptsize$\pm$12} & \textbf{81}{\scriptsize$\pm$4} & 25{\scriptsize$\pm$43} & \textbf{72}{\scriptsize$\pm$42} & \textbf{97}{\scriptsize$\pm$4} & 40{\scriptsize$\pm$9} & 30{\scriptsize$\pm$0} & 45{\scriptsize$\pm$0} \\
    Opus 5, low & 57{\scriptsize$\pm$6} & 67{\scriptsize$\pm$8} & 74{\scriptsize$\pm$7} & 0{\scriptsize$\pm$0} & 30{\scriptsize$\pm$37} & 57{\scriptsize$\pm$35} & \textbf{51}{\scriptsize$\pm$6} & 50{\scriptsize$\pm$4} & \textbf{60}{\scriptsize$\pm$0} \\
    \midrule
    DeepSeek V4 & 38{\scriptsize$\pm$4} & 43{\scriptsize$\pm$4} & 44{\scriptsize$\pm$4} & 0{\scriptsize$\pm$0} & 0{\scriptsize$\pm$0} & 0{\scriptsize$\pm$0} & 20{\scriptsize$\pm$14} & 29{\scriptsize$\pm$12} & 30{\scriptsize$\pm$12} \\
    Sonnet 5 & 45{\scriptsize$\pm$8} & 46{\scriptsize$\pm$8} & 46{\scriptsize$\pm$8} & 0{\scriptsize$\pm$0} & 0{\scriptsize$\pm$0} & 0{\scriptsize$\pm$0} & 32{\scriptsize$\pm$6} & 34{\scriptsize$\pm$7} & 34{\scriptsize$\pm$7} \\
    Kimi K3 & 39{\scriptsize$\pm$3} & 43{\scriptsize$\pm$7} & 46{\scriptsize$\pm$6} & 0{\scriptsize$\pm$0} & 0{\scriptsize$\pm$0} & 2{\scriptsize$\pm$4} & 30{\scriptsize$\pm$14} & 47{\scriptsize$\pm$33} & 55{\scriptsize$\pm$29} \\
    Fable 5.1 & 44{\scriptsize$\pm$12} & 57{\scriptsize$\pm$12} & 64{\scriptsize$\pm$13} & 24{\scriptsize$\pm$43} & 32{\scriptsize$\pm$39} & 32{\scriptsize$\pm$39} & 49{\scriptsize$\pm$17} & 49{\scriptsize$\pm$17} & 56{\scriptsize$\pm$17} \\
    GPT-6 Astra & 43{\scriptsize$\pm$2} & 47{\scriptsize$\pm$4} & 48{\scriptsize$\pm$5} & 0{\scriptsize$\pm$0} & 0{\scriptsize$\pm$0} & 0{\scriptsize$\pm$0} & 20{\scriptsize$\pm$4} & 25{\scriptsize$\pm$7} & 30{\scriptsize$\pm$11} \\
    Opus 5.5 & \textbf{67}{\scriptsize$\pm$8} & \textbf{74}{\scriptsize$\pm$8} & 77{\scriptsize$\pm$7} & 25{\scriptsize$\pm$29} & 40{\scriptsize$\pm$41} & 55{\scriptsize$\pm$34} & 46{\scriptsize$\pm$14} & 50{\scriptsize$\pm$11} & 50{\scriptsize$\pm$11} \\
    GPT-6 Sol & 33{\scriptsize$\pm$7} & 41{\scriptsize$\pm$9} & 53{\scriptsize$\pm$7} & 0{\scriptsize$\pm$0} & 0{\scriptsize$\pm$0} & 0{\scriptsize$\pm$0} & 30{\scriptsize$\pm$11} & 30{\scriptsize$\pm$11} & 30{\scriptsize$\pm$11} \\
    \bottomrule
  \end{tabular}}
  \par\medskip
  \resizebox{\ifdim\width>\linewidth\linewidth\else\width\fi}{!}{%
  \begin{tabular}{l|ccc|ccc|ccc}
    \toprule
    Model & \multicolumn{3}{c|}{Gear Meshing} & \multicolumn{3}{c|}{USB-C Insertion} & \multicolumn{3}{c}{Cable Routing} \\
    \cmidrule(lr){2-4} \cmidrule(lr){5-7} \cmidrule(lr){8-10}
    & @0 & @1 & @2 & @0 & @1 & @2 & @0 & @1 & @2 \\
    \midrule
    Human & \multicolumn{3}{c|}{100} & \multicolumn{3}{c|}{87} & \multicolumn{3}{c}{47} \\
    \midrule
    Opus 5, xhigh & 100{\scriptsize$\pm$0} & 100{\scriptsize$\pm$0} & 100{\scriptsize$\pm$0} & 0{\scriptsize$\pm$0} & 50{\scriptsize$\pm$35} & 100{\scriptsize$\pm$0} & 73{\scriptsize$\pm$2} & 72{\scriptsize$\pm$1} & 73{\scriptsize$\pm$2} \\
    Opus 5, high & 67{\scriptsize$\pm$41} & 100{\scriptsize$\pm$0} & 100{\scriptsize$\pm$0} & 0{\scriptsize$\pm$0} & 33{\scriptsize$\pm$41} & 87{\scriptsize$\pm$16} & 75{\scriptsize$\pm$5} & 76{\scriptsize$\pm$5} & 76{\scriptsize$\pm$5} \\
    Opus 5, low & 100{\scriptsize$\pm$0} & 100{\scriptsize$\pm$0} & 100{\scriptsize$\pm$0} & 60{\scriptsize$\pm$28} & 80{\scriptsize$\pm$12} & 80{\scriptsize$\pm$12} & 73{\scriptsize$\pm$13} & 76{\scriptsize$\pm$6} & 76{\scriptsize$\pm$6} \\
    \midrule
    DeepSeek V4 & 100{\scriptsize$\pm$0} & 100{\scriptsize$\pm$0} & 100{\scriptsize$\pm$0} & 0{\scriptsize$\pm$0} & 10{\scriptsize$\pm$12} & 12{\scriptsize$\pm$16} & 70{\scriptsize$\pm$11} & 77{\scriptsize$\pm$4} & 77{\scriptsize$\pm$4} \\
    Sonnet 5 & 100{\scriptsize$\pm$0} & 100{\scriptsize$\pm$0} & 100{\scriptsize$\pm$0} & 25{\scriptsize$\pm$38} & 25{\scriptsize$\pm$38} & 25{\scriptsize$\pm$38} & 67{\scriptsize$\pm$8} & 72{\scriptsize$\pm$4} & 72{\scriptsize$\pm$4} \\
    Kimi K3 & 97{\scriptsize$\pm$4} & 97{\scriptsize$\pm$4} & 97{\scriptsize$\pm$4} & 0{\scriptsize$\pm$0} & 0{\scriptsize$\pm$0} & 0{\scriptsize$\pm$0} & 69{\scriptsize$\pm$4} & 71{\scriptsize$\pm$3} & 73{\scriptsize$\pm$0} \\
    Fable 5.1 & 67{\scriptsize$\pm$41} & 100{\scriptsize$\pm$0} & 100{\scriptsize$\pm$0} & 0{\scriptsize$\pm$0} & 25{\scriptsize$\pm$43} & 50{\scriptsize$\pm$50} & 79{\scriptsize$\pm$4} & 79{\scriptsize$\pm$4} & 79{\scriptsize$\pm$4} \\
    GPT-6 Astra & 100{\scriptsize$\pm$0} & 100{\scriptsize$\pm$0} & 100{\scriptsize$\pm$0} & 15{\scriptsize$\pm$11} & 30{\scriptsize$\pm$21} & 30{\scriptsize$\pm$21} & 80{\scriptsize$\pm$0} & 80{\scriptsize$\pm$0} & 80{\scriptsize$\pm$0} \\
    Opus 5.5 & 100{\scriptsize$\pm$0} & 100{\scriptsize$\pm$0} & 100{\scriptsize$\pm$0} & \textbf{85}{\scriptsize$\pm$26} & \textbf{100}{\scriptsize$\pm$0} & 100{\scriptsize$\pm$0} & 79{\scriptsize$\pm$5} & 79{\scriptsize$\pm$5} & 79{\scriptsize$\pm$5} \\
    GPT-6 Sol & 50{\scriptsize$\pm$35} & 50{\scriptsize$\pm$35} & 50{\scriptsize$\pm$35} & 0{\scriptsize$\pm$0} & 40{\scriptsize$\pm$28} & 100{\scriptsize$\pm$0} & \textbf{83}{\scriptsize$\pm$0} & \textbf{83}{\scriptsize$\pm$0} & \textbf{83}{\scriptsize$\pm$0} \\
    \bottomrule
  \end{tabular}}
  \caption{Table~\ref{tab:models} with run-to-run variation: mean $\pm$ standard deviation (percentage points) over the five runs of each entry; Human is VABAR's Level-1 result, without runs.}
  \label{tab:models_sd}
\end{table}

\paragraph{Other studies.}
The feedback study's kept programs fail at placement more often (63\%) than Table~\ref{tab:models}'s @2 programs
(39\%) and never at the grasp. The restricted skill-library arms' perception share (25\%) comes entirely from the
tools-only arm on Cable Routing, which segments no cable in any held-out episode.

\FloatBarrier

\FloatBarrier
\appendixpart{Curriculum Study}
\section{Privileged-to-Perception Curriculum}
\label{sec:app-curriculum}

This appendix provides the full experimental details for the curriculum study in Section~\ref{sec:curriculum-design} (Figure~\ref{fig:mimicgen_pv} and Table~\ref{tab:curriculum-comparison}). Section~\ref{sec:curriculum-setup} describes the two learning settings, tool access, tasks, training and evaluation protocol, and the conversion from privileged geometry to perception. Section~\ref{sec:curriculum-transfer} then analyzes representative cases in which transfer is unsuccessful.

\subsection{Experimental Setup}
\label{sec:curriculum-setup}

\subsubsection{Settings and Agent}
\paragraph{Learning settings.}
We compare two settings that use the same self-learning loop (Section~\ref{sec:protocol}), model, instructions, skill library, feedback, training layouts, and total iteration budget. They differ only in how the policy observes the scene.
\begin{itemize}[nosep, leftmargin=*]
  \item \emph{Privileged $\rightarrow$ perception} (w/ curr.). For the first $K_p$ iterations, the policy observes exact simulator geometry and has no camera access. We then convert the best privileged checkpoint to a perception policy and continue learning for $K_v$ iterations using perception only.
  \item \emph{Perception only} (w/o curr.). The policy uses cameras and perception tools for all $K_p + K_v$ iterations.
\end{itemize}

MimicGen and VABAR runs use different configurations, described in Section~\ref{sec:curriculum-suite-diff}. Within each suite, the privileged $\rightarrow$ perception and perception-only settings use the same model, probe allowance, feedback format, training and held-out layouts, iteration budget, and evaluation protocol; they differ only in the scene information available to the policy.

On MimicGen, we use Claude Fable~5.1 with $K_p = K_v = 8$ (16 iterations per run). On VABAR, we use Claude Opus~5 with $K_p = K_v = 5$ (10 iterations per run). We run each setting three times per task (runs 000, 001, and 002), giving $7 \times 3 = 21$ curriculum runs and 21 perception-only runs.

\paragraph{Initial policies.}
Iteration 0 is a zero-shot policy written by the agent from the task instruction and skill library. In the perception-only setting, each run starts from a fixed zero-shot perception policy shared with the other perception-only studies in the paper. In the privileged setting, the agent writes a separate zero-shot policy using the privileged tool surface, because a perception policy cannot execute with the privileged tool list. On VABAR, run 000 in each setting uses the first workflow graph produced for that run, while runs 001 and 002 are generated in new sessions from the same frozen prompt and model without additional training.

\paragraph{Agent harness.}
The agent is Claude Code with temperature 0.2. It can read, write, and search files in the policy workspace, run \texttt{./validate.sh}, and copy library skills into the workspace. It cannot invoke the simulator directly, except on MimicGen, where it may call \texttt{./probe.sh} for probe episodes (Appendix~\ref{sec:curriculum-training}). Each improvement step starts a new session that receives the current policy, the latest feedback report, and a summary of previous iterations and code changes. On MimicGen, a session is limited to 35 turns, 10 minutes of active time, and 45 minutes total. On VABAR, a session is limited to 80 turns with no wall-clock limit, as in the other VABAR experiments. The conversion session uses the same turn limit and no wall-clock limit.

\subsubsection{Tool Access}
\paragraph{Privileged tools.}
During the privileged phase, the policy observes the scene only through simulator geometry queries: \texttt{sim.list\_objects}, \texttt{sim.get\_object\_pose}, and \texttt{sim.get\_object\_obb} provide exact object poses and oriented bounding boxes, while \texttt{sim.list\_features} and \texttt{sim.get\_feature} provide world-frame geometry for named functional parts. These parts are defined by the task assets; examples include the needle \emph{handle} and \emph{shaft} in Threading, the stove \emph{burner} in Kitchen, the machine \emph{holder} and \emph{lid} in Coffee, the drawer \emph{handle} and \emph{floor} in Mug Cleanup, and each wrench \emph{ring} and hook \emph{shank} in Tool Hanging. VABAR additionally provides \texttt{sim.list\_regions} and \texttt{sim.get\_region\_obb} for regions, and \texttt{sim.get\_fixture\_layout} for multi-part fixtures without a single object pose, such as the peg centers, pinion center, and button on the Gear Meshing board.

All camera and perception tools are disabled during this phase, including RGB-D observations, Grounding DINO detection, SAM3 segmentation and tracking, and geometry helpers that recover 3D structure from masks and depth. Robot control, motion, and grasp skills are unchanged. The agent is explicitly instructed to learn task decomposition and manipulation logic that can later be executed from RGB-D observations, and not to hard-code poses observed in feedback.

\paragraph{Perception tools.}
The perception policy observes the scene only through calibrated cameras that provide RGB, depth, intrinsics, and camera pose. MimicGen uses the fixed \texttt{agentview} camera and a wrist camera; VABAR uses each task's own camera set: fixed overhead, front, side, or top-down views, plus wrist cameras in Tool Hanging and USB-C Insertion. The policy can use Grounding DINO for text-prompted bounding boxes, SAM3 for masks from text, boxes, or points, and geometry utilities that back-project masks and depth into world coordinates and fit boxes, lines, and planes. All \texttt{sim.*} geometry queries are disabled.

In both settings, the policy never receives task flags, goal predicates, success checks, or task-specific planners. These signals are used only for scoring and feedback.

\paragraph{Tool enforcement.}
Each setting uses a fixed tool list. Before an edited policy can run, a validator checks every policy script and rejects any call outside the allowed tool surface, including calls constructed dynamically at run time. The agent must pass this check with \texttt{./validate.sh} before the session ends. If the check fails, the violation is returned to the agent for up to two repair rounds.

\subsubsection{Tasks}
Table~\ref{tab:curriculum-tasks} lists the seven tasks used in the study. We select tasks that require both accurate state estimation and non-trivial manipulation, so that separating perception from manipulation can meaningfully affect learning.
\begin{itemize}[nosep, leftmargin=*]
  \item \emph{Threading} and \emph{Coffee} require precise insertion. A failure can therefore arise from target localization, the object's pose in the gripper, or the insertion motion itself.
  \item \emph{Kitchen} and \emph{Mug Cleanup} are long, multi-stage tasks with articulated objects, so errors early in the trajectory can affect later stages.
  \item \emph{Tool Hanging} (Level~2) contains two wrenches, two hooks, and distractor tools. Each wrench must be grasped so that its ring remains available for hanging and the object remains stable during transport.
  \item \emph{Gear Meshing} (Level~2) requires placing three gears on thin pegs so that they mesh with a fixed pinion, followed by a button press.
  \item \emph{USB-C Insertion} (Level~1) requires aligning a small plug with a receptacle to within a few millimeters and inserting it to sufficient depth. 
\end{itemize}
In the privileged phase, task-relevant parts such as rings, pegs, and ports are available through exact geometry, which largely removes localization error. After conversion, the same structures must be recovered from RGB-D observations, including small or low-contrast parts.

\begin{table}[h]
\centering
\small
\setlength{\tabcolsep}{2pt}
\begin{tabular}{@{}p{0.11\linewidth}p{0.18\linewidth}p{0.43\linewidth}p{0.08\linewidth}p{0.13\linewidth}@{}}
\toprule
Suite & Task & Instruction & Stages & Episode \\
\midrule
MimicGen & Coffee\newline (D1) & Insert the coffee pod into the coffee machine and close the lid. & 7 & 1500\newline steps \\
 & Kitchen\newline (D1) & Cook the bread in the pot on the stove, serve the pot with the bread retained, and finish with the stove off. & 7 & 1500\newline steps \\
 & Mug Cleanup\newline (D1) & Open the target drawer, pick up the mug, place it upright inside the drawer, release it, close the drawer. & 7 & 1500\newline steps \\
 & Threading\newline (D1) & Pick up the needle by its handle, align its shaft with the tripod ring, and thread it through the ring. & 5 & 1500\newline steps \\
\midrule
VABAR & Tool Hanging\newline (Level 2) & Hang both wrenches on the two hooks. & 7 & task\newline budget \\
 & Gear Meshing\newline (Level 2) & Put each gear on its own peg so they mesh, then press the red button at the near end of the board. & 5 & task\newline budget \\
 & USB-C Insertion\newline (Level 1) & Insert the USB-C plug into the receptacle. & 4 & task\newline budget \\
\bottomrule
\end{tabular}
\caption{Tasks used in the curriculum study. ``Stages'' denotes the number of ordered reference stages used for stage feedback and stage reward, including the final success stage.}
\label{tab:curriculum-tasks}
\end{table}

\subsubsection{Differences Between MimicGen and VABAR Runs}
\label{sec:curriculum-suite-diff}
For each benchmark, the two curriculum settings use the same experimental configuration. The VABAR experiments use the same setup as the other VABAR experiments in this paper: Claude Opus~5, text-only feedback, a fixed set of four training layouts, no probe episodes, and sessions of up to 80 turns with no time limit.

For MimicGen, we adapt this setup to make self-learning more tractable because its tasks present a harder learning problem. MimicGen tasks are long-horizon, multi-stage, and involve multiple objects. For example, Kitchen requires the agent to interact with the stove, pot, bread, and serving area in the correct sequence over a 1500-step episode. Moreover, the skill library contains no task-specific skills for most MimicGen tasks, so the agent must develop the task-specific skills needed to solve them during self-learning. Object configurations are also randomized over a wide range, making it easy for a policy trained on a fixed set of layouts to overfit.

We therefore use a stronger model for MimicGen (Claude Fable~5.1), provide failure-image feedback, allow more learning iterations and training episodes, resample training seeds at every iteration to reduce overfitting, and include probe episodes so that the agent can test a hypothesis before modifying its policy. Because MimicGen learning runs are longer (16 iterations), we additionally limit each session to 35 turns and 10 minutes of agent update time. This encourages each iteration to make a more localized policy update. The differences between the VABAR and MimicGen configurations are summarized in Table~\ref{tab:curriculum-suite-diff}.

\begin{table}[h]
\centering
\small
\setlength{\tabcolsep}{4pt}
\begin{tabular}{p{0.28\linewidth}p{0.33\linewidth}p{0.33\linewidth}}
\toprule
 & MimicGen & VABAR \\
\midrule
Tasks & Coffee, Kitchen, Mug Cleanup, Threading (D1) & Tool Hanging, Gear Meshing (Level 2); USB-C Insertion (Level 1) \\
Model & Claude Fable 5.1 & Claude Opus 5 \\
Iterations (privileged + perception) & 8 + 8 & 5 + 5 \\
Iterations (perception only) & 16 & 10 \\
Training episodes per iteration & 10 & 4 \\
Training layouts & 10 new every iteration (seeds 0--159 over 16 iterations) & same 4 every iteration \\
Probe episodes & up to 2 per iteration (seeds 160--449) & none \\
Held-out & 10 layouts (seeds 450--459) & 10 layouts (layouts 012--021) \\
Episode length & 1500 steps & task time limit \\
Feedback & text+2 end-of-episode \texttt{agentview} frames & text only \\
Success measure & native task success & per-object success \\
Turns per session & 35 & 80 \\
Time per session & 10 min active, 45 min total & no limit \\
Cameras & \texttt{agentview}, wrist & fixed views per task; wrist cameras in Tool Hanging and USB-C \\
\bottomrule
\end{tabular}
\caption{Configuration differences between MimicGen and VABAR curriculum runs. Both settings (privileged $\rightarrow$ perception and perception only) use the same configuration within a suite. The conversion session uses the suite's turn limit with no time limit.}
\label{tab:curriculum-suite-diff}
\end{table}

\subsubsection{Training and Evaluation}
\label{sec:curriculum-training}
\paragraph{Training, held-out, and probe layouts.}
On MimicGen, each iteration trains on 10 newly sampled layouts, with training layouts not reused across iterations. The agent may additionally run up to two unscored probe episodes per iteration to test a hypothesis. Because VABAR contains fewer layouts, each iteration instead uses the same four fixed training layouts, one episode per layout, and probe episodes are disabled. In both cases, the agent is told that evaluation uses unseen layouts and that movable objects must therefore be localized at run time.

After every iteration, the policy is evaluated on 10 fixed held-out layouts that are disjoint from training. Held-out results are never shown to the agent; they are used only to construct the learning curves and to select the privileged checkpoint for conversion. Within each benchmark, the curriculum and perception-only settings use the same training sequence and the same held-out layouts.

\paragraph{One learning iteration.}
Each iteration has three stages. (1)~\emph{Preflight:} the current policy runs on two unscored episodes of at most 300 steps to detect program errors. If it crashes, the agent receives the error and may perform up to two repair sessions. (2)~\emph{Training:} the policy runs on the training layouts and a feedback report is generated. (3)~\emph{Improvement:} a new agent session reads the report and edits the policy; the resulting policy must pass the tool validator. Held-out evaluation of the new policy runs in parallel with the next learning iteration. We do not revert edits or stop runs early, so every run uses its full iteration budget.

\paragraph{Feedback.}
Both settings receive the same feedback structure. The report describes object states and inter-object relations using named object parts and object-centered frames, organized by the named steps of the policy (Section~\ref{sec:protocol}). For each episode, it also reports which ordered reference stages were reached (Table~\ref{tab:curriculum-tasks}) and the corresponding stage reward: the number of stages completed in order before the first missed stage, divided by the total number of stages and averaged across episodes. Up to two episodes are shown in full detail, and the previous six iterations are summarized as history. MimicGen reports additionally include two end-of-episode \texttt{agentview} frames to provide richer feedback and further facilitate learning, whereas VABAR reports are text only. The privileged and perception-only settings therefore differ in policy observations, not in the feedback used for self-improvement.

\paragraph{Metrics.}
We report held-out success after every iteration: native task success on MimicGen and per-object success on VABAR (for example, each wrench or gear is scored separately). We also report held-out stage reward. In Figure~\ref{fig:mimicgen_pv}, each line is the mean across the three runs and the shaded band shows $\pm1$ standard error. In Table~\ref{tab:curriculum-comparison}, \#~Skills is the number of nodes in the final policy workflow graph, where each node is one code-executing step, and Success is the mean held-out success over the last three iterations; both are averaged across the three runs.

\subsubsection{Conversion to Perception}
At the end of the privileged phase, we select the privileged checkpoint with the highest held-out success, breaking ties by stage reward and then by the latest iteration. A dedicated conversion session receives the selected policy, its most recent feedback report, and the perception tool list, together with the following instruction:
\begin{quote}\small
\emph{The privileged perception phase is over. This checkpoint must now run on the vision-only tool surface. Preserve its task decomposition, manipulation logic, recovery branches, and bounded diagnostics. Replace every privileged sim geometry read and every privileged perception skill with calibrated RGB-D perception using the shipped vision skills and tools. Do not hard-code object poses observed in the privileged phase. Finish only when ./validate.sh reports OK on the vision surface.}
\end{quote}
The conversion session uses the same turn limit as a normal improvement session and has no wall-clock limit. It does not execute the policy or receive new environment feedback, so the first perception iteration evaluates the converted policy exactly as produced by the conversion session. All 21 conversions passed the validator on the first attempt, with conversion costs of 12--39 minutes and \$5--15 in API usage. The converted checkpoint then becomes the starting policy for the perception phase.

\subsection{Failure Analysis of Unsuccessful Transfer}
\label{sec:curriculum-transfer}

\begin{figure}[t]
  \centering
  \includegraphics[width=\linewidth]{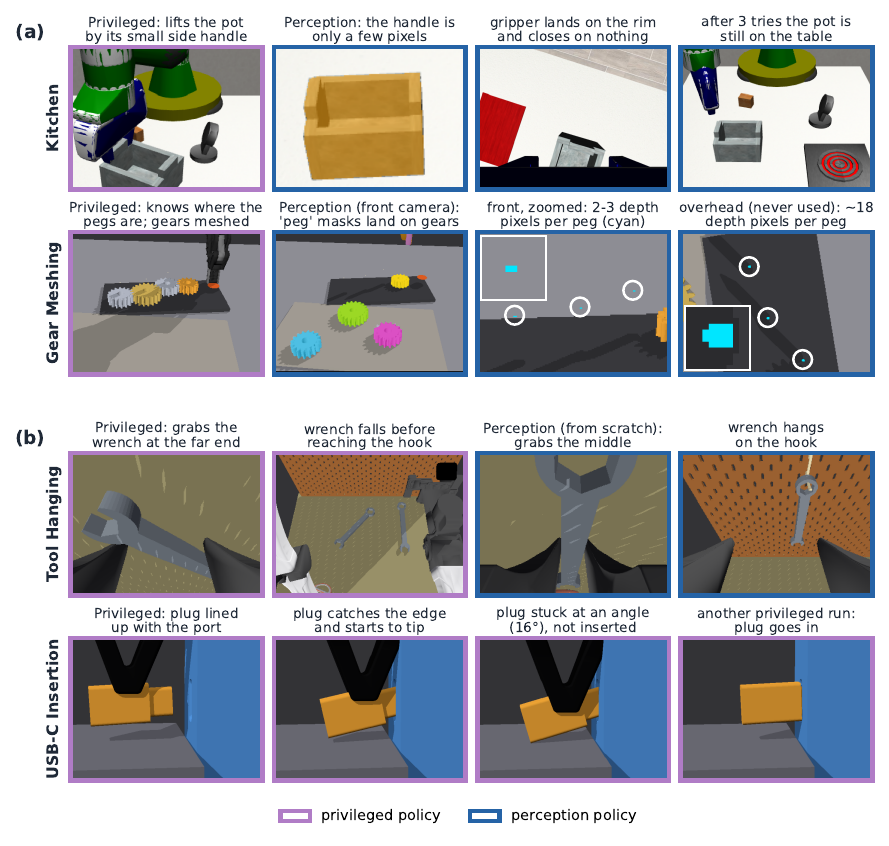}
  \caption{
  Representative failures of privileged-to-perception learning.
  \textbf{(a)} Policies learned with exact geometry can rely on geometric precision that perception cannot reliably recover after transfer. In Kitchen, the narrow handle tab is difficult to localize precisely; in Gear Meshing, the short pegs are difficult to detect.
  \textbf{(b)} Exact geometry alone may be insufficient for learning robust manipulation strategies. In Tool Hanging, the policy learns an unstable wrench grasp; in USB-C Insertion, it aligns the plug but fails during contact.
  Purple borders indicate privileged policies, and blue borders indicate perception policies.
  }
  \label{fig:ears}
\end{figure}

We further analyze the failure modes observed in the curriculum study from Section~\ref{sec:curriculum-design}. Figure~\ref{fig:ears} illustrates representative examples and reveals two distinct limitations of privileged-to-perception learning. First, policies learned with exact geometry may rely on a level of geometric precision that perception cannot reliably recover after transfer. Second, exact geometry alone may be insufficient for learning the object affordances and contact strategies required for robust manipulation.

\paragraph{Dependence on privileged geometric precision.}

Policies learned with exact geometry can inherit manipulation strategies that are highly sensitive to localization errors after transfer to perception, as illustrated in Figure~\ref{fig:ears}(a). In Kitchen, the privileged policy in one run achieves 90--100\% held-out success by pinching a narrow tab on the pot handle using exact geometry. After transfer to perception, success drops to 0--30\% and eventually reaches 0\%. The agent repeatedly identifies the tab grasp as the source of failure, but modifies its reaching and retry behavior rather than adopting a more robust grasp. Because the usable grasp region is only a few millimeters wide, the inherited strategy is highly sensitive to perceptual error.

Gear Meshing shows a related failure in one run. Held-out success drops from 70--90\% before transfer to at most 7\% afterward. The inherited policy can still grasp and carry the gears, but after transfer it must recover the peg locations from vision. The policy uses the front camera, where each peg occupies only 2--3 raised depth pixels, rather than the overhead camera, where each occupies about 18 pixels (Figure~\ref{fig:ears}(a)). As a result, the short pegs are often merged with the plate or missed entirely, leaving the policy without a valid placement target. In contrast, a perception-only run that uses the overhead camera reaches 97--100\% held-out success. Subsequent iterations mainly try to repair peg detection, rather than changing the inherited camera choice or manipulation strategy. Together with Kitchen, this shows that privileged learning can produce effective strategies that become brittle when they depend on geometric precision that perception cannot reliably recover.

\paragraph{Manipulation failures despite exact geometry.}

Exact geometry alone may also be insufficient for understanding object affordances and selecting robust manipulation strategies, as illustrated in Figure~\ref{fig:ears}(b). In Tool Hanging, the privileged state provides the exact wrench pose but does not specify where or how the wrench should be grasped for reliable transport. Consequently, the agent learns to grasp the wrench near its far end, resulting in an unstable grasp during the subsequent carry motion. Because the resulting failure becomes apparent only during transport, the agent struggles to attribute it to the earlier grasp decision and fails to learn an effective strategy. Privileged training never exceeds 10\% held-out success in any of the three runs, whereas perception-only learning reaches 60--100\%.

USB-C Insertion reveals a related limitation in contact reasoning. In one privileged training run, the policy consistently aligns the plug with the bore but fails during insertion because contact with the socket causes the plug to tilt and become wedged. The controller fails to recover from this contact-induced error, resulting in 0\% success. These examples show that exact geometry can reduce localization error without providing the affordance understanding and contact reasoning needed for robust manipulation.

\end{document}